\documentclass[default,iicol]{sn-jnl}

\usepackage{graphicx}
\usepackage{subfigure}

\jyear{2021}%

\theoremstyle{thmstyleone}%
\theoremstyle{thmstyletwo}%

\theoremstyle{thmstylethree}%

\begin{document}

\title[Article Title]{Large-scale Multi-Modal Pre-trained Models: A Comprehensive Survey}  


\author[1,2]{\fnm{Xiao} \sur{Wang}}
\author[1,3]{\fnm{Guangyao} \sur{Chen}}
\author[1]{\fnm{Guangwu} \sur{Qian}}
\author[1]{\fnm{Pengcheng} \sur{Gao}}
\author[1,4]{\fnm{Xiao-Yong} \sur{Wei}}
\author[1]{\fnm{Yaowei} \sur{Wang}$^{(\textrm{\Letter})}$} 
\author[1,3]{\fnm{Yonghong} \sur{Tian}$^{(\textrm{\Letter})}$}
\author[1,3]{\fnm{Wen} \sur{Gao}}

\affil[1]{\orgname{Pengcheng Laboratory}, \orgaddress{\city{Shenzhen} \postcode{518055},  \country{China}}}
\affil[2]{\orgdiv{School of Computer Science and Technology}, \orgname{Anhui University}, \orgaddress{\city{Hefei} \postcode{230601},  \country{China}}} 
\affil[3]{\orgdiv{School of Computer Science}, \orgname{Peking University}, \orgaddress{\city{Beijing} \postcode{100871},  \country{China}}}
\affil[4]{\orgdiv{College of Computer Science}, \orgname{Sichuan University}, \orgaddress{\city{Chengdu} \postcode{610065},  \country{China}}}



\abstract{
With the urgent demand for generalized deep models, many pre-trained big models are proposed, such as BERT, ViT, GPT, etc. Inspired by the success of these models in single domains (like computer vision and natural language processing), the multi-modal pre-trained big models have also drawn more and more attention in recent years. In this work, we give a comprehensive survey of these models and hope this paper could provide new insights and helps fresh researchers to track the most cutting-edge works. Specifically, we firstly introduce the background of multi-modal pre-training by reviewing the conventional deep learning, pre-training works in natural language process, computer vision, and speech. Then, we introduce the task definition, key challenges, and advantages of multi-modal pre-training models (MM-PTMs), and discuss the MM-PTMs with a focus on data, objectives, network architectures, and knowledge enhanced pre-training. After that, we introduce the downstream tasks used for the validation of large-scale MM-PTMs, including generative, classification, and regression tasks. We also give visualization and analysis of the model parameters and results on representative downstream tasks. Finally, we point out possible research directions for this topic that may benefit future works. In addition, we maintain a continuously updated paper list for large-scale pre-trained multi-modal big models: \url{https://github.com/wangxiao5791509/MultiModal_BigModels_Survey}. 
}


\keywords{Multi-modal, Pre-trained Model, Information Fusion, Representation Learning, Deep Learning} 



\maketitle

\section{Introduction} \label{sec1}
Along with the breakthroughs of recognition performance of AlexNet~\cite{krizhevsky2012AlexNet} on the ImageNet competition~\cite{deng2009imagenet}, the artificial intelligence have developed greatly. Many representative deep neural networks are proposed, such as VGG~\cite{simonyan2014VGG}, ResNet~\cite{he2016resnet}, Inception~\cite{szegedy2017inception}, LSTM~\cite{hochreiter1997long}. The researchers usually collect and annotate some samples for their task, and train their models based on pre-trained backbones on large-scale datasets (such as ImageNet~\cite{deng2009imagenet} for computer vision, Glove~\cite{pennington2014glove} and Skip-thought vectors~\cite{kiros2015skipthoughtvectors} for natural language processing). Many tasks can be solved well in such an end-to-end manner compared with traditional handcrafted features, such as object detection, segmentation, and recognition. However, the generalization ability of obtained deep model is still limited. Collecting and annotating a larger dataset can address these issues to some extent, but this procedure is expensive and tedious.

To address this issue, Ashish et al. propose the Transformer network~\cite{vaswani2017attention} which achieves new SOTA (State-Of-The-Art) performance on machine translation task. After that, the self-supervised pre-training on large-scale corpus, then, fine-tuning on downstream tasks attracts more and more researchers' attention. Many pre-trained big models are proposed by following such paradigm, such as BERT~\cite{kenton2019bert}, GPT~\cite{xia2021xgpt, brown2020language}, T5~\cite{raffel2020T5}, XLNet~\cite{yang2019xlnet} which also trigger new research highlights of pre-training in CV community. More and more large-scale NLP and CV models demonstrate the powerful effect by pretrain-and-finetuning paradigm, including ViT~\cite{dosovitskiy2020ViT} and Swin-Transformer~\cite{liu2021swinTransformer}.

\begin{figure*}[!htp]
\center
\includegraphics[width=6in]{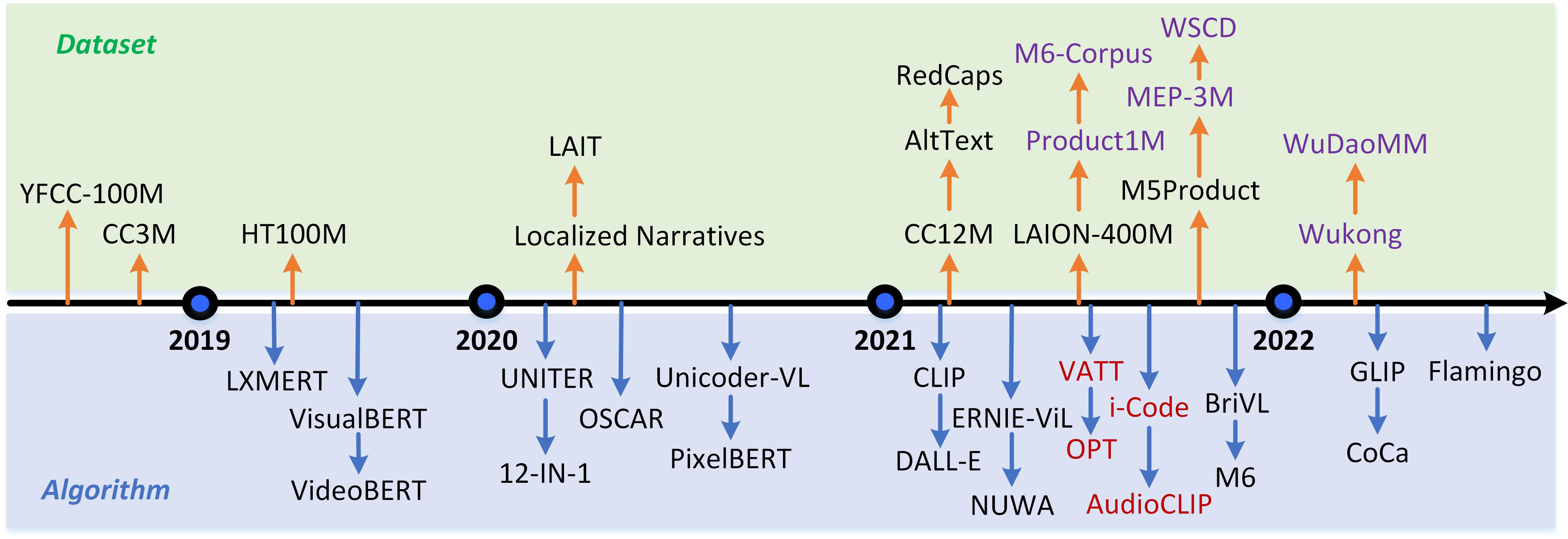}
\caption{The chronological milestones on multi-modal pre-trained big models from 2019 to the present (June 2022), including multi-modal datasets (as shown by the orange arrow) and representative models (as shown by the blue arrow). The purple font indicates that the dataset contains Chinese text (other datasets contain English text). The models highlighted in wine red are trained on more than two modalities.}   
\label{milestone}
\end{figure*}

Although the progress brings new impetus to the development of artificial intelligence, however, the issues caused by the defect of single modality are still hard to solve. Researchers attempt to incorporate more modalities to bridge the data gap for deep models. Many multi-modality fusion based tasks are also explored in a traditional deep learning manner, such as RGB, Depth, Natural Language, Point Cloud, Audio, Event stream, etc. Many large-scale pre-trained multi-modal models~\cite{li2020oscar, chen2020uniter, li2021DeCLIP, huang2020pixelBERT, jia2021ALIGN, liu2021opt, cheng2022hybridDistillation} are proposed which set new SOTA on downstream tasks one after another, as shown in Fig.~\ref{milestone}. In this paper, we give a comprehensive review of these works which target to help the new researchers who are interested in this area to understand the history and latest developments quickly.

\begin{figure*}[!htp]
\center
\includegraphics[width=6in]{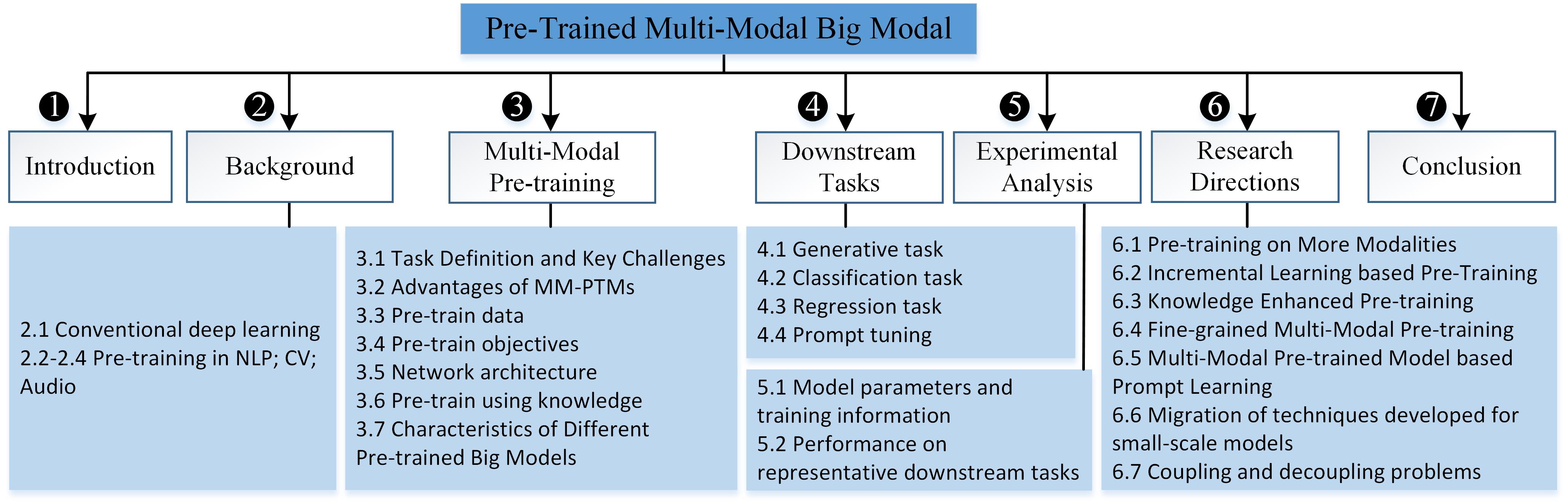}
\caption{\textcolor{black}{The overall framework of this survey.}}  
\label{framework}
\end{figure*}

\textbf{Organization of our review.}   
In this paper, we firstly review the background of multi-modal pre-training technique in Section~\ref{background}, from the traditional deep learning paradigm to pre-training in single modality tasks, including natural language processing, computer vision, and automatic speech processing. 
Then, we focus on MM-PTMs and describe the task definition, key challenges, and benefits, in Section~\ref{taskDefinitionChallenges} and~\ref{benefitsMMPTMs}. The key components are also reviewed in the following sub-sections, including large-scale data, network architectures, optimization objectives, and knowledge-enhanced pre-training. 
To validate the effectiveness of pre-trained models, many downstream tasks are used for quantitative assessment. In Section~\ref{downstreamTasks}, we provide detailed reviews on the task definition and evaluation metrics of these tasks. 
In Section~\ref{experimentalAnalysis}, we review the model parameters and hardware for training and also report the experimental results of several representative downstream tasks. 
Finally, in Section~\ref{researchDict}, we conclude this survey and propose multiple research directions needed to be studied. The architecture of this survey is visualized in Fig.~\ref{framework}.

\textbf{Difference from existing reviews.} 
Although there are already two surveys~\cite{chen2022vlp, du2022VLPsurvey} proposed for MM-PTMs, the difference between our survey and existing ones can be summarized as follows: 
\begin{itemize}
\item \textbf{Scope:} Existing multi-modal surveys~\cite{chen2022vlp, du2022VLPsurvey} focus on vision-language only, however, the multi-modal information problem is a wider research topic. This paper is more comprehensive than the aforementioned reviews by introducing more modalities, such as audio, video, table, etc.  
\item \textbf{Timeliness:} This paper introduces the latest datasets and algorithms (from the year 2019 to June 2022) proposed for multi-modal pre-training which is a long survey, meanwhile, their work belongs to short paper. 
\item \textbf{New insights to MM-PTMs:} By classifying and analyzing the existing MM-PTMs from different perspectives, this article can help readers master the cutting-edge methods and techniques from both detailed and high-level perspectives. In addition, our proposed research directions on the MM-PTMs are deliberate and will provide new clues for the follow-up research. 
\end{itemize}

\begin{table*}[htp]
\center
\scriptsize 
\caption{Summary of related single- and multi-modal pre-training surveys. SC and DC denotes Single Column and Double Column. Pub. is short for Publication.}     
\label{datasetList}
\begin{tabular}{c|l|c|c|c|ccccccccccc}
\hline
\textbf{No.} &\textbf{Title} &\textbf{Year} &\textbf{Pub.}    &\textbf{Topic} &\textbf{Pages} \\ 
\hline  
01  &\makecell[l]{A short survey of pre-trained \\ language models for conversational \\ ai-a new age in nlp~\cite{zaib2020shortsurvey}}    
    &2020   &ACSWM   &NLP   &DC, 4	\\  
\hline
02  &\makecell[l]{A Survey of Controllable Text  \\ Generation using Transformer-based  \\ Pre-trained Language Models ~\cite{zhangHanqing2022survey}}    
    &2022   &arXiv  &NLP   &SC, 34	\\  
\hline
03  &\makecell[l]{A Survey of Knowledge  \\ Enhanced Pre-trained Models ~\cite{yang2021survey}}    
    &2021   &arXiv   &KE   &DC, 20	\\  
\hline
04  &\makecell[l]{A Survey of Knowledge-Intensive \\ NLP with Pre-Trained Language Models ~\cite{yin2022survey}}   
    &2022   &arXiv   &KE   &DC, 8	\\  
\hline 
05  &\makecell[l]{Commonsense Knowledge Reasoning \\  and Generation with Pre-trained \\ Language Models: A Survey  ~\cite{bhargava2022CKRSurvey}}    
    &2022   &arXiv   &KE   &DC, 11	\\  
\hline
06  &\makecell[l]{A survey on contextual embeddings  ~\cite{liu2020contextualembedsurvey}}    
    &2020   &arXiv   &NLP   &DC, 13	\\  
\hline
07  &\makecell[l]{Pre-train, prompt, and predict: \\ A systematic survey of prompting methods \\ in natural language processing  ~\cite{liu2021prepromptpredsurvey}}    &2021   &arXiv   &NLP   &SC, 46	\\  
\hline
08  &\makecell[l]{Pre-trained Language Models in \\ Biomedical Domain: A Systematic Survey  ~\cite{wang2021pretrainBiomedicalSurvey}}    
   &2021   &arXiv   &NLP   &SC, 46	\\  
\hline
09  &\makecell[l]{Pre-trained models for natural \\ language processing: A survey  ~\cite{qiu2020pretrainNLPsurvey}}    
   &2020   &SCTS   &NLP   &DC, 26	\\  
\hline
10  &\makecell[l]{Pre-Trained Models: \\ Past, Present and Future  ~\cite{han2021pretrainSurveys}}    
   &2021   &AI Open   &NLP, CV, MM   &DC, 45	\\  
\hline
11  &\makecell[l]{Recent Advances in Natural  \\ Language Processing via Large Pre-Trained  \\ Language Models: A Survey  ~\cite{han2021pretrainSurveys}}    
   &2021   &arXiv   &NLP   &DC, 49	\\  
\hline
12  &\makecell[l]{A Survey of Vision-Language  \\ Pre-Trained Models ~\cite{du2022surveyVLP}}    
   &2022   &arXiv   &MM   &DC, 9	\\  
\hline
13  &\makecell[l]{Survey: Transformer based \\ video-language pre-training  ~\cite{ruan2022TransformerVideosurvey}}    
   &2022   &AI Open  &CV   &DC, 13	\\  
\hline
14  &\makecell[l]{Vision-Language Intelligence: \\ Tasks, Representation Learning, \\ and Large Models  ~\cite{li2022VLISurvey}}    
   &2022   &arXiv   &MM   &DC, 19	\\  
\hline
15  &\makecell[l]{A survey on vision transformer  ~\cite{han2022transformersurvey}}    
   &2022   &TPAMI   &CV   &DC, 23	\\  
\hline
16  &\makecell[l]{Transformers in vision: A survey  ~\cite{khan2021transformerssurvey}}    
   &2021   &CSUR   &CV   &SC, 38	\\  
\hline
17  &\makecell[l]{A Survey of Visual Transformers  ~\cite{liu2021surveyTransformers}}    
   &2021   &arXiv   &CV   &DC, 21	\\  
\hline
18  &\makecell[l]{Video Transformers: A Survey ~\cite{selva2022videoSurvey}}    
   &2022   &arXiv   &CV   &DC, 24	\\  
\hline
19  &\makecell[l]{Threats to Pre-trained Language \\ Models: Survey and Taxonomy~\cite{guo2022threats}}    
   &2022   &arXiv   &NLP   &DC, 8	\\  
\hline
20  &\makecell[l]{A survey on bias in deep NLP~\cite{garrido2021survey}}    
   &2021   &AS   &NLP   &SC, 26	\\  
\hline
21  &\makecell[l]{A Survey of Controllable Text Generation \\ using Transformer-based \\ Pre-trained Language Models~\cite{zhangHanqing2022survey}}    
   &2022   &arXiv   &NLP   &SC, 34	\\  
\hline
22  &\makecell[l]{An Empirical Survey of the Effectiveness \\ of Debiasing Techniques for \\ Pre-Trained Language Models~\cite{meade2021empiricalSurvey}}    
   &2021   &arXiv   &NLP   &DC, 21	\\  
\hline
23  &\makecell[l]{A multi-layer bidirectional transformer \\  encoder for pre-trained word embedding: \\ A survey of BERT~\cite{kaliyar2020BERTsurvey}}    
  &2020   &CCDSE   &NLP   &DC, 5	\\  
\hline
24  &\makecell[l]{Survey of Pre-trained Models \\ for Natural Language Processing~\cite{pengJiajia2021survey}}    
  &2021   &ICEIB   &NLP   &DC, 4	\\  
\hline
25  &A Roadmap for Big Model~\cite{yuan2022roadmapBIGModels} &2022   &arXiv   &NLP, CV, MM   &SC, 200   \\ 
\hline 
26 &\makecell[l]{Vision-and-Language Pretrained \\ Models: A Survey~\cite{Long2022VLPTsurvey}} &2022   &IJCAI   &MM   &DC, 8   \\ 
\hline
27 &\makecell[l]{Multimodal Learning with \\ Transformers: A Survey~\cite{Xu2021MMTransSurvey}}  &2022   &arXiv   &MM   &DC, 23   \\ 
\hline 
\end{tabular}
\end{table*}

\section{Background} \label{background}

\subsection{Conventional Deep Learning} 
With the release of AlexNet~\cite{krizhevsky2012AlexNet}, a series of deep learning models are proposed in the artificial intelligence community. 
These deep models show better capabilities for fitting complex data than conventional machine learning models. 
From the perspective of its development (LeNet~\cite{lecun1998LeNet} $\rightarrow$ AlexNet~\cite{krizhevsky2012AlexNet} $\rightarrow$ VGG~\cite{simonyan2014VGG} $\rightarrow$ ResNet~\cite{he2016resnet} $\rightarrow$ DenseNet~\cite{Huang_2017CVPR_densenet}), we can find that their architectures become deeper and deeper, and the corresponding performance accordingly becomes better. 
The success of these approaches is supported by large-scale annotated training data, such as the ImageNet~\cite{deng2009imagenet} for the classification task. 
The scale of used data is much larger than traditional methods, but it's still limited. 
The pursuit of robustness and generalization performance of machine learning models has never stopped. 

Recently, the results of large-scale pre-trained models obtained by pre-training on massive data are constantly refreshing people's cognition of artificial intelligence.
Compared with previous small-scale deep learning methods, pre-trained big models show obvious advantages in Natural Language Processing (NLP), Computer Vision (CV), and Multi-Modal fields. 
Such a pre-training scheme take full advantage of the large-scale unlabeled data, therefore, getting rid of expensive annotation costs. 
Therefore, the study of large-scale pre-trained models is a feasible and necessary way to explore real intelligence.

\subsection{Pre-training in Natural Language Processing} 
The large-scale pre-trained models~\cite{qiu2020pre, zaib2020short, guo2022threats, yin2022survey, min2021recent, garrido2021survey, liu2021pre} first appeared in the NLP field. Their success is mainly attributed to self-supervised learning and network structures like Transformer~\cite{vaswani2017attention}. Specifically, the advent of Bidirectional Encoder Representations (BERT)~\cite{kenton2019bert} based on self-supervised learning has led to revolutionary performance improvements on a wide variety of downstream tasks by fine-tuned on fewer training data~\cite{wang2019glue}. 
%
Generative Pre-trained Transformers (GPT)~\cite{radford2018improving, radford2019language, brown2020language} further extends the number of parameters and the training data for better performance.
Note that, the GPT-3~\cite{brown2020language} has ten times more parameters than TuringNLP~\cite{rosset2020turing}. It can not only better fulfill the functions of general NLP tasks, but also has some mathematical calculation ability. The success of the GPT-3 model has made it widely used in various fields, such as search engines, chatbots, music composition, graphics, and coding.
XLNet~\cite{yang2019xlnet} is developed based on a generalized permutation language modeling objective, which achieves unsupervised language representation learning. 
PanGu-$\alpha$~\cite{zeng2021pangualpha} is a large-scale pre-trained Chinese model with 200 billion parameters and implemented based on MindSpore Auto-parallel. 
%
NEZHA~\cite{wei2019nezha} is another Chinese pre-trained big model based on BERT proposed by Wei et al. 
More large-scale pre-trained models for NLP can be found in surveys~\cite{zhangHanqing2022survey, qiu2020pretrainNLPsurvey}.

\subsection{Pre-training in Computer Vision}
Inspired by the revolutionary advancement of Transformer for NLP tasks, many large-scale Transformer-based vision models are also proposed in recent years. Chen et al.~\cite{chen2020generative} attempt to auto-regressively predict pixels using a sequence Transformer. The model obtained by pre-training on the low-resolution ImageNet dataset demonstrates strong image representations. The ViT (Vision Transformer) model~\cite{dosovitskiy2020image} directly adopts the pure Transformer to handle the sequence of image patches for classification. Many new SOTA performances are achieved on several downstream CV tasks, including object detection~\cite{carion2020end}, semantic segmentation~\cite{zheng2021rethinking}, image processing~\cite{chen2021pre}, video understanding~\cite{chen2021pre}. The Swin-Transformer~\cite{liu2021swinTransformer} is another milestone for computer vision, as a hierarchical Transformer, it adopts shifted windows for representation learning. 

For the pre-training methods, the Masked Image Modeling (MIM)~\cite{dosovitskiy2020image, chen2020generative} is proposed to learn rich visual representations via masked parts prediction by conditioning on visible context. MIM provides another direction for the exploration of the visual large-scale pre-training model. He et al. propose the MAE~\cite{he2021masked} to re-explore pixel regression in MIM and show more comparable performance on multiple image recognition tasks. BEiT~\cite{bao2021beit} greatly improves MIM's performance via masked visual token prediction, and PeCo~\cite{dong2021peco} finds injecting perceptual similarity during visual codebook learning benefits MIM pre-trained representation.

\subsection{Pre-training in Audio and Speech} 
As one of the most popular modalities, the audio and speech based pre-training also draws the researcher's attention. 
For example, the wav2vec~\cite{schneider2019wav2vec} is the first work that applies contrastive learning to improve supervised speech recognition by learning the future raw audio based on the past raw audio. 
The vq-wav2vec~\cite{schneider2019wav2vec} uses context prediction tasks from wav2vec to learn the representations of audio segments. 
Discrete-BERT~\cite{baevski2019effectiveness} is BERT-style model by finetuning the pre-trained BERT models on transcribed speech.
HuBERT~\cite{hsu2021hubert} uses self-supervised speech learning where an offline clustering step is used to generate discrete labels of masked speech signals.  
wav2vec 2.0~\cite{baevski2020wav2vec} solves a contrastive task to predict the masked latent representation. 
w2v-BERT~\cite{chung2021w2v} uses contrastive learning and masked speech modeling simultaneously, where a model predicts discretized speech tokens and another model solves a masked prediction task.

\begin{figure}[!htp]
\centering
\includegraphics[width=0.5\textwidth]{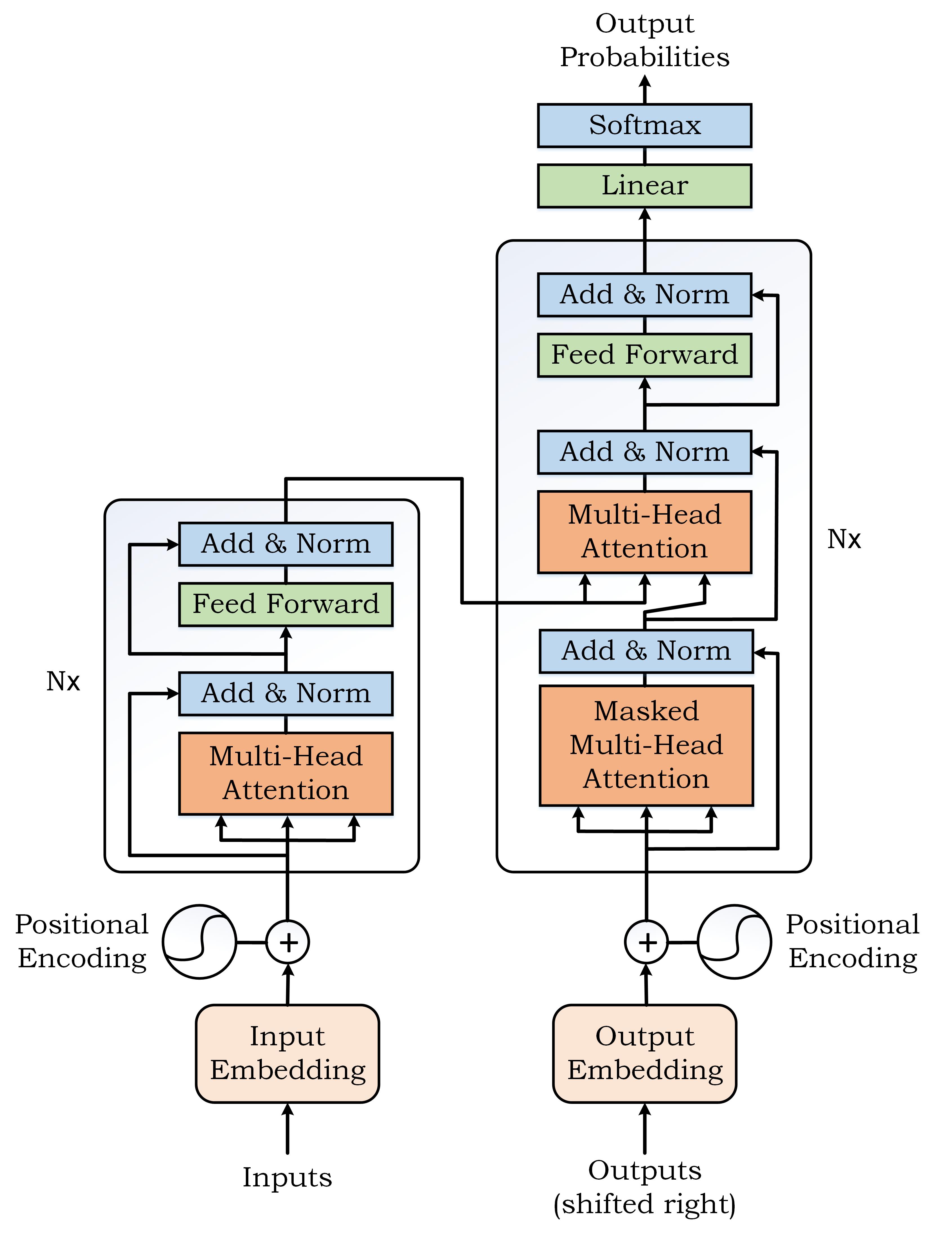}
\caption{The detailed network architecture of Transformer network~\cite{vaswani2017attention}.} 
\label{fig_transformer}
\end{figure}

\section{Multi-Modal Pre-training}   

\subsection{Task Definition and Key Challenges}  \label{taskDefinitionChallenges}

\begin{figure}
\center
\includegraphics[width=3in]{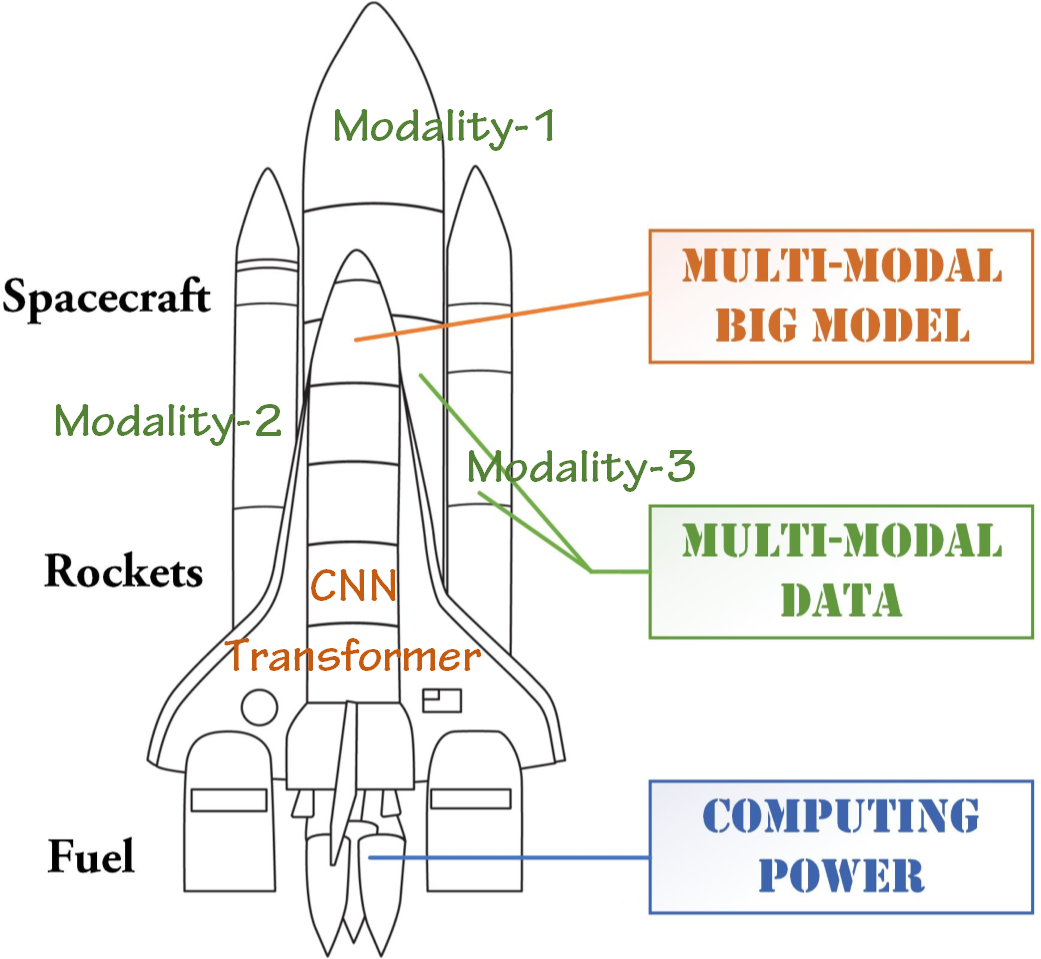}
\caption{\textcolor{black}{The relations between multi-modal data, model, and computing power.}}   
\label{MM_PTMs}
\end{figure}

\textbf{Task Definition.} 
Usually, the deep neural networks are trained on a large-scale dataset, for example, the widely used residual network~\cite{he2016resnet} are pre-trained using a classification task on the ImageNet dataset~\cite{deng2009imagenet}. In contrast, the multi-modal pre-training big models are usually trained on a massive training dataset. Usually, these data are not annotated with labels due to the scale are too large to annotate. On the other hand, the parameters need to reach a certain scale. As illustrated in Fig.~\ref{MM_PTMs}, the multi-modal data, big model, and computing power are tightly connected. All in all, with the support of computing power, the multi-modal pre-training usually denotes the task that the multi-modality model with huge parameters pre-trained on the massive multi-modal data in an unsupervised way.

\noindent \textbf{Key Challenges.} It is challenging to attain a great multi-modal pre-training big model according to aforementioned process. More in detail, we summarize the following key challenging factors: 

$\bullet$ \textbf{Acquisition and clean of large-scale multi-modal data.} 
The multi-modal data is one of the most important elements in MM-PTMs. The collection of multi-modal data is significantly harder than the single one, due to the scarce of multi-modal imaging devices. The frequently used multi-modal cameras are usually covers two modalities only, such as RGB-Depth, RGB-Thermal, RGB-Radar, RGB-Event cameras, etc. Most of current MM-PTMs are vision-language models, because of the easy access to image and text data from the Internet. But the additional cleaning of these data is also necessary due to the noisy samples.

$\bullet$ \textbf{Design of network architectures for large-scale multi-modal pre-training.}
The network architecture is another key component for multi-modal pre-training. The networks used for feature encoding of multiple input modalities are worthy carefully tailored, as different modalities may have their own features and particular networks are needed. For example, the Transformer or CNN are suggested for image and text modality, the spiking networks can be used for event streams. Another problem is the design of multi-modal fusion or cross-modality matching modules. Whether similar modules designed for small-scale multi-modal tasks work for large-scale pre-trained models or not are still remain to be verified.

$\bullet$ \textbf{Design of pre-training objectives.} 
Due to the massive unlabelled multi-modal data, the pre-training tasks usually need to be done in an unsupervised learning manner. Many current works adopt the masked region prediction for each modality as their learning objective. Obviously, the objectives for multi-modal tasks can be directly borrowed from single-modality pre-training, however, the pre-training objectives designed for the multi-modal tasks are also necessary, intuitive and effective. The widely used contrastive learning, modality based matching, and modality translation are all valid and meaningful attempts. How to design new multi-modal pre-training objectives is one of the most challenging tasks for MM-PTMs.

$\bullet$ \textbf{Support of large-scale computing power.} 
The training for traditional deep neural networks can be executed on a server with limited number of GPUs. In contrast, the MM-PTMs needs more computing power due to the large-scale multi-modal data and the super large-scale model parameters. Therefore, the first thing is to prepare a supercomputing device and the subsequent model training also requires a lot of power to support.

$\bullet$ \textbf{Skills on parameter tuning.} 
It is never a simple task to train an effective large model considering aforementioned challenging factors. The tricks used for training the neural networks are also very important. As the research and techniques for the small scale pre-training are relatively more mature, however, there is less accumulation of experience on large-scale pre-training techniques.

\subsection{Advantages of MM-PTMs}  \label{benefitsMMPTMs}

Compared with \emph{single modality pre-trained big models}, the MM-PTMs are more suitable for practical application scenarios. Specifically, the problems like multi-modal collaborative generation, modal completion, cross-domain retrieval, etc, can be addressed well using MM-PTMs. Also, the multi-modal data contains more information which can make up for the defects of a single modality. Therefore, the MM-PTMs can help extracting the common features of multi-modalities. Many recent works demonstrate that the utilization of MM-PTMs indeed brings in the additional prior knowledge~\cite{zhu2022promptUIC, radford2021CLIP, xing2022classprompt}.  

Compared with \emph{small-scale multi-modal models}, the generalizability of MM-PTMs which are obtained by self-supervised/unsupervised learning can be improved significantly. As some prior knowledge is only contained in massive big data, and a small amount of artificially selected annotated data is biased, therefore, it is hard for the small-scale models to master such knowledge.

\begin{table*}[htp]
\center
\scriptsize 
\caption{An overview of multi-modal datasets proposed for large-scale pre-training. Lang. and Ava. is short for Language and Available, respectively.}  
\label{datasetList}
\begin{tabular}{c|l|cccccccccccccc}
\hline
\textbf{No.} &\textbf{Datasets}  &\textbf{Year} &\textbf{Scale}   &\textbf{Modal} &\textbf{Lang.} &\textbf{Ava.}  &\textbf{URL}  \\ 
\hline  
01  & SBU Captions~\cite{ordonez2011im2text}    &2011     &1M  &image-text &English     &\checkmark     	&\href{Link}{http://www.cs.virginia.edu/~vicente/sbucaptions/}     	\\  
\hline
02  & Flickr30k~\cite{young2014image}   &2014     &145K     &image-text     	&English     &\checkmark     	&\href{Link}{http://nlp.cs.illinois.edu/}     	\\
\hline
03  & COCO~\cite{lin2014coco}   &2014     &567K     &image-text     	&English     &\checkmark     	&\href{Link}{https://cocodataset.org/#home}     	\\         
\hline
04  & Visual Genome~\cite{krishna2017visualgenome}     &2017     &5.4M     &image-text     	&English     &\checkmark     	&\href{Link}{https://visualgenome.org/}     	\\         
\hline
05  & VQA v2.0~\cite{goyal2017VQA}    &2017     &1.1M     &image-text &English     &\checkmark     	&\href{Link}{https://visualqa.org/}     	\\  
\hline
06  & FashionGen~\cite{rostamzadeh2018fashion}     &2018     &300k     &image-text     	&English     &\checkmark     	&\href{Link}{https://fashion-gen.com/}     	\\         
\hline
07  & CC3M~\cite{sharma2018conceptual}  &2018     &3M     &image-text     	&English     &\checkmark     	&\href{Link}{https://github.com/google-research-datasets/conceptual-captions}     	\\         
\hline
08  & GQA~\cite{hudson2019gqa}       &2019     &1M     &image-text     &English     &\checkmark &\href{Link}{https://cs.stanford.edu/people/dorarad/gqa/}     	\\         
\hline
09 & LAIT~\cite{qi2020imagebert}      &2020     &10M     &image-text     	&English   &$\times$ &-     	\\       
\hline
10  & CC12M~\cite{changpinyo2021conceptual}     &2021     &12M     &image-text    &English  &\checkmark 	&\href{Link}{https://github.com/google-research-datasets/conceptual-12m}     	\\       
\hline
11  & AltText~\cite{jia2021scaling}   &2021     &1.8B     &image-text     &English  &$\times$	&-     	\\       
\hline
12  & TVQA~\cite{lei2018tvqa}        &2018     &21,793     &video-text     &English     &\checkmark	&\href{Link}{http://tvqa.cs.unc.edu/}     	\\       
\hline
13  & HT100M~\cite{miech2019howto100m}    &2019     &136M     &video-text     &English     &\checkmark	&\href{Link}{https://www.di.ens.fr/willow/research/howto100m}     	\\       
\hline
14  & WebVid2M~\cite{bain2021frozen}  &2021     &2.5M     &video-text     &English    	&\checkmark	&\href{Link}{https://github.com/m-bain/webvid}     	\\       
\hline
15  & YFCC-100M~\cite{thomee2016yfcc100m} &2015     &100M     &image-text     &English     &\checkmark	&\href{Link}{http://projects.dfki.uni-kl.de/yfcc100m/} \\ 
\hline
16  & LAION-400M~\cite{schuhmann2021laion}   &2021     &400M     &image-text     &English     &\checkmark	&\href{Link}{https://laion.ai/laion-400-open-dataset/} \\ 
\hline
17  & RedCaps~\cite{desai2021redcaps}   &2021     &12M     &image-text     &English     &\checkmark	&\href{Link}{https://redcaps.xyz/} \\        
\hline
18  & Wukong~\cite{gu2022wukong}   &2022     &100M     &image-text &Chinese     &\checkmark	&\href{Link}{https://wukong-dataset.github.io/wukong-dataset/index.html} \\    
\hline
19  & CxC~\cite{parekh2021crisscrossed} &2021     &24K     &image-text &English     &\checkmark	&\href{Link}{https://github.com/google-research-datasets/Crisscrossed-Captions} \\   
\hline
20  & Product1M~\cite{zhan2021product1m} &2021     &1M     &image-text &Chinese     &\checkmark	&\href{Link}{https://github.com/zhanxlin/Product1M} \\  
\hline
21  & WIT~\cite{srinivasan2021wit} &2021     &37.5M     &image-text &Multi-lingual     &\checkmark	&\href{Link}{https://github.com/google-research-datasets/wit} \\ 
\hline
22  & JFT-300M~\cite{sun2017JFT300M} &2017     &30M     &image-text &English     &$\times$	&- \\ 
\hline
23  & JFT-3B~\cite{yang2021JFT3B} &2021     &3000M     &image-text &English     &$\times$	&- \\ 
\hline
24  & IG-3.5B-17k~\cite{mahajan2018exploring} &2018     &350M     &image-text &English     &$\times$	&- \\ 
\hline
25  & M6-Corpus~\cite{lin2021m6} &2021   &60M     &image, image-text &Chinese     &$\times$	&- \\ 
\hline
26  & M5Product~\cite{dong2021m5product}  &2021     &6M     &\makecell[c]{image, text, table \\ video, audio \\} &English     &\checkmark	&\href{Link}{https://xiaodongsuper.github.io/M5Product_dataset/index.html} \\ 
\hline
27  & \makecell[l]{Localized \\ Narratives~\cite{pont2020connecting}}  &2020    &849k     &\makecell[c]{image, audio, text, \\ mouse trace \\} &English     &\checkmark	&\href{Link}{https://google.github.io/localized-narratives/} \\  
\hline
28  & RUC-CAS-WenLan~\cite{huo2021wenlan} &2021     &30M     &image-text &Chinese     &$\times$	&- \\ 
\hline
29  & WuDaoMM~\cite{Sha2022WuDaoMM} &2022    &600M     &image-text &Chinese     &\checkmark	&\href{Link}{https://data.wudaoai.cn/home} \\ 
\hline
30  & MEP-3M~\cite{chenmep} &2021 &3M &image-text &Chinese     &\checkmark	&\href{Link}{https://github.com/ChenDelong1999/MEP-3M} \\ 
\hline
31  & WSCD~\cite{fei2021wenlanV2} &2021 &650M &image-text &Chinese &$\times$	&- \\ 
\hline
\end{tabular}
\end{table*}

\subsection{Pre-training Data} 
As shown in Table~\ref{datasetList}, many large-scale multi-modal datasets are proposed for the pre-training task. In this subsection, we will briefly introduce these datasets to help readers quickly master the data information for pre-training. 

$\bullet$ \textbf{SBU Captions}~\cite{ordonez2011im2text} is originally collected by querying Flickr \footnote{\url{https://www.flickr.com/}} using plentiful query terms. Then, they filter the obtained large-scale but noisy samples to get the dataset, which contains more than 1M images with high-quality captions.

$\bullet$ \textbf{Flickr30k}~\cite{young2014image} is obtained by extending Hodosh et al.~\cite{hodosh2013framing} ’s corpus with 31,783 photographs collected from Flickr. These images cover everyday activities, events, and scenes. Five sentences are annotated for each collected image via crowdsourcing, therefore, Flickr30k contains 158,915 captions.

$\bullet$ \textbf{COCO}~\cite{chen2015cococaptions} is developed based on MS-COCO dataset~\cite{chen2015cococaptions} which contains 123,000 images. The authors recruit the Amazon Mechanical Turk \footnote{\url{https://www.mturk.com/}} to annotate each image with five sentences.

$\bullet$ \textbf{Visual Genome}~\cite{krishna2017visualgenome} is proposed to help develop machine learning models that can understand the image by mining the interactions and relationships between objects. Therefore, they perform well on the cognitive tasks, such as the image description and visual question answering, etc. Statistically, the Visual Genome dataset contains more than 108K images and each image has about 35 objects, 26 attributes, 21 pairwise relationships.

$\bullet$ \textbf{VQA v2.0}~\cite{goyal2017VQA} is proposed to reduce the language biases that existed in previous VQA datasets which contain about 1.1M image-question samples and 13M associated answers on 200K visual images from the COCO dataset.

$\bullet$ \textbf{FashionGen}~\cite{rostamzadeh2018fashion} contains 325,536 high-resolution images (1360 $\times$ 1360), each image has a paragraph-length descriptive captions sourced from experts. Six different angles are photographed for all fashion items.

$\bullet$ \textbf{CC3M}~\cite{sharma2018conceptual} is a dataset annotated with conceptual captions proposed in 2018. The image-text samples are mainly collected from the web, then, about 3.3M image-description pairs remained after some necessary operations, such as extract, filter, and transform.

$\bullet$ \textbf{CC12M}~\cite{changpinyo2021conceptual} is the outcome of urgent need of MM-PTMs for large-scale data. The released CC3M dataset is far failed to meet the demand, therefore, the authors further relax the filters used in CC3M for the image and text cleaning. Correspondingly, a four times larger dataset CC12M can be obtained with a slight loss of accuracy.

$\bullet$ \textbf{GQA}~\cite{hudson2019gqa} is mainly proposed for visual reasoning and compositional question answering. A robust question engine is carefully refined by considering \emph{content} and \emph{structure} information. Then, the associated semantic representations are adopted to greatly reduce biases within the dataset and control for its question type composition. Finally, a balanced dataset with 1.7M samples is obtained.

$\bullet$ \textbf{LAIT}~\cite{qi2020imagebert} (Large-scale weAk-supervised Image-Text) is a large-scale image-text dataset collected from the Internet in a weak-supervised manner. It contains about 10M visual images, and each image has a corresponding natural language description which contains about 13 words.

$\bullet$ \textbf{AltText}~\cite{jia2021scaling} is collected by following the rules for constructing Conceptual Captions dataset~\cite{sharma2018conceptual}. To get a large-scale dataset (1.8B image-text pairs), the authors only apply minimal frequency-based filtering for data cleaning. Although the obtained resulting dataset is noisy, the big models obtained by pre-training on this dataset still beats many SOTA works on many downstream tasks.

$\bullet$ \textbf{TVQA}~\cite{lei2018tvqa} is build based on six long-running TV shows from 3 genres, including sitcoms, medical dramas, and crime drama. Then, the Amazon Mechanical Turk is used for VQA collection of video clips. Finally, this dataset contains about $152,545$ question-answer pairs from 21,793 video clips.

$\bullet$ \textbf{HT100M}~\cite{miech2019howto100m} contains about 136 million video clips, which are collected from 1.22 million narrated instructional videos. The content of these videos are mainly focus on humans with a total of 23,000 various tasks. The language description for each clip is an automatically transcribed narration. Therefore, the video and text are weakly-paired, compared with other captioning datasets. 

$\bullet$ \textbf{WebVid2M}~\cite{bain2021frozen} is a video-text captioning dataset which contains over two million video alt-text pairs. These data are collected from the Internet following a similar procedure to CC3M dataset. The authors find that more than $10\%$ of CC3M images are thumbnails from videos, therefore, they scrape these video sources (a total of 2.5M text-video pairs) and create the WebVid2M dataset.

$\bullet$ \textbf{YFCC-100M}~\cite{thomee2016yfcc100m} totally contains 100 million media objects (99.2 million photos, 0.8 million videos) collected from Flickr, the time span of these videos from 2004 and 2014. Note that the YFCC100M dataset is constantly evolving, various expansion packs are unscheduled released.

$\bullet$ \textbf{LAION-400M}~\cite{schuhmann2021laion} contains 400 million image-text pairs which is released for vision-language related pre-training. It is worthy to note that this dataset is filtered using CLIP~\cite{radford2021CLIP} which is a very popular pre-trained vision-language model.

$\bullet$ \textbf{RedCaps}~\cite{desai2021redcaps} is a large-scale dataset with 12M image-text samples collected from 350 subreddits. The authors firstly define the range of subreddit, then, filter the image post and clean the captions. The ethical issue is also considered when building the dataset, and the problematic images are filtered according to privacy, harmful stereotypes, etc.

$\bullet$ \textbf{Wukong}~\cite{gu2022wukong} is the currently largest dataset collected from the Internet which contains 100 million image-text pairs. A list of 200K queries is maintained to ensure the collected samples cover diverse visual concepts. These queries are fed into the Baidu Image Search Engine, then, the image and its corresponding captions can be obtained. Note that each query can get at most 1000 samples to keep a balance between different queries and a series of filtering strategies are adopted for the final Wukong dataset.

$\bullet$ \textbf{CxC}~\cite{parekh2021crisscrossed} is extended based on MS-COCO dataset by rating existing and new pairs with continuous (0-5) semantic similarity. In general, the CxC contains human ratings for 267,095 pairs which is a significant extension in scale and detail. It can be used for a variety of tasks, such as the image-text, text-text, and image-image retrieval, etc.

$\bullet$ \textbf{Product1M}~\cite{zhan2021product1m} contains 1,182,083 image-caption pairs, 458 categories, 92,200 instance. Each image contains about 2.83 objects. Different from regular object detection benchmark datasets, this dataset obtains the instance locations in a paste manner. They first segment the target object, then, paste them into other images based on a given bounding box. It can be used for multiple tasks, including weak-supervised, multi-modal, and instance-level retrieval.

$\bullet$ \textbf{WIT}~\cite{srinivasan2021wit} is constructed by crawling on Wikipedia \footnote{\url{https://www.wikipedia.org/}}. Then, a set of rigorous filtering operations are executed on these data which finally resulting the dataset containing over 37.5 million image-text sets. Note that, the WIT dataset contains multi-lingual, in contrast, other image-text datasets only contain single lingual (for example, English or Chinese).

$\bullet$ \textbf{JFT-300M}~\cite{sun2017JFT300M} contains about 300M images and 375M labels, and each image has about 1.26 labels. Note that, 18291 categories are annotated in this dataset, including 1165 animals and 5720 vehicles, etc. A rich hierarchy is formed according to these categories. It is worthy to note that this dataset is not available online.

$\bullet$ \textbf{JFT-3B}~\cite{yang2021JFT3B} is also an internal Google dataset, which contains about 3 billion images. These samples are annotated in a semi-automatic way with a class hierarchy of 30,000 labels. In other words, this dataset contains large amount of noisy samples. Note that, this dataset is also not available online.

$\bullet$ \textbf{IG-3.5B-17k}~\cite{mahajan2018exploring} is constructed for weakly supervised pre-training by collecting images from Instagram \footnote{\url{https://www.instagram.com/}}. Similar with JFT-300M~\cite{sun2017JFT300M} and JFT-3B~\cite{yang2021JFT3B}, the dataset is also inaccessible and can only be used within the Facebook.

$\bullet$ \textbf{M6-Corpus}~\cite{lin2021m6} is specifically constructed for the pre-training of vision-Chinese big model M6~\cite{lin2021m6}. The samples are collected from various sources, such as the product description, community question answering, forum, etc. It contains 60.5M images and 111.8B tokens.

$\bullet$ \textbf{M5Product}~\cite{dong2021m5product} is a benchmark dataset specifically proposed for E-commerce. It contains 6 million multi-modal samples which cover 6,000 categories, 5,000 attributes, and five modalities, including the visual image, table, video, language description, and audio. It is worthy to note that the M5Product dataset is different from standard multimodal datasets which have completely paired samples, that is to say, each sample may only contain only a subset of modalities. It also has a challenging long-tailed distribution issue.

$\bullet$ \textbf{Localized Narratives}~\cite{pont2020connecting} is proposed by Jordi et al. in 2020, which provides a new form of multi-modal image annotations for the connection of vision and language. The image and corresponding spoken description, textual description, and mouse trace are all embodied in this dataset which provides dense grounding between language and vision. It contains 849k images and covers the whole COCO, Flickr30k, and ADE20K~\cite{zhou2017ade20k} datasets and 671k images of Open Images.

$\bullet$ \textbf{RUC-CAS-WenLan}~\cite{huo2021wenlan} is obtained by crawling multi-source image-text data and totally contains about 30M image-text pairs. These samples covers a wide range of topics and categories, such as the sports, entertainment, news, art, and culture, etc. It plays a fundamental role in the WenLan project and supports the training of the BriVL model~\cite{huo2021wenlan}.

$\bullet$ \textbf{WSCD}~\cite{fei2021wenlanV2} (Weak Semantic Correlation Dataset) is a multi-source dataset, which contains large-scale image-text data samples (650 million). The English texts are all translated into Chinese to support the pre-training of BriVL.

$\bullet$ \textbf{MEP-3M}~\cite{chenmep} is a large-scale image-text dataset collected from several Chinese large E-commerce platforms which contains 3 million image-text pairs of products and 599 classes. Another key feature of this dataset is the hierarchical category classification, in detail, it covers 14 classes, 599 sub-classes, and 13 sub-classes have further sub-subclasses.


\subsection{Pre-training Objectives}  
How to design the learning objectives is a very important step for multi-modal pre-training. Currently, the following learning objectives are proposed, including contrastive loss, generative loss, etc.

\begin{figure*}[!htp]
\center
\includegraphics[width=6.5in]{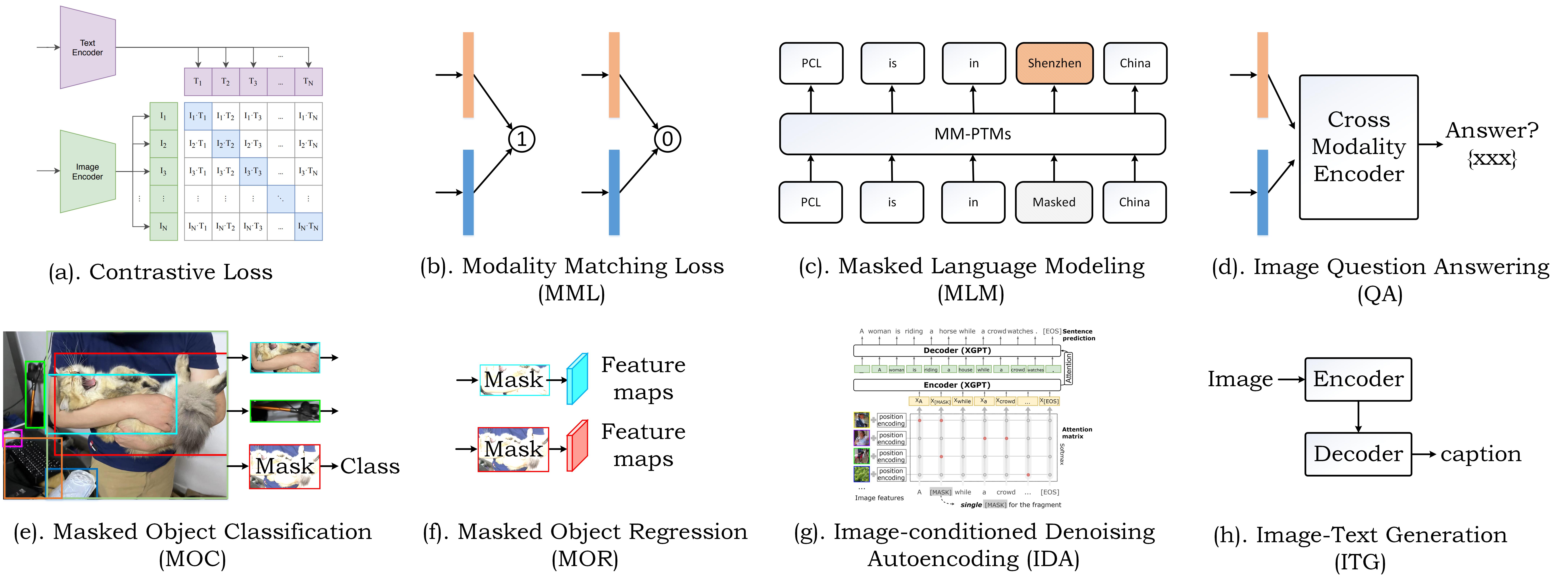}
\caption{\textcolor{black}{Representative pre-training objectives used in MM-PTMs.}}   
\label{pretrainObjectives}
\end{figure*}

$\bullet$ \textbf{Contrastive loss (CS)} function usually constructs positive and negative training samples which is widely used in dual-modality. For example, CLIP~\cite{radford2021CLIP}, ALIGN~\cite{jia2021ALIGN} are all trained using contrastive learning loss. The authors of VinVL~\cite{zhang2021vinvl} adopt the \emph{3-way contrastive loss} for the pre-training to replace the binary contrastive loss function utilized in the Oscar model~\cite{li2020oscar}. 

The contrastive losses in ALIGN are defined as follows:
\begin{equation}
    \begin{split}
        \mathcal{L}_{i2t} &= -\frac{1}{N}\sum_i^N log \frac{exp(x_i^T y_i / \sigma)}{\sum_{j=1}^N exp(x_i^T y_j / \sigma)}\\
        \mathcal{L}_{t2i} &= -\frac{1}{N}\sum_i^N log \frac{exp(y_i^T x_i / \sigma)}{\sum_{j=1}^N exp(y_i^T x_j / \sigma)}\\
        \mathcal{L}_{CL} &= \mathcal{L}_{i2t} + \mathcal{L}_{t2i}
    \end{split}
\end{equation}
where $\mathcal{L}_{i2t}, \mathcal{L}_{t2i}, \mathcal{L}_{CL}$ are an image-to-text classification loss function, a text-to-image classification loss function and the total contrastive loss respectively. The $x_i$ is used to denote the normalized image embedding in the $i$-th pair, while the $y_j$ denote the normalized embedding of text in the $j$-th pair. The $N$ and $\sigma$ are batch size and temperature parameter.

$\bullet$ \textbf{Modality Matching loss (MML)} is widely used in multi-modal pre-training big models due to the explicit or implicit alignment relationships between various modalities. For instance, Unicoder-VL~\cite{li2020Unicodervl} utilizes the Visual-linguistic Matching (VLM) for vision-language pre-training. They extract the positive and negative image-sentence pairs and train their model to predict whether the given sample pairs are aligned or not (in other words, to predict the matching scores). Different from regular negative image-text samples, the authors of InterBERT~\cite{lin2020interbert} design the image-text matching with hard negatives (i.e., ITM-hn) by selecting the highest TF-IDF similarities.

$\bullet$ \textbf{Masked Language Modeling (MLM)} is another widely pre-training objective, usually, the researchers usually mask and fill the input words randomly using special tokens. The surrounding words and corresponding image regions can be used as a reference for the masked word prediction. Wang et al. train SIMVLM~\cite{wang2021simvlm} using the Prefix Language Modeling (PrefixLM), which executes the bi-directional attention on the prefix sequence and auto-regressive factorization on the rest tokens, respectively. The words are denoted as $w = \{x_1,\cdots,x_K\}$, and the image regions as $v = \{v_1,\cdots,v_T\}$. For MLM, the input words is masked as $x_m$ by the mask indices $m$ by generated randomly with a probability of $p\%$
The optimizing goal is to predict the masked words based on all image regions $v$ and remaining words $x_{\lnot m}$, by minimizing the negative log-likelihood:
\begin{equation}
    \mathcal{L}_{MLM}(\theta) = -\mathbb{E}_{(x,v)}\log P_{\theta}(x_m \vert x_{\lnot m},v),
\end{equation}
where $\theta$ is the trainable parameters. Beside MLM, PrefixLM in SIMVLM can also be adopted to pretrain vision-language representation:
\begin{equation}
    \mathcal{L}_{PrefixLM}(\theta) = -\mathbb{E}_{\boldsymbol{x} \sim D}\log P_{\theta}(\boldsymbol{x}_{ \geq T_p} \vert \boldsymbol{x}_{<T_p}),
\end{equation}
where $\boldsymbol{x}$ is the given text sequence, $D$ is the pretraining data and $T_p$ is the length of a prefix sequence of tokens.

$\bullet$ \textbf{Masked Segment Modeling (MSM)} masks a continuous segment of given text using the special token, meanwhile, the MLM masks random words.

$\bullet$ \textbf{Image Question Answering (QA)} is used in LXMERT~\cite{tan2019lxmert} to further expand the pre-training data, as many image-sentence pairs are image and question. The authors train their model to predict the answers as one of their pre-training objectives.

$\bullet$ \textbf{Masked Object Classification (MOC)} mainly focuses on masking the visual images using zero values. Then, people often take the predicted labels by object detector as the ground truth labels. This pre-training objective is widely used, such as Unicoder-VL~\cite{li2020Unicodervl}. Similar to MLM, the image regions can be masked by masking their viusal feature with a prabability of $p\%$. The goal is predict the object category of the masked image regions $v_m^{i}$. The encoder output of the masked image regions $v_m^{i}$ is feed into an FC layer to predict the scores of $T$ object classes, which further goes through a softmax function to be be transformed into a normalized distribution $g_{\theta}(v_m^{i})$. The final objective is:
\begin{equation}
    \mathcal{L}_{MOC}(\theta) = -\mathbb{E}_{(w,v)}\sum^M_{i=1}CE(c(v_m^{i}), g_{\theta}(v_m^{i})),
\end{equation}
where $c(v_m^{i})$ is the ground-truth label.

$\bullet$ \textbf{Masked Object Regression (MOR)} is implemented to regress the masked feature or image regions. For example, the LXMERT~\cite{tan2019lxmert} considers both MOC and MOR for their pre-training.

$\bullet$ \textbf{Image-Text Matching (ITM)} aims to align the image-text data. Negative training data is generated by randomly sampling, including negative sentences for each image, and negative images for each sentence. $y$ is denoted by the gourd truth label for each image-text pair $(v,t)$. A binary classification loss function is used for optimization:
\begin{equation}
    \begin{split}
    \mathcal{L}_{ITM}(\theta) &= -\mathbb{E}_{(v,t)}[y\log s_\theta(v,t) \\
    &+(1-y)\log(1-s_\theta(v,t))],
    \end{split}
\end{equation}
where $s_\theta$ is the image-text similarity score.

$\bullet$ \textbf{Unidirectional LM (UiDT)}  
Single direction history information is used for masked token prediction only, such as \emph{left-to-right} and \emph{right-to-left} language model objectives. Successful stories includes the ELMo~\cite{ELMo2018}, UNILM~\cite{dong2019unilm}.

$\bullet$ \textbf{Bidirectional LM (BiDT)} 
Different from Unidirectional LM which predicts the masked token from a single direction only, the Bidirectional LM considers contextual information from both directions. Therefore, the contextual representations of text can be encoded more accurately. BERT~\cite{kenton2019bert}, UNIML~\cite{dong2019unilm} and VLP~\cite{chen2022vlp} all adopt BiDT as one of their pre-training objective.

$\bullet$ \textbf{Sequence-to-Sequence LM (Seq2seq)} 
is a pre-training objective used in VLP~\cite{chen2022vlp}, etc. It treats the inputs as different parts, each part can attend to different contexts.

$\bullet$ \textbf{Word-Region Alignment (WRA)} is used in UNITER~\cite{chen2020uniter}  which target at explicitly achieves the fine-grained alignment between the multi-modal inputs via Optimal Transport (OT)~\cite{peyre2019computationalOT}. Specifically, the authors learn a transport plan which is a 2D matrix to optimize the alignment and resort to the IPOT algorithm~\cite{xie2020fast} for approximate OT distance estimation. Then, the authors take this distance as the WRA loss to optimize their networks.

$\bullet$ \textbf{Action Prediction (AP)} target at evaluating whether the agent developed for vision-language navigation (VLN) can select the right actions based on the current image and instruction~\cite{hao2020PREVALENT}.

$\bullet$ \textbf{Image-conditioned Denoising Autoencoding (IDA)} is adopted in XGPT~\cite{xia2021xgpt} to align the underlying image-text using an attention matrix. Even without the prior length of the masked fragment, the IDA could still reconstruct the whole sentence successfully.

$\bullet$ \textbf{Attribute Prediction (AttP)} is used to recover the masked tokens of attribute pairs, as indicated in ERNIE-ViL~\cite{yu2021ernieViL}.

$\bullet$ \textbf{Relation Prediction (RelP)} is used in ERNIE-ViL~\cite{yu2021ernieViL} to predict the probability for each masked relation tokens to recover the masked relationship tokens.

$\bullet$ \textbf{Aligned Kaleido Patch Modeling (AKPM)} is proposed for the pre-training of Kaleido-BERT~\cite{zhuge2021kaleidoBERT}, which contains five kaleido sub-tasks, i.e., Rotation Recognition (RR), Jigsaw Puzzle Solving (JPS), Camouflage Prediction (CP), Grey-to-Color Modeling (G2CM), and Blank-to-Color Modeling (B2CM):
\begin{equation}
\begin{split}
    \mathcal{L}_{RR} &= CE(y_r, \mathcal{F}(T,K,\theta)_{K_{1}\_hidden}) \\
    \mathcal{L}_{JPS} &= CE(y_j, \mathcal{F}(T,K,\theta)_{K_{2}\_hidden}) \\
    \mathcal{L}_{CP} &= CE(y_c, \mathcal{F}(T,K,\theta)_{K_{3}\_hidden}) \\
    \mathcal{L}_{G2CM} &= \sum KLD(k_{4i}, \mathcal{F}(T,K,\theta)_{K_{4}\_hidden}) \\
    \mathcal{L}_{B2CM} &= \sum KLD(k_{5i}, \mathcal{F}(T,K,\theta)_{K_{5}\_hidden})
\end{split}
\end{equation}
where $CE$ represents the cross-entropy loss function, $y_r$ denotes the rotation angle, $K_p$ is the hidden output patch of size $p \times p$, $KLD$ denotes the KL-divergence, and $K_p$ are kaleido patches, among which $k_{pi}$ is the masked out ones.

$\bullet$ \textbf{OBject Detection (OBD)} is introduced in the~\cite{xu2021E2EVLP} as a direct set prediction to enhance the pre-training. Also, the authors consider object attribute prediction to learn the fine-grained semantic information. A negative log-likelihood loss is defined for OBD as follows:
\begin{equation}
    \begin{split}
        \hat{\sigma} &= \mathop{\arg\min}\limits_{\sigma \in \mathcal{\phi}_N} \sum_i^N \mathcal{L}_{match}(y_i, \hat{y}_{\sigma(i)}) \\
        \mathcal{L}_{OBD}(y, \hat{y}) &= \sum_{i=1}^N [-log \hat{p}_{\hat{\sigma}(i)}(a_i) - log \hat{p}_{\hat{\sigma}(i)}(c_i) \\
        &+ \mathcal{L}_{box}(b_i, \hat{b}_{\hat{\sigma}(i)}(i))]
    \end{split}
\end{equation}
where $y$ denotes the ground truth set of objects and $\hat{y} = \{\hat{y}_i\}_{i=1}^N$, the number of elements is $N$, $\sigma$ is the cost of a permutation of $N$ elements, $\mathcal{L}_{match}(y_i, \hat{y}_{\sigma(i)})$ denotes the pair-wise matching loss between a prediction with index $\sigma(i)$ and ground truth $y_i$, $\hat{p}_{\hat{\sigma}(i)}(a_i), \hat{p}_{\hat{\sigma}(i)}(c_i)$ denotes the attribute and class probability, $\mathcal{L}_{box}(b_i, \hat{b}_{\hat{\sigma}(i)}(i))$ is a normalized loss of bounding box regression.

$\bullet$ \textbf{Image-Text Generation (ITG)} also plays an important role in the vision-language related pre-training tasks. The aligned image and text are capable of training a model for text generation based on a given image, for example, Xu et al. train the E2E-VLP~\cite{xu2021E2EVLP} with ITG objective:
\begin{equation}
    \mathcal{L}_{ITG} = - \sum_{(x,y) \in (\mathcal{X}, \mathcal{Y})} log \prod_{t=1}^n P(y_t \vert y_{<t}, x)
\end{equation}
where $\mathcal{X}$ represents the visual sequence with context, $\mathcal{Y}$ denotes the generated set of text, and the length of tokens in text $y$ is $n$.

$\bullet$ \textbf{Video-Subtitle Matching (VSM)} considers two targets for the video-text pre-training task, i.e., (i) local alignment, (ii) global alignment, as used in HERO~\cite{li2020hero}. The score functions and the corresponding loss functions are defined as follows:
\begin{equation}
    \begin{split}
        &S_{local}(s_q, \boldsymbol{v}) = \boldsymbol{V}^{temp} \boldsymbol{q} \in \mathbb{R}^{N_v} \\
        &S_{global}(s_q, \boldsymbol{v}) = max(\frac{\boldsymbol{V}^{temp}}{\|\boldsymbol{V}^{temp}\|} \frac{\boldsymbol{q}}{\|\boldsymbol{q}\|}) \\
        &\mathcal{L}_h(S_{pos}, S_{neg}) = max(0, \delta + S_{pos} - S_{neg})\\
        &\mathcal{L}_{local} = -\mathbb{E}_D log(\boldsymbol{p}_{st}[y_{st}] + log(\boldsymbol{p}_{ed}[y_{ed}]) \\
        &\mathcal{L}_{global} = -\mathbb{E}_D [\mathcal{L}_h(S_{global}(s_q, \boldsymbol{v}), S_{global}(\hat{s}_q, \boldsymbol{v})) \\
        & \qquad \qquad + \mathcal{L}_h(S_{global}(s_q, \boldsymbol{v}), S_{global}(s_q, \hat{\boldsymbol{v}}))] \\
        &\mathcal{L}_{VSM} = \lambda_1 \mathcal{L}_{local} + \lambda_2 \mathcal{L}_{global}
    \end{split}
\end{equation}
where $s_q$ denotes the sampled query from all subtitle sentences, $\boldsymbol{v}$ is the whole video clip, $\boldsymbol{V}^{temp} \in \mathbb{R}^{N_v \times d}$ is the final visual frame representation generated by temporal transformer, $\boldsymbol{q}\in \mathbb{R}^d$ is the final query vector, $y_{st}, y_{ed} \in \{1,...,N_v\}$ are the start and end index respectively, $\boldsymbol{p}_{st}, \boldsymbol{p}_{ed} \in \mathbb{R}^{N_v}$ represent probability vectors generated from the scores, $\boldsymbol{p}[y]$ indexes the $y$-th element of the vector $\boldsymbol{p}$, $\mathcal{L}_h$ denotes the combined hinge loss over positive and negative query-video pairs, $(s_q, \boldsymbol{v})$ is a positive pair while $(s_q, \hat{\boldsymbol{v}}), (\hat{s_q}, \boldsymbol{v})$ are negative ones replaced with one other sample in $\boldsymbol{v}$ and $s_q$ respectively, $\delta$ is the margin hyper-parameter and $\lambda_1, \lambda_2$ are balancing factors.

$\bullet$ \textbf{Frame Order Modeling (FOM)} is treated as a classification problem in HERO~\cite{li2020hero}, which targets reconstructing the timestamps of selected video frames. The objective of FOM is defined as follows:
\begin{equation}
    \mathcal{L}_{FOM} = -\mathbb{E}_D \sum_{i=1}^R log \boldsymbol{P}[r_i,t_i]
\end{equation}
where the number of reordered frames is $R$, $i\in[1,R], t_i\in \{1,...,N_v\}$, $r_i$  is the reorder index, $\boldsymbol{P} \in \mathbb{R}^{N_v \times N_v}$ is the probability matrix.

$\bullet$ \textbf{Textual Aspect-Opinion Extraction (AOE)} aims to extract aspect and opinion terms from the text, as noted in~\cite{ling2022vision}. To handle the lack of label information required for supervised learning, the authors resort to other models for aspect extraction and opinion extraction. The obtained aspect and opinion terms are treated as labels for the AOE task.

$\bullet$ \textbf{Visual Aspect-Opinion Generation (AOG)} targets at generating the aspect-opinion pair detected from the input image~\cite{ling2022vision}.

$\bullet$ \textbf{Multimodal Sentiment Prediction (MSP)} enhance the pre-trained models by capturing the subjective information from vision-language inputs~\cite{ling2022vision}.

$\bullet$ \textbf{Modality-Level Masking (MoLM)} is used in~\cite{liu2021opt} to learn the alignment among the text, vision, and audio. The authors mask out each modality independently with a certain probability.

$\bullet$ \textbf{Structural Knowledge Masking (SKM)} is proposed in~\cite{cui2021rosita} which attempts to mask the tokens selectively based on the cue provided by the knowledge entry. The masking probabilities is calculated to obtain mask indices $M_{w}$ and $M_{r}$ for each knowledge entry, the two items denote the words of sentences and visual regions of images need to be masked, respectively. The loss function of Structural Knowledge Masking Language Model can be formulated as:
\begin{equation}
\mathcal{L}_{SKMLM}(\theta)=-\mathbb{E}_{(W,R)\sim D}logP_{\theta}(\mathcal{W}_{M_{w}}\vert\mathcal{W}_{\setminus M_{w}},\mathcal{R}_{\setminus M_{r}})
\end{equation}
where $\theta$ is the parameters. $\mathcal{W}_{\setminus M_{w}}$ and $\mathcal{R}_{\setminus M_{r}}$ represent the non-masked words of sequences and the remaining regions of images, respectively.

\subsection{Pre-training Network Architecture}

\subsubsection{Self-attention and Transformer} 
In the large-scale pre-training era, most of current pre-trained models are inspired by the Transformer (which is mainly consisted of self-attention layers). It is originally developed for natural language processing tasks in 2017~\cite{vaswani2017attention} which sets new SOTA performance on many downstream tasks by a large margin. Such framework is also introduced into the computer vision community, therefore, the design of unified network architectures for various tasks and inputs is the current research hotspot.

\begin{figure}[!htp]
\center
\includegraphics[width=3in]{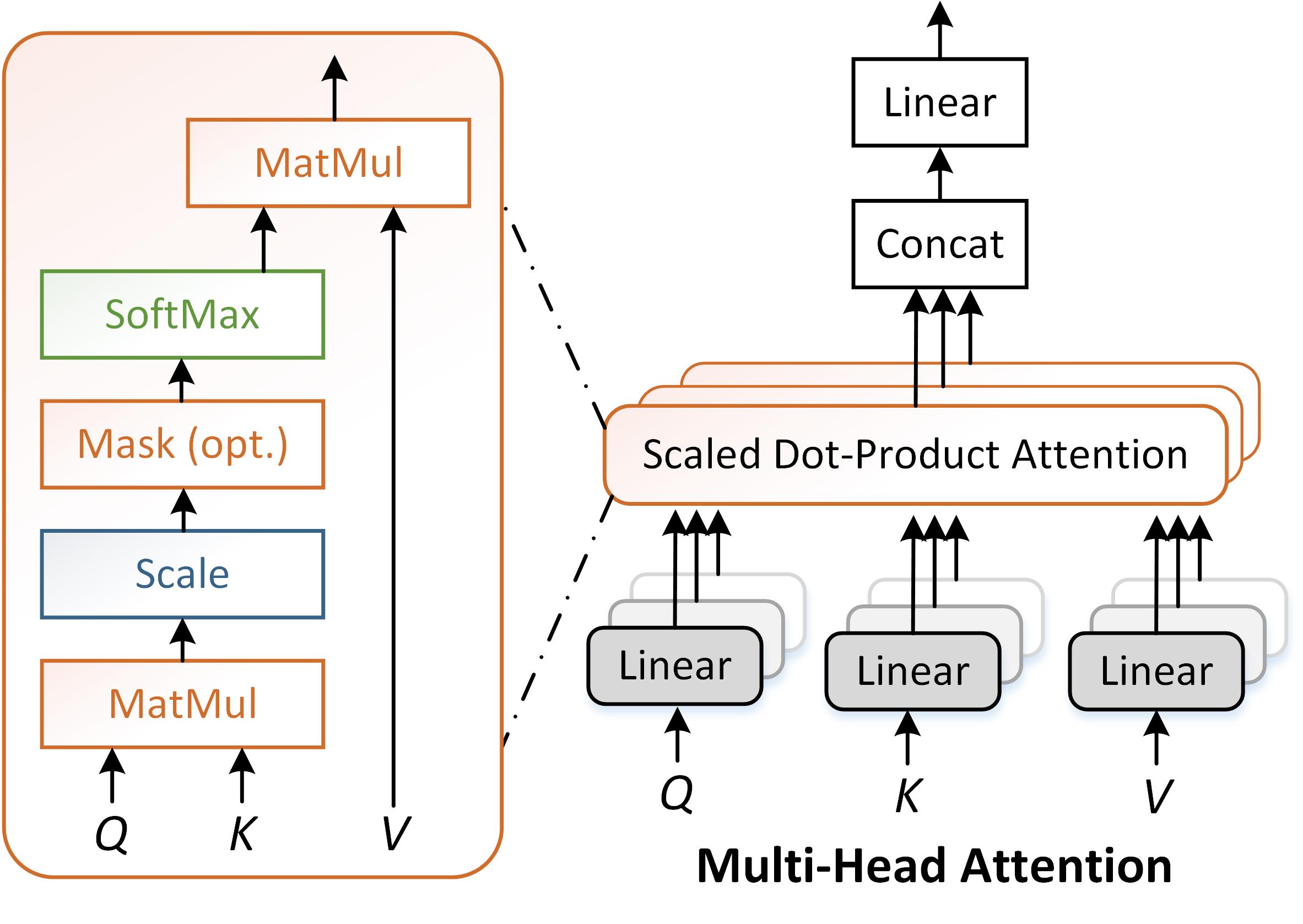}
\caption{An illustration of multi-head self-attention (MHSA)~\cite{vaswani2017attention}.}     
\label{selfattention_MHA}
\end{figure}

Given the input x, an attention module A(x) is used to generate attention weights, then, some procedures are conducted based on input x and A(x) to get the attended input x' = f(A(x), x). Many attention models are designed based on this idea, such as the channel attention, spatial attention, temporal attention, branch attention~\cite{guo2022attentionSurvey}. The self-attention scheme is a special case of attention mechanism, as shown in Fig.~\ref{selfattention_MHA}. More in detail, 
\begin{flalign}
&~~~~ Q, K, V = Linear(x) & \\
&~~~~ A(x) = Softmax(QK) & \\
&~~~~ f(A(x), x) = A(x)V
\end{flalign}
where the Linear denotes fully connected layers. On the basis of self-attention, the work mechanism of multi-head attention is the aggregation of parallel attention layers. Mathematically speaking, 
\begin{flalign} 
\scriptsize 
& MultiHead(Q, K, V) = [head_1, ... , head_h]W^O & \\ 
& head_i = Attention(QW_i^Q, KW_i^K, VW_i^V). 
\end{flalign}
where $[ , ]$ denotes the concatenate operation, $W_i^Q, W_i^K, W_i^V$ and $W^O$ are parameter matrices.

\subsubsection{Single- and Multi-stream} 

The multi-layer transformer is widely used in many current MM-PTMs. The input of each modality is first extracted as feature embeddings by the independent encoder and then interacted with other modalities. According to the manner of multi-modal information fusion, two categories of MM-PTMs can be concluded, i.e., single- and cross-stream. In this subsection, we will present these two architectures separately.

\begin{figure}
\centering
\subfigure[Architecture of single-stream pre-training multi-modal model]
{
\begin{minipage}[b]{0.4\textwidth}
\includegraphics[width=1\textwidth]{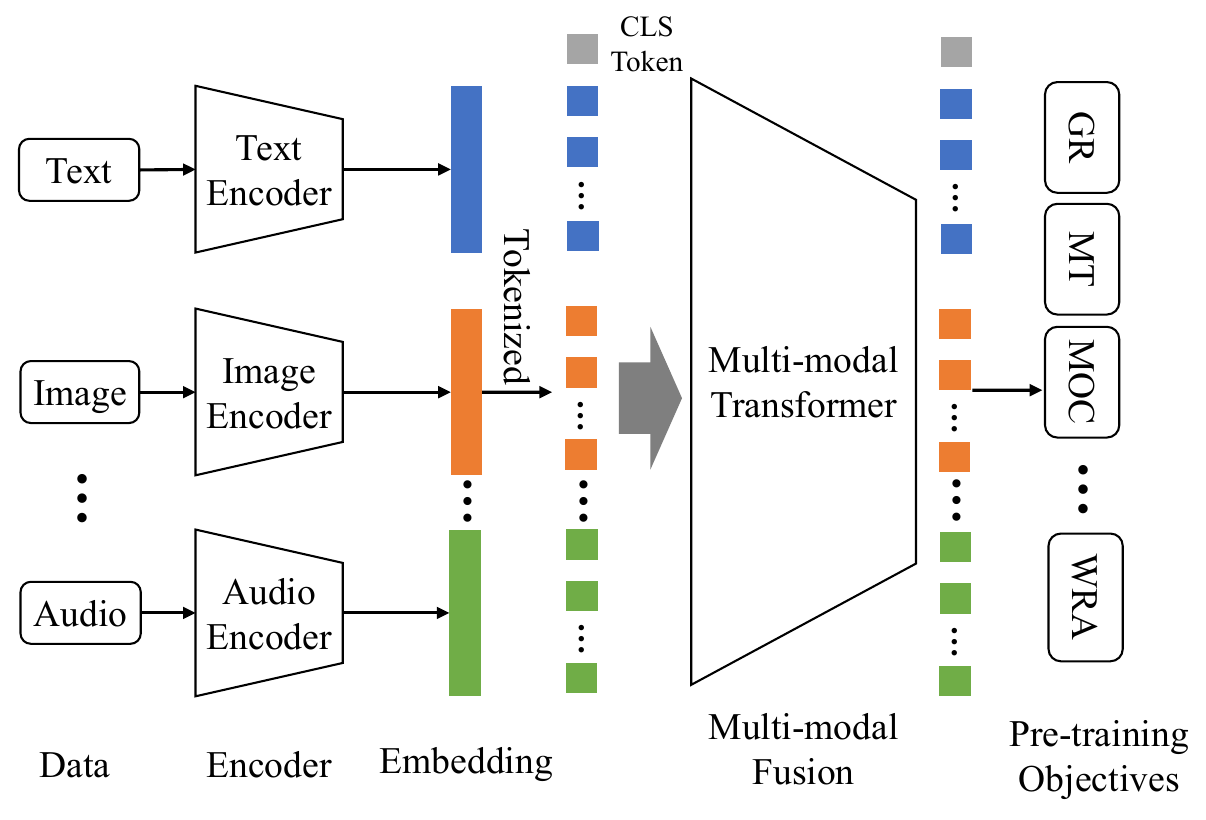}
\label{single}
\end{minipage}
}
\subfigure[Architecture of Cross-stream pre-training multi-modal model]{
\begin{minipage}[b]{0.4\textwidth}
\includegraphics[width=1\textwidth]{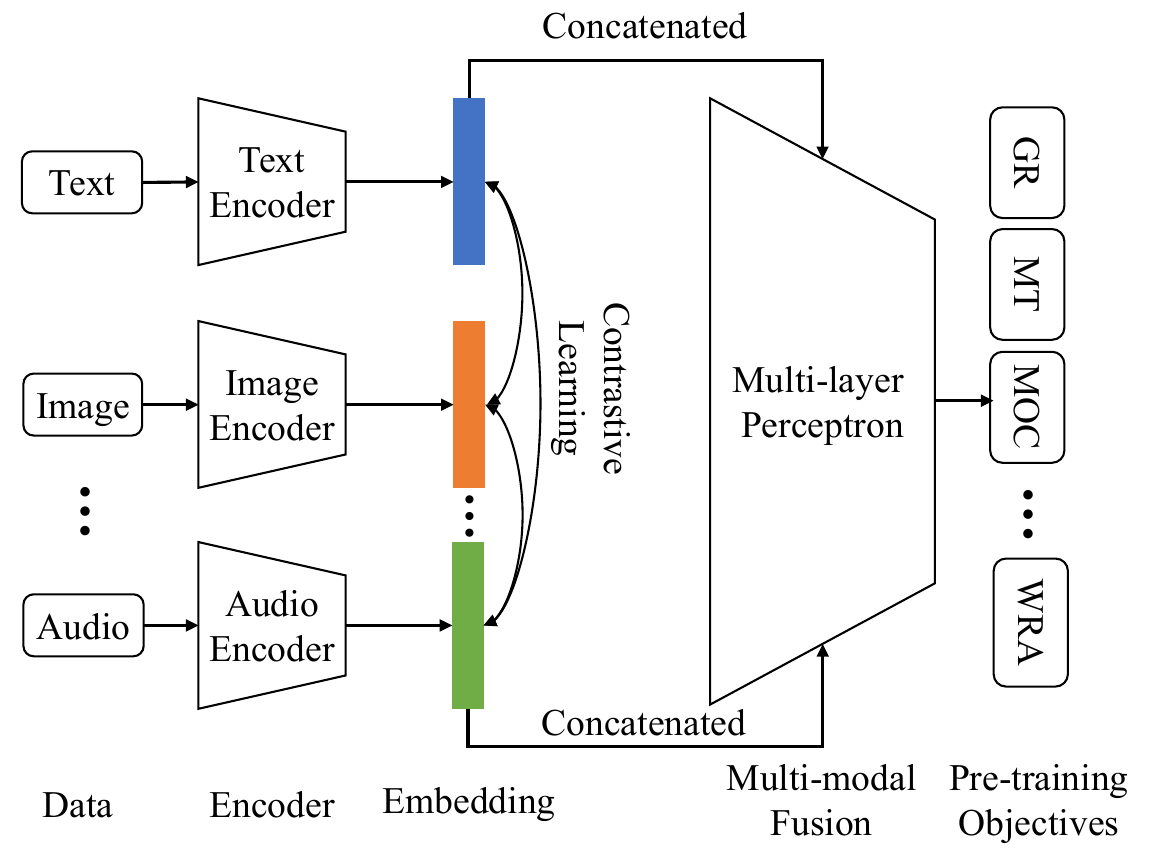}
\label{cross}
\end{minipage}
}
\caption{Pre-training network architecture.}  
\label{architecture}
\end{figure}

$\bullet$ \textbf{Single-stream} Multi-modal inputs such as images and text are treated equally and fused in a unified model. The uni-modal features extracted from each modality are tokenized and concatenated by the separators as the input of the multi-modal transformer for multi-modal fusion, as shown in Fig.~\ref{single}. In the transformer, the MHSA (multi-head self-attention) mechanism is usually adopted to interactively fuse the uni-modal features, then, the multi-modal fusion features are output from the class token of the transformer. Large-scale MM-PTMs based on single-stream structure includes VL PTMs (e.g., Oscar~\cite{li2020oscar} and ALBEF~\cite{li2021ALBEF}) and vision-language-audio pre-training model OPT~\cite{liu2021opt}. Single-stream pre-training models perform token-level matching based on strong semantic correlation, e.g. object features of the image are matched with semantic features of object tags. It provides realistic interaction between uni-modal features, and multi-modal fusion features contain information from different modalities with better characterization capability.

$\bullet$ \textbf{Cross-stream} Features of different modalities are extracted in parallel by independent models and then are aligned by self-supervised contrastive learning in cross-stream architecture. The pre-training models obtain aligned uni-modal features rather than fused multi-modal features. As shown in Fig.~\ref{cross}, multi-modal fusion features are obtained by concatenating uni-modal features and fed into a MLP (Multi-Layer Perceptron) for pre-training objective learning. Representative large-scale MM-PTMs  based on cross-stream structure include BriVL~\cite{huo2021wenlan} and CLIP~\cite{radford2021CLIP}, etc. Compared with pre-training models based on single-stream, cross-stream models align different modality features into a consistent high-dimensional feature space, such as text semantics and visual image representation. Cross-stream pre-training models generally contain the CS pre-training objective and achieve embedding-level matching based on ``weak semantic correlation"~\cite{huo2021wenlan}. The structure of cross-stream models is more flexible, and modifying the branching structure of one modality of the model does not affect other modalities, making it easy to deploy in real scenarios. However, cross-stream models extract the aligned multi-modal common features, and how to effectively exploit the information differences and complementarity between multi-modal data is an issue to be studied.

In addition, depending on the needs of the pre-training objectives, the structure of pre-training models can be divided into with and without a decoder. If pre-training objectives contain generative tasks, such as masked image reconstruction, generating matching images based on the text description, etc., the pre-training model adds a decoder after the encoder for converting multi-modal fusion features into the corresponding output.

\begin{figure*}[!htp]
\center
\includegraphics[width=6in]{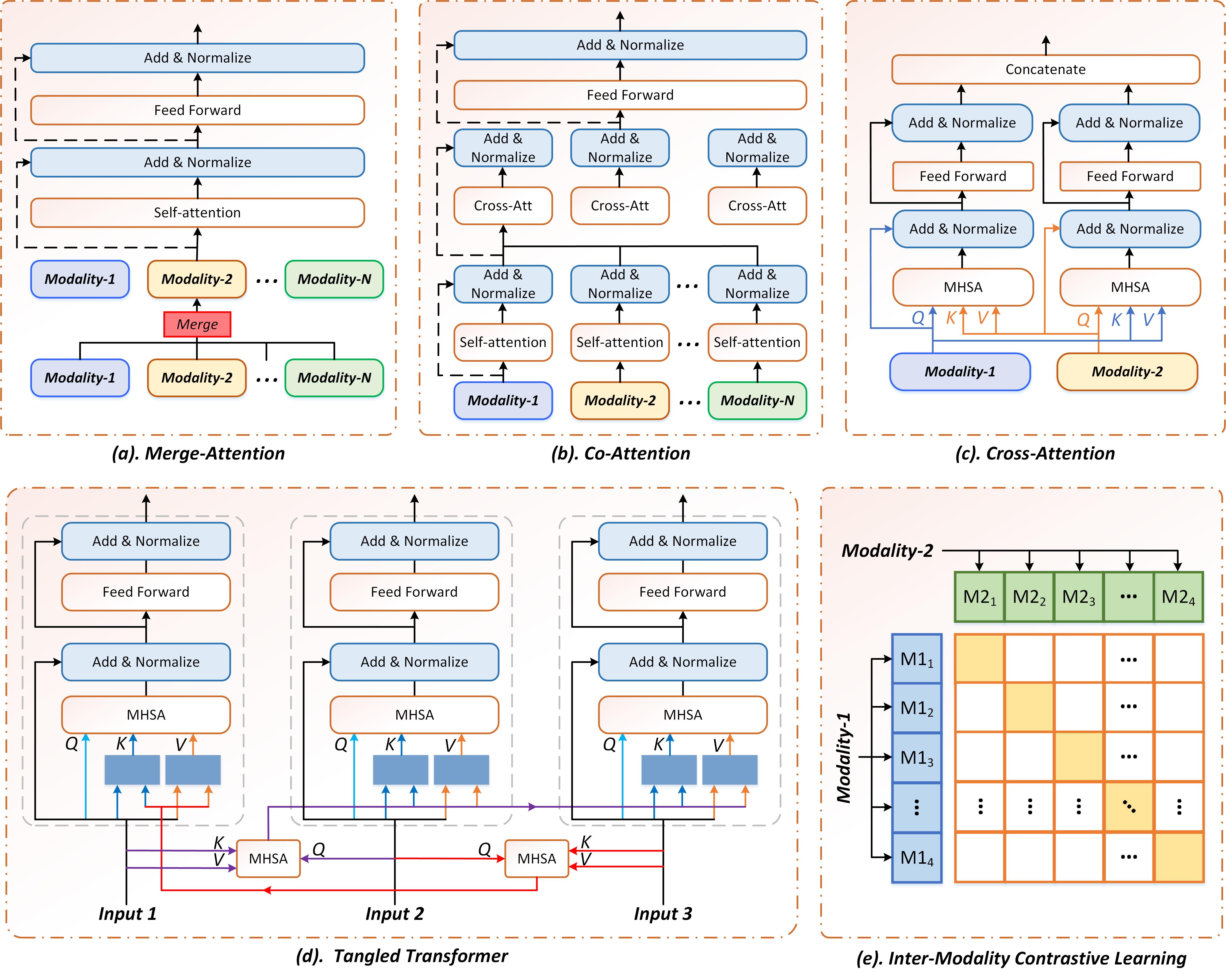}
\caption{\textbf{The widely used modality interactive learning modules for MM-PTMs.} (a) Merge-attention~\cite{yang2022Icode}, (b) Co-attention~\cite{yang2022Icode}, (c) Cross-attention~\cite{SuoS0021ijcai}, (d) Tangled-transformer~\cite{zhu2020actbert}, and (e) Contrastive learning~\cite{radford2021CLIP}.}     
\label{modalityInteractiveLearning}
\end{figure*}

\subsubsection{Modality Interactive Learning} 
Most of current large-scale pre-trained multi-modal models adopt concatenate, add, Merge-attention, Co-attention, and Cross-attention~\cite{SuoS0021ijcai} to achieve interactive learning between modalities. An introduction to these modules are given in the  following paragraphs.

\textbf{$\bullet$ Merge-attention:} 
As shown in Fig.~\ref{modalityInteractiveLearning} (a), a unified feature representation is obtained by concatenating the input modalities. Then, this feature is fed into the fusion network. For example, the i-Code~\cite{yang2022Icode} flatten the visual inputs along the temporal and spatial dimensions. Note that the parameters of this attention model is shared by these input modalities.

\textbf{$\bullet$ Co-attention:} 
For the co-attention module, as shown in Fig.~\ref{modalityInteractiveLearning}, each input modality has its own self-attention layers for modality-specific feature embedding. Then, the multiple embeddings are fused using a cross-attention layer.

\textbf{$\bullet$ Cross-attention:} 
For the multi-modal task, the key step is how to design a fusion module to connect the multi-modality inputs effectively. For instance, the cross-attention layer is proposed by Suo et al.~\cite{SuoS0021ijcai}, which integrate the image and language subtly for visual question answering. Specifically, they mutually input one modality into the Q-branch of another self-attention network. Then, the output of two modalities are concatenated as one unified representation for final prediction.

\textbf{$\bullet$ Tangled-transformer:} 
The TaNgled Transformer (TNT)~\cite{zhu2020actbert} is proposed to handle the action-, regional object-, and linguistic-features, simultaneously, using three Transformer modules. As shown in Fig.~\ref{modalityInteractiveLearning} (d), the authors inject one modality to the Transformer network designed for other modality to enhance the interactions.

\textbf{$\bullet$ Inter-Modality Contrastive Learning:} 
The contrastive learning is widely used for inter-modality relation modelling, such as the CLIP~\cite{radford2021CLIP} and its following-up works~\cite{wang2021actionclip, li2022clipevent, cui2022CLIPbenchmark, li2021DeCLIP, shen2021CLIPViL, Chen2022ProtoCLIP, dong2021m5product}. The representative work SCALE \cite{dong2021m5product} is trained with Self-harmonized Inter-Modality Contrastive Learning (SIMCL), which can be written as: 
\begin{equation}
\tiny  
\label{SIMCLFunction} 
\mathcal{L}_{CL}(d_i^{(0)}, d_i^{(1)}) = -log \frac{exp(Sim(f_i^{(0)}, f_i^{(1)})/\tau)}{\sum^{1}_{m=0} \sum^{N}_{k=1} \mathbf{1}_{[k \neq i]} exp(Sim(f_i^{(m)}, f_k^{(1-m)})/\tau))}, 
\end{equation}
where $(d_i^{(0)}, d_i^{(1)})$ is a positive pair, and the pairing of $d_i^{(0)}$ and other samples will bring us negative training data. $f_i^{(0)}, f_i^{(1)}$ are feature embedding of $(d_i^{(0)}, d_i^{(1)})$ respectively. The $Sim$ denotes the cosine similarity, $\mathbf{1}_{[k \neq i]}$ is the binary indicator function, $\tau$ is a temperature parameter.

\begin{table*}[htp]
\center
\tiny   
\caption{The summary of mainstream multi-modal pre-trained big models (Part-I).}   
\label{PTMList1}
\resizebox{\textwidth}{95mm}{ 
\begin{tabular}{c|l|l|c|c|c|l|c|ccccccc}
\hline
\textbf{No.} &\textbf{Model}  &\textbf{Pub.}   &\textbf{Modality} &\textbf{Architecture} &\textbf{Objective}  &\textbf{Highlights} & \textbf{Parameters} &\textbf{Code}\\  
\hline  
01  &VisualBERT~\cite{li2019visualbert}   &arXiv-2019   &image-text  &\makecell[c]{Trans, \\ BERT}   &GR, MML   &\makecell[l]{A simple and strong \\ baseline for VLP}
& 170M 
&\href{URL}{https://github.com/uclanlp/visualbert}   \\
\hline
02  &ViLBERT~\cite{lu2019vilbert}   &NeurIPS-2019      &image-text   &Trans   &CS, GR   
&\makecell[l]{First adopt co-attention \\ for MM pre-training}
& 274M
&\href{URL}{https://github.com/jiasenlu/vilbert_beta}   \\
\hline
03  &LXMERT~\cite{tan2019lxmert}  &EMNLP-2019      &image-text   &Trans   & \makecell[c]{QA, MOR, MOC, \\ MML, MLM}   
&\makecell[l]{Propose a cross-modality \\ encoder for vision-language \\ pre-training}
& 183M 
&\href{URL}{https://github.com/airsplay/lxmert}   \\
\hline
04  &B2T2~\cite{alberti2019B2T2}   &EMNLP-2019      &image-text &\makecell[c]{ResNet, \\ BERT}   &MML, GR  
&\makecell[l]{Embed bounding box into \\ text transformer in a \\ early fusion manner}   
& - 
&\href{URL}{https://github.com/google-research/language/tree/master/language/question_answering/b2t2}   \\
\hline
05  &Unicoder-VL~\cite{li2020Unicodervl}    &AAAI-2020 &image-text   &Trans   &GR, MML, MOC   
&\makecell[l]{Single transformer encoder \\ for VLP}    
& 170M 
&\href{URL}{https://github.com/microsoft/Unicoder}   \\
\hline
06  &VL-BERT~\cite{su2019vlbert}   &ICLR-2019 &image-text   &BERT   &GR, MOC    
&\makecell[l]{MM PTMs and faster rcnn \\ are jointly trained}
& -
&\href{URL}{https://github.com/jackroos/VL-BERT}   \\
\hline
07  &VLP~\cite{zhouLuowei2020VLP}   &AAAI-2020      &image-text   &Trans   &BiDT, Seq2seq    
&\makecell[l]{Unified encoder-decoder \\ network architecture}
& - 
&\href{URL}{https://github.com/LuoweiZhou/VLP}   \\
\hline
08  &UNITER~\cite{chen2020uniter}   &ECCV-2020      &image-text   &Trans   &MRA, MML   
&\makecell[l]{Propose an OT-based Word-\\Region Alignment objective} 
& 110M 
&\href{URL}{https://github.com/ChenRocks/UNITER}     \\
\hline
09  &12-IN-1~\cite{lu20201212in1}     &CVPR-2020     &image-text   &Trans   &CS, GR 
&\makecell[l]{Training jointly on 12 different \\ datasets in a multi-task \\ learning manner}
& 270M
&\href{URL}{https://github.com/facebookresearch/vilbert-multi-task}   \\
\hline
10  &VisDial-BERT~\cite{murahari2020vdialogbert}   &ECCV-2020      &image-text   &Trans   &MLM, NSP, MIR   
&\makecell[l]{Pre-training on image-text \\ corpus and finetuning \\ on visual dialog}
& - 
&\href{URL}{https://github.com/vmurahari3/visdial-bert/}   \\
\hline
11  &ImageBERT~\cite{qi2020imagebert}   &arXiv-2020  &image-text   &Trans   &\makecell[c]{MOC, MLM, \\ MML, MOR}   
&\makecell[l]{Indicating that multi-stage \\ pre-training works better} 
& 170M
&-   \\  
\hline
12  &PREVALENT~\cite{hao2020PREVALENT}   &CVPR-2020  &image-text   &Trans   &MLM, AP    
&\makecell[l]{Pre-training for vision \\ and language navigation} 
& - 
&\href{URL}{https://github.com/weituo12321/PREVALENT}   \\
\hline
13  &XGPT~\cite{xia2021xgpt}   &NLPCC-2021  &image-text   &Trans  &\makecell[c]{IC, MLM, \\ IDA, MOR}   
&\makecell[l]{Novel IDA pre-training; \\ Share parameters between \\ encoder and decoder}  
& - 
&-   \\
\hline
14  &InterBERT~\cite{lin2020interbert}   &arXiv-2020 &image-text   &Trans   &\makecell[c]{MSM, MOC, \\ ITM-hn} 
&\makecell[l]{Finding that all-attention \\ works better than co-attention \\ for modal interaction}  
& 173M
&\href{URL}{https://github.com/black4321/InterBERT}   \\
\hline
15  &PixelBERT~\cite{huang2020pixelBERT}   &arXiv-2020  &image-text &\makecell[c]{CNN, \\ Trans}    &MLM, MML   
&\makecell[l]{First to align vision \\ and language in pixel \\ and text-level} 
& 142M
&-   \\
\hline
16  &OSCAR~\cite{li2020oscar}   &ECCV-2020  &image-text &Trans   &CS, MLM   
&\makecell[l]{Align the visual patches \\ with word embeddings by using \\ object tags as anchor points} 
& 155M 
&\href{URL}{https://github.com/microsoft/Oscar}   \\ 
\hline
17  &pyramidCLIP~\cite{gaoPyramidCLIP2022}   &arXiv-2022     &image-text   &\makecell[c]{CNN+Trans}   &CS
&\makecell[l]{Hierarchical image-text \\ contrastive learning}
& - 
&-   \\
\hline 
18  &FashionBERT~\cite{gao2020fashionbert}   &RDIR-2020  &image-text   &BERT   &MLM, MOR, MML 
&\makecell[l]{Use image patches for fashion \\ domain instead of RoIs}   
& - 
&\href{URL}{https://github.com/alibaba/EasyTransfer}   \\ 
\hline
19  &VILLA~\cite{gan2020VILLA}   &NeurIPS-2020      &image-text   &Trans   &MLM, MOR, MML   
&\makecell[l]{Pre-training with \\ adversarial learning}  
& - 
&\href{URL}{https://github.com/zhegan27/VILLA}   \\ 
\hline
20  &ERNIE-ViL~\cite{yu2021ernieViL}   &AAAI-2021  &image-text &Trans   &\makecell[c]{MOC, AttP, RelP, \\ MLM, MOR, MML}   
&\makecell[l]{Use the knowledge obtained \\ from scene graph}   
& - 
&\href{URL}{https://github.com/Muennighoff/vilio}   \\ 
\hline
21  &KVL-BERT~\cite{song2021kvlbert}   &KBS-2021  &image-text   &BERT   &MOC, MLM   
&\makecell[l]{Integrate commonsense \\ knowledge for visual \\ commonsense reasoning} 
& - 
&-   \\
\hline
22  &VinVL~\cite{zhang2021vinvl}   &CVPR-2021 &image-text   &Trans   &\makecell[c]{MTL, \\ 3-way CS}   
&\makecell[l]{Verifying that visual feature \\ matters in VLP, i.e., strong \\ object detector brings \\ better results}   
& 157M 
&\href{URL}{https://github.com/pzzhang/VinVL}   \\
\hline
23  &VL-T5~\cite{cho2021VLT5}    &ICML-2021  &image-text   &Trans   &\makecell[c]{MLM, VQA, MML, \\ VG, GC}    
&\makecell[l]{Unified framework for VL \\ via generating texts}  
& 400M 
&\href{URL}{https://github.com/j-min/VL-T5}   \\
\hline
24  &ViLT~\cite{kim2021vilt}   &ICML-2021      &image-text &Trans   &MLM, MML   
&\makecell[l]{Use linear embedding only \\ for Fast VL transformer}
& 87M
&\href{URL}{https://github.com/dandelin/vilt}   \\
\hline
25  &ALIGN~\cite{jia2021ALIGN}  &ICML-2021      &image-text   &\makecell[c]{EfficientNet, \\ BERT}   &CS   
&\makecell[l]{Milestone for image-text \\ pre-training using noisy data}
& 300M 
&-   \\ 
\hline
26  &Kaleido-BERT~\cite{zhuge2021kaleidoBERT}   &CVPR-2021 &image-text  &Trans  &\makecell[c]{MLM, MML, \\ AKPM}   
&\makecell[l]{Use saliency detector to \\ generate multi-grained patches}   
& - 
&\href{URL}{http://dpfan.net/Kaleido-BERT}   \\ 
\hline
27  &MDETR~\cite{kamath2021mdetr}   &ICCV-2021  &image-text   &CNN+Trans &STP, MML    
&\makecell[l]{A text-modulated detection system \\ which can be trained \\ in an end to end way}  
& - 
&\href{URL}{https://github.com/ashkamath/mdetr}   \\ 
\hline
28  &SOHO~\cite{huang2021SOHO}  &CVPR-2021 &image-text   &CNN+Trans &MLM, MOR, MML
&\makecell[l]{Use a dynamic-updated \\ visual dictionary for \\ vision-language alignment}    
& - 
&\href{URL}{https://github.com/researchmm/soho}   \\ 
\hline
29  &E2E-VLP~\cite{xu2021E2EVLP}    &ACL-2021  &image-text   &Trans &OBD, ITG  
&\makecell[l]{The first PTM for \\ vision-language understanding \\ and generation}   
& 94M
&-   \\ 
\hline
30  &PIM~\cite{xue2021VisualParsing}   &NeurIPS-2021     &image-text   &Trans   &MLM, MML, MOR   
&\makecell[l]{Measure and reveal the \\ V+L fusion using the proposed \\ inter-modality flow metric}  
& 48M 
&-   \\ 
\hline
31  &$CLIP-ViL_{p}$~\cite{shen2021CLIPViL}   &arXiv-2021  &image-text   &Trans   &MLM, VQA, MML   
&\makecell[l]{Take the CLIP visual encoder \\  as its visual backbone}   
& - 
&\href{URL}{https://github.com/clip-vil/CLIP-ViL}   \\ 
\hline
32  &ALBEF~\cite{li2021ALBEF}    &NeurIPS-2021  &image-text &Trans  &CS, GR   
&\makecell[l]{Design a momentum model to \\ address noisy data}   
& 210M
&\href{URL}{https://github.com/salesforce/ALBEF}	\\  
\hline
33  &SimVLM~\cite{wang2021simvlm}   &arXiv-2021    &image-text   &Trans   &PrefixLM   
&\makecell[l]{Simple VL model using \\ single PrefixLM pre-training \\ objective only}  
& - 
&-   \\ 
\hline
34  &MURAL~\cite{jain2021mural}   &arXiv-2021  &image-text   &Trans &CS   
&\makecell[l]{Adopt multi-task contrastive \\ learning objective \\ (image-text, text-text)}   
& 430M
&-  \\ 
\hline
35  &VLMo~\cite{wang2021vlmo}   &arXiv-2021 &image-text   &Trans   &MLM, MML, CS 
&\makecell[l]{Jointly learns visual-, \\ text-encoder and a \\ fusion encoder}  
& - 
&\href{URL}{https://aka.ms/vlmo}   \\   
\hline
\end{tabular}} 
\end{table*}

\begin{table*}[htp]
\center
\tiny   
\caption{The summary of mainstream multi-modal pre-trained big models (Part-II).}   
\label{PTMList2}
\resizebox{\textwidth}{95mm}{
\begin{tabular}{c|l|l|c|c|c|l|c|cccccccc}
\hline
\textbf{No.} &\textbf{Model}  &\textbf{Pub.}   &\textbf{Modality} &\textbf{Architecture} &\textbf{Objective}  &\textbf{Highlights}  &\textbf{Params} &\textbf{Code}\\  
\hline  
36  &METER~\cite{dou2021METER}    &CVPR-2022   &image-text   &Trans   &\makecell[c]{MLM, MOR, \\ MOC, MML}  
&\makecell[l]{An empirical study on VLP}   
&-
&\href{URL}{https://github.com/zdou0830/METER}   \\  
\hline
37  &VideoBERT~\cite{sun2019videobert}    &ICCV-2019      &video-text   &BERT   &MLM
&\makecell[l]{A simple model for \\ video-text feature learning}  
&-
&\href{URL}{https://github.com/ammesatyajit/VideoBERT}   \\ 
\hline
38  &CBT~\cite{sun2019CBT}   &arXiv-2019    &video-text   &Trans   &NCE   
&\makecell[l]{Self-supervised contrastive \\ bidirectional Transformer}   
&15M
&-   \\ 
\hline
39  &UniVL~\cite{luo2020univl}   &arXiv-2020    &video-text   &Trans   &\makecell[c]{MLM, MFM, \\ MML, ITG}    
&\makecell[l]{A unified model for multimodal \\ understanding and generation}   
&-
&\href{URL}{https://github.com/microsoft/UniVL}   \\ 
\hline
40  &HERO~\cite{li2020hero}   &EMNLP-2020   &video-text   &Trans   &\makecell[c]{MLM, MFM, \\ VSM, FOM}  
&\makecell[l]{Hierarchical Transformer-based \\ model trained with newly \\ proposed VSM and FOM}   
&-
&\href{URL}{https://github.com/linjieli222/HERO}   \\  
\hline
41  &MMFT-BERT~\cite{urooj2020mmftBERT}   &EMNLP-2020    &image-text   &BERT   &\makecell[c]{Classification}  
&\makecell[l]{Adopt multiModal fusion \\ Transformer for modality fusion}
&-
&\href{URL}{https://github.com/aurooj/MMFT-BERT}   \\ 
\hline
42  &ActBERT~\cite{zhu2020actbert}   &CVPR-2020      &image-text   &Trans &CS, GR   
&\makecell[l]{Extract actions explicitly \\ as one of the inputs}   
&-
&-   \\
\hline
43  &CLIP~\cite{radford2021CLIP}   &ICML-2021      &image-text   &\makecell[c]{Resnet, \\ Trans} &CS   
&\makecell[l]{Milestone for image-text \\ pre-training using noisy data}  
&88.6M
&\href{URL}{https://github.com/OpenAI/CLIP}   \\  
\hline
44  &Frozen~\cite{bain2021frozen}   &ICCV-2021  &video/image-text   &Trans   &MML   
&\makecell[l]{Jointly optimize the model \\ on both images and videos}
&180.4M
&\href{URL}{https://github.com/m-bain/frozen-in-time}   \\ 
\hline
45  &RegionLearner~\cite{yan2021RegionLearner}   &arXiv-2021 &video-text   &Trans   &MML   
&\makecell[l]{Implicitly learning object region \\ without position supervision}  
&-
&\href{URL}{https://github.com/showlab/Region_Learner}   \\ 
\hline
46 &UNIMO~\cite{li2020unimo}   &arXiv-2020      &image-text   &Trans   &CS   
&\makecell[l]{Adapt to single-, multi-modal \\ understanding and generation \\ tasks effectively}   
&-
&\href{URL}{https://github.com/PaddlePaddle/Research/tree/master/NLP/UNIMO}   \\ 
\hline 
47 &DALL-E~\cite{ramesh2021DALLE}  &ICML-2021   &image-text   &Trans   &ELB   
&\makecell[l]{Achieve high quality image \\ generation without using any \\ of the training labels}   
&12B
&\href{URL}{https://github.com/openai/DALL-E}   \\ 
\hline
48 &BriVL~\cite{huo2021wenlan}  &arXiv-2021   &image-text   &Trans   &InfoNCE   
&\makecell[l]{The first Chinese \\ large-scale MM-PTMs}   
& 10B
&\href{URL}{https://github.com/chuhaojin/WenLan-api-document}  \\ 
\hline
49 &VLC~\cite{Gui2022VLC} 
&arXiv-2022  &image-text   &ViT   &\makecell[c]{MIM, MLM \\ ITM}
&\makecell[l]{Built on top of MAE that does \\ not require trained on ImageNet} 
& 87M
&\href{URL}{https://github.com/guilk/VLC}   \\ 
\hline 
50 &M6~\cite{lin2021m6}  &arXiv-2021  &image-text   &Trans   &\makecell[c]{LM}   
&\makecell[l]{The largest pretrained \\ model in Chinese}  
& 100B
&-   \\   
\hline
51 &CogView~\cite{ding2021cogview}  &NeurIPS-2021   &image-text   &Trans   &\makecell[c]{NLL}
&\makecell[l]{The first open-source \\ large text-to-image \\ transformer}   
& 4B
&\href{URL}{https://github.com/THUDM/CogView}   \\ 
\hline
52 &VATT~\cite{akbari2021vatt}  &NeurIPS-2021   &\makecell[c]{Video, Audio, \\ Text}   &Trans   &\makecell[c]{NCE, MIL-NCE}  
&\makecell[l]{Modality-specific or \\ Modality-agnostic triplet \\ modality pre-trained model}  
& 306.1M
&\href{URL}{https://github.com/google-research/google-research/tree/master/vatt}   \\ 
\hline
53 &OPT~\cite{liu2021opt}  &arXiv-2021     &\makecell[c]{image, Audio, \\ Text}   &Trans   &\makecell[c]{MLM, MVM, MoLM \\ MAM, DTR, DIR}   
&\makecell[l]{The first model pre-trained \\ using triplet modalities}  
&-
&-   \\ 
\hline
54 &Florence~\cite{yuan2021florence}  &arXiv-2021     &image-text   &CoSwin   &UniCL
&\makecell[l]{Multi-dimensional expansion \\ of representations}  
&893M
&-  \\ 
\hline
55 &ROSITA~\cite{cui2021rosita}  &MM-2021  &image-text   &Trans   &\makecell[c]{SKM, MLM, \\ MRM}  
&\makecell[l]{Fuse the intra-, cross-modality \\ knowledge, and SKM}   
&-
&-  \\ 
\hline
56 &VLCDoC~\cite{Bakkali2022VLCDoC} 
&arXiv-2022  &image-text   &Trans   &\makecell[c]{CS}  
&\makecell[l]{Contrastive Pre-Training \\ for document classification}  
&-
&-   \\
\hline 
57 &MVP~\cite{wei2022mvp} &arXiv-2022    &image-text   &ViT   &MIM
&\makecell[l]{Multimodality-guided visual \\ pre-training leads to \\ impressive gains} 
&-
&-   \\ 
\hline 
58 &GilBERT~\cite{hong2021gilbert} &IR-2021     &image-text   &BERT   &MLM, MOR   
&\makecell[l]{Considers both realistic \\ and synthetic data for VLP} 
&-
&-   \\ 
\hline 
59 &COTS~\cite{Lu2022COTS} &arXiv-2022     &image-text   &Trans   &\makecell[c]{CS, KLD, \\ MVLM}
&\makecell[l]{Token- and task-level interaction \\ are proposed to enhance \\ cross-modal interaction} 
&-
&-   \\ 
\hline 
60  &U-VisualBERT~\cite{li2021Uvisualbert} &NAACL-2021   &image-text  &Trans, BERT   &GR, MML   
&\makecell[l]{\emph{Unpaired} image-text \\ data for pre-training}
&-
&\href{URL}{https://github.com/uclanlp/visualbert}   \\
\hline 
61 &Flamingo~\cite{alayrac2022flamingo}  &arXiv-2022     &image-text   &NFNet   &CS   
&\makecell[l]{Pre-training on interleaved \\ visual and text data as input}  
&80B
&\href{URL}{https://github.com/lucidrains/flamingo-pytorch}   \\ 
\hline
62 &M3P~\cite{ni2021m3p} &CVPR-2021  &image-text   &BERT   &\makecell[c]{xMLM, MC-MLM, \\ MC-MRM}    
&\makecell[l]{Multitask, Multilingual, \\ Multimodal Pre-training}  
&-
&\href{URL}{https://github.com/microsoft/M3P}   \\   
\hline 
63 &BLIP~\cite{li2022BLIP} &arXiv-2022  &image-text  &BERT   &CS, MML, MLM   
&\makecell[l]{Propose the multimodal mixture \\ of encoder-decoder, and \\ captioning-filtering scheme}
&224M
&\href{URL}{https://github.com/salesforce/BLIP}   \\ 
\hline 
64 &NUWA~\cite{wu2021nvwa} &arXiv-2021  &image-text   &Trans   &T2I, T2V, V2V   
&\makecell[l]{A 3D transformer framework \\ can handle image, text, \\ and video, simultaneously}  
&809M
&\href{URL}{https://github.com/microsoft/NUWA}   \\ 
\hline 
65 &TCL~\cite{yang2022vision} &CVPR-2022  &image-text   &BERT   &\makecell[c]{CMA, IMC, LMI \\ ITM, MLM}   
&\makecell[l]{The first work considers \\ local structure information \\ for  multi-modality \\ representation learning}  
&123.7M
&\href{URL}{https://github.com/uta-smile/TCL}   \\ 
\hline 
66 &SCALE~\cite{weim5product} &CVPR-2022  &\makecell[c]{image, text, table \\ video, audio}   &BERT   &\makecell[c]{MRP, MLM, MEM \\ MFP, MFP, MAM}
&\makecell[l]{A unified model to \\ handle five modalities}   
&-
&\href{URL}{https://xiaodongsuper.github.io/M5Product_dataset/}   \\ 
\hline 
67 &Clinical-BERT~\cite{yan2022clinical} &AAAI-2022  &image-text   &BERT   &\makecell[c]{CD, MMM \\ MLM, IMM}    
&\makecell[l]{The first work to learn \\ domain knowledge during \\ pre-training for \\ the medical domain}  
&102M
&-   \\ 
\hline 
68 &RegionCLIP~\cite{zhong2021regionclip} &CVPR-2022  &image-text   &Trans   &\makecell[c]{Distillation loss, CS}   
&\makecell[l]{Learn region-level visual \\ representations based on CLIP}   
&-
&\href{URL}{https://github.com/microsoft/RegionCLIP}   \\ 
\hline 
69 &ProbES~\cite{liang2022visual} &ACL-2022  &image-text   &LSTM, ViLBERT   &Ranking loss 
&\makecell[l]{Prompt-based learning \\ for VLN based on CLIP}  
&-
&\href{URL}{https://github.com/liangcici/Probes-VLN}   \\ 
\hline 
70 &GLIP~\cite{li2021GLIP} 
&CVPR-2022  &image-text   &BERT   &CS
&\makecell[l]{Unifying the object detection \\ and grounding into a \\ unified framework}  
&394M
&\href{URL}{https://github.com/microsoft/GLIP}   \\ 
\hline 
\end{tabular} } 
\end{table*}

\begin{table*}[htp]
\center
\tiny   
\caption{The summary of mainstream multi-modal pre-trained big models (Part-III).}   
\label{PTMList3}
\resizebox{\textwidth}{70mm}{
\begin{tabular}{c|l|l|c|c|c|l|c|cccccccc}
\hline
\textbf{No.} &\textbf{Model}  &\textbf{Pub.}   &\textbf{Modality} &\textbf{Architecture} &\textbf{Objective}  &\textbf{Highlights} &\textbf{Parameters} &\textbf{Code}\\  
\hline  
71 &VLP-MABSA~\cite{ling2022vision} &ACL-2022  &image-text   &BERT   &\makecell[c]{MLM, AOE, MRM \\ AOG, MSP}   
&\makecell[l]{Task-specific VL-PTMs \\ for multimodal aspect-based \\ sentiment analysis}  &-   
&\href{URL}{https://github.com/NUSTM/VLP-MABSA}   \\ 
\hline 
72 &R2D2~\cite{Xie2022zeroR2D2} &arXiv-2022  &image-text   &ViT, BERT   &GCPR, FGR, MLM   
&\makecell[l]{A two-way distillation strategy \\ is proposed, i.e., target- and \\ feature-guided distillation}   
&-
&-   \\ 
\hline 
73 &DeCLIP~\cite{li2021DeCLIP} &ICLR-2022  &image-text   &ViT   &\makecell[c]{InfoNCE, SS \\ MVS, NNS}   
&\makecell[l]{Learn generic visual features \\ in a data efficient way}   
&276M &\href{URL}{https://github.com/Sense-GVT/DeCLIP}   \\ 
\hline 
74 &DeFILIP~\cite{cui2022CLIPbenchmark} &arXiv-2022  &image-text   &ViT, ResNet   &CS   
&\makecell[l]{A benchmark for CLIP \\ and its variants}  
&- &\href{URL}{https://github.com/Sense-GVT/DeCLIP}   \\ 
\hline 
75 &SLIP~\cite{mu2021slip} 
&arXiv-2021  &image-text   &ViT   &CS, InfoNCE
&\makecell[l]{Combine the self-supervised \\ learning and CLIP pre-training \\ in a multi-task framework}  
&38M &\href{URL}{https://github.com/facebookresearch/SLIP}   \\
\hline 
76 &FILIP~\cite{yao2021filip} 
&arXiv-2021  &image-text   &ViT   &CS
&\makecell[l]{Cross-modal interactive learning \\ for finer-level alignment}  
&- &-   \\ 
\hline 
77 &SemVLP~\cite{li2021semvlp}  
&arXiv-2021  &image-text   &Trans   &\makecell[c]{MLM, MOP, \\ ITM, QA}  
&\makecell[l]{Fuse the single- and \\ two-stream architectures}  
&2.1B &-   \\
\hline 
78 &CoCa~\cite{yu2022coca}  
&arXiv-2022  &image-text   &Trans   &\makecell[c]{CS, ITG}  
&\makecell[l]{Jointly pre-train image \\ text model with \\ contrastive loss and \\ captioning loss}  
&- &-   \\
\hline 
79 &HiVLP~\cite{Chen2022HiVLP} 
&arXiv-2022  &image-text   &Trans   &\makecell[c]{LRM, HRL, \\ VLM}  
&\makecell[l]{Accelerate image-text \\ retrieval via hierarchical \\ retrieval}  
&- &- \\ 
\hline 
80 &CLIP-Event~\cite{li2022clipevent}  
&CVPR-2022  &image-text   &Trans    &\makecell[c]{CS}  
&\makecell[l]{Consider event structural \\ knowledge and prompts in \\ the pre-training phase.}
&- &\href{URL}{https://github.com/limanling/clip-event}   \\
\hline 
81 &AudioCLIP~\cite{guzhov2022audioclip} 
&ICASSP-2022  &image-text-audio   &Trans    &\makecell[c]{CS}  
&\makecell[l]{Build a triplet modality \\ based PTMs like CLIP} 
&30M &\href{URL}{https://github.com/AndreyGuzhov/AudioCLIP}   \\
\hline 
82 &VL-BEiT~\cite{bao2022vlbeit} 
&arXiv-2022  &image-text  &Trans  &\makecell[c]{MLM, MIM, MVLM} 
&\makecell[l]{Share the Transformer \\ network on both monomodal- \\ and multimodal-data}  
&- &\href{URL}{https://github.com/microsoft/unilm} \\ 
\hline 
83 &MV-GPT~\cite{seo2022MVGPT} 
&arXiv-2022  &image-text  &BERT  &\makecell[c]{MLM, LG}  
&\makecell[l]{Pre-train both a multi-modal \\ video encoder and a sentence \\ decoder jointly.}  
&117M &- \\ 
\hline 
84 &MMKD~\cite{Fan2022MMKD} 
&arXiv-2022  &image-text  &BERT  &\makecell[c]{ITM}  
&\makecell[l]{Iteratively execute knowledge \\ discovery and model pre-training \\ for continuous learning}  
&- &- \\ 
\hline
85 &GLIPv2~\cite{Zhang2022GLIPv2} 
&arXiv-2022  &image-text  &Swin, BERT  &\makecell[c]{PGL, CS, MLM}  
&\makecell[l]{Serves both the localization \\ and understanding tasks.}  
&- &\href{URL}{https://github.com/microsoft/GLIP} \\ 
\hline 
86 &LIMoE~\cite{Basil2022LIMoE} 
&arXiv-2022  &image-text  &Trans  &\makecell[c]{CS}  
&\makecell[l]{multi-modal pre-training \\ with a sparse mixture \\ of experts model}  
&675M &- \\ 
\hline 
87 &VLMixer~\cite{wang2022vlmixer} 
&arXiv-2022  &image-text  &Trans  &\makecell[c]{MLM, CMCL, MTM}   
&\makecell[l]{Implicit cross-modal alignment \\ learning in unpaired VLP.}  
&- &\href{URL}{https://github.com/ttengwang/VLMixer} \\ 
\hline 
88 &ProtoCLIP~\cite{Chen2022ProtoCLIP} 
&arXiv-2022  &image-text  &Trans  &\makecell[c]{CS}   
&\makecell[l]{Combine the CLIP \\ loss and prototypical \\ supervisions for VLP.}  
&- &\href{URL}{https://github.com/megvii-research/protoclip} \\ 
\hline 
89 &i-Code~\cite{yang2022Icode}  
&arXiv-2022  &image-text-audio  &Trans  &\makecell[c]{MLM, MVM \\ MSM, CS}   
&\makecell[l]{It can handle different combinations \\ of modalities (such as single-, dual-, \\ and triple-modality) into \\ a single representation space.}  
&906M &- \\ 
\hline 
\end{tabular} } 
\end{table*}

\subsection{Pre-training using Knowledge}
\label{subsec_KEPTMs}
\begin{figure*}[!htp]
\centering
\includegraphics[width=\textwidth]{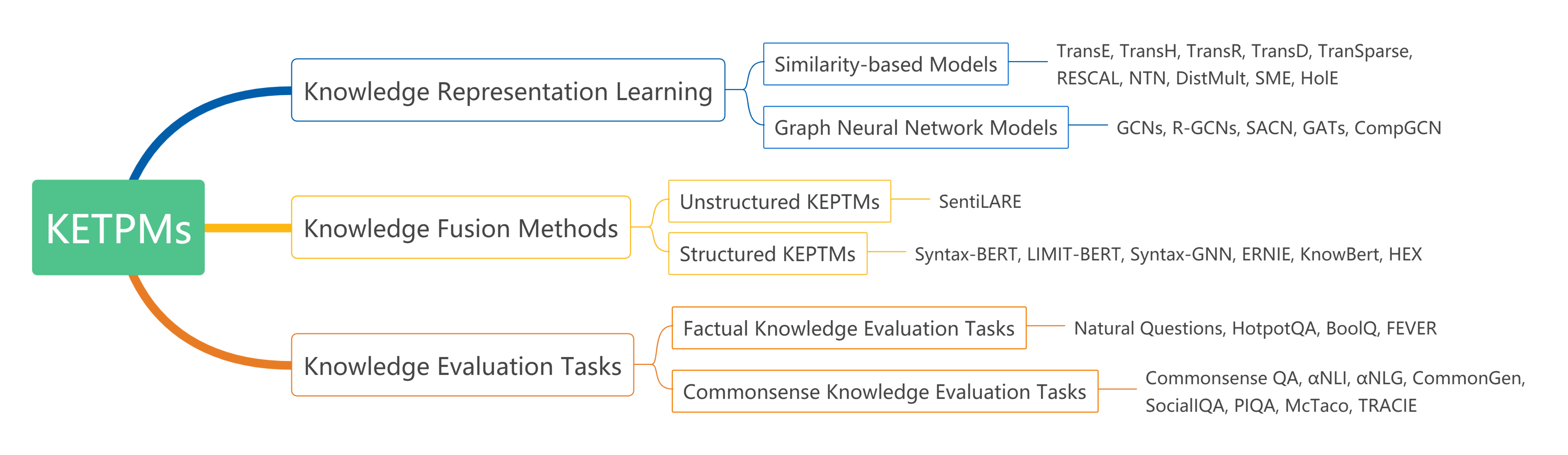}
\caption{The taxonomy of Knowledge Enhanced Pre-Trained Models (KEPTMs).}
\label{fig_KEPTMs}
\end{figure*}

Conventional pre-trained models suffer from poor logical reasoning and lack of interpretability. To alleviate those problems, it is straightforward to involve knowledge, deep understanding of data, in pre-training models, i.e., pre-training using knowledge also known as Knowledge Enhanced Pre-Trained Models (KEPTMs) shown in Fig.~\ref{fig_KEPTMs}.

\textbf{Knowledge Representation Learning} By learning to represent symbolic knowledge, usually in the form of entities and relations, knowledge representation learning enables neural network based models to fuse knowledge and improve their reasoning capabilities. Similarity-based models and graph neural network (GNN) models are two major methods of knowledge representation learning.

$\bullet$ \textbf{Similarity-based Models} Given similarity-based scoring functions, similarity-based models measure the similarity of latent semantics between two entities. Translation-based models are representatives of similarity-based models, as the distance in the vector space is often used to describe the similarity. TransE firstly models relations by translations, which operates on entity embeddings at low-dimension ~\cite{NIPS2013_1cecc7a7}. To deal with mapping properties of relations efficiently in complex models, such as reflexive, one-to-many, many-to-one and many-to-many, TransH is proposed to model a relation as a translation operation on a hyperplane ~\cite{Wang_Zhang_Feng_Chen_2014}. TransR is proposed to embed entity and relation in a separated spaces to capture different aspects of entities over various relations ~\cite{ji-etal-2015-knowledge}. Compared with TransR, not only the diversity of relations but also entities are considered in TransD ~\cite{10.5555/2886521.2886624}. To deal with heterogeneity and imbalance issues brought by knowledge graphs but ignored by aforementioned translation-based models, transfer matrices are replaced with adaptive sparse matrices in TranSparse, because the number of entities linked by relations determines sparse degrees ~\cite{Ji_Liu_He_Zhao_2016}. Besides translation-based models, tensor or matrix factorization approaches have also been proposed for multi-relational data by introducing scoring or ranking functions to measure how likely the semantic matching is correct. With the latent components, RESCAL is capable of collective learning and can provide an efficient algorithm of the factorization of a three-way tensor~\cite{10.5555/3104482.3104584}. NTN introduces an expressive neural tensor network for reasoning over relationships between two entities~\cite{NIPS2013_b337e84d}. DistMult presents a general framework for multi-relational learning and shows the effectiveness of a simple bilinear formulation~\cite{https://doi.org/10.48550/arxiv.1412.6575}. SME designs a new neural network architecture to encode multi-relational graphs or tensors into a flexible continuous vector space, so that multi-relational semantics can be learnt ~\cite{Bordes2014}. HolE is proposed to learn compositional vector space representations of entire knowledge graphs by employing holographic models of associative memory and circular correlation to create compositional representations~\cite{Nickel_Rosasco_Poggio_2016}.

$\bullet$ \textbf{Graph Neural Network Models} To further leverage the structure of the graph rather than collections of triplets, graph neural network models are employed to embed entities and relations. As convolutional neural networks (CNNs) are extremely efficient architectures in recognition tasks over different domains, they are generalized to graphs based on hierarchical clustering of the domain and the spectrum of the graph Laplacian in~\cite{ae482107de73461787258f805cf8f4ed}. Inspired by the pioneering work, further efforts have been done on graph convolutional networks (GCNs), such as semi-supervised classification~\cite{kipf2017semi}, unsupervised learning based on the variational auto-encoder (VAE)~\cite{https://doi.org/10.48550/arxiv.1611.07308}, inductive representation learning to sample and aggregate features from a node's local neighborhood~\cite{NIPS2017_5dd9db5e}, and attention mechanism by leveraging masked self-attentional layers~\cite{https://doi.org/10.48550/arxiv.1710.10903}. Beyond GCNs, R-GCNs is developed to deal with the highly multi-relation data characteristic of realistic knowledge bases~\cite{10.1007/978-3-319-93417-4_38}. A structure-aware convolutional network (SACN) takes the benefit of GCN and ConvE~\cite{Shang_Tang_Huang_Bi_He_Zhou_2019} together, where GCN as the encoder utilizes knowledge graph node structure and ConvE as the decoder enables the translational feature~\cite{Dettmers_Minervini_Stenetorp_Riedel_2018}. To further enhance Graph Attention Networks (GATs) and capture both entity and relation features within any entity's neighborhood, another model is proposed for attention-based feature embedding ~\cite{KBGAT2019}. To leverage various composition operations for embedding entities and relations in KGs and ever-increasing number of relations, a composition-based GCN named CompGCN is proposed to embed both nodes and relations jointly~\cite{vashishth2020compositionbased}.

\textbf{Knowledge Fusion Methods} How to fuse knowledge into pre-trained models and improve their logical understanding of data after knowledge representation learning remains a challenge to researchers. According to the category of knowledge provided, KEPTMs roughly contain two categories: unstructured knowledge and structured knowledge enhanced pre-trained models.

$\bullet$ \textbf{Unstructured KEPTMs}
Unstructured knowledge often refers to the knowledge without structures involved, which is in the form of plain text, like the words or phrases. Although some literatures introduce entities as supervised data and achieve promising performance, structural information is ignored while only entities are used to enable PTMs to learn semantics or attain extra key features from them. Word-aligned attention aligns the character-level attention to the word level to exploit explicit word information in Chinese  ~\cite{li-etal-2020-enhancing}. SentiLARE also introduces part-of-speech tag and sentiment polarity to build word-level linguistic knowledge ~\cite{ke-etal-2020-sentilare}. As unstructured text trained neural language models can store knowledge implicitly, PTMs can be further fine-tuned to explicitly retrieve knowledge without access to external knowledge or context ~\cite{roberts-etal-2020-much}.

$\bullet$ \textbf{Structured KEPTMs}
Contrary to unstructured KEPTMs, structured KEPTMs take account of sorts of structural information, including syntax-tree, rules and knowledge graphs. Syntax-BERT incorporates syntax trees effectively and efficiently into pre-trained Transformers~\cite{sachan-etal-2021-syntax}. LIMIT-BERT learns language representations across multiple linguistics tasks including constituent and dependency syntactic parsing~\cite{zhou-etal-2020-limit}. Syntax-GNN is proposed to learn syntax representations by using dependency trees and fusing the embeddings into transformers~\cite{sachan-etal-2021-syntax}. Knowledge graphs (KGs) provide structural knowledge in the form of entities and relations between them. An enhanced language representation model ERNIE is trained by utilizing both large-scale textual corpora and knowledge graphs, so that it can simultaneously leverage lexical, syntactic and knowledge ~\cite{zhang2019ernie}. Similar work named KnowBert is also proposed for large-scale models to embed multiple knowledge bases with entity linkers, which retrieves relevant entity embeddings and updates contextual word representations by the word-to-entity attention~\cite{peters-etal-2019-knowledge}. Moreover, the reasoning capability is also developed by finding supporting-facts, based on a large external knowledge base~\cite{ijcai2017-179,8046084}. Rules, in the form of constraints or even logical expressions, are preferred due to their interpretability and accountability. HEX graphs are proposed to enhance existing models by capturing semantic relations between labels applied to the same object~\cite{10.1007/978-3-319-10590-1_4}.

\textbf{Knowledge Evaluation Tasks} Besides conventional performance metrics, more knowledge-oriented tasks are required to evaluate the capability of KEPTMs and inspect whether external knowledge really helps models understand data semantically. Knowledge evaluation tasks are severed as testbeds to ensure the effectiveness of knowledge fusion methods. Currently, knowledge evaluation tasks mainly focus on NLP tasks and can be categorized into two groups based on the types of required knowledge: factual knowledge and commonsense knowledge evaluation tasks.

$\bullet$ \textbf{Factual Knowledge Evaluation Tasks}
Factual knowledge is the knowledge of facts, including specific details and elements to describe the objective facts~\cite{yang2021survey}. Factual knowledge evaluation tasks focus on testing models' reasoning ability on factual knowledge over various domains, like answering questions by giving a fact or judging the correctness of a given fact. Natural Questions is the first large publicly available dataset and robust metrics are also introduced to evaluate the performance of question answering (QA) systems~\cite{kwiatkowski-etal-2019-natural}. HotpotQA, another QA dataset, provides supporting facts at sentence-level for reasoning and new factoid comparison questions~\cite{yang-etal-2018-hotpotqa}. Different from the above two open-domain QA tasks, BoolQ only involves yes/no naturally occurring questions, namely verifying facts generated in unprompted and unconstrained settings, but those queries involve with complicated and non-factoid information so that make it unexpectedly challenging ~\cite{clark-etal-2019-boolq}. Another fact extraction and verification task FEVER is proposed and a new type of claims NotEnoughInfo is introduced beside Supported and Refuted~\cite{thorne-etal-2018-fever}. Entity linking, linking entities from a knowledge base to the corresponding textual mentions in a corpus, can also evaluate how well a model understands the factual knowledge~\cite{10.1145/2661829.2661887}.

$\bullet$ \textbf{Commonsense Knowledge Evaluation Tasks}
Commonsense knowledge refers to the information generally accepted by the majority of people concerning everyday life, i.e. the practical knowledge about how the world works~\cite{yin2022survey}. Like factual knowledge evaluation tasks, Commonsense QA also focuses on QA, but such QA requires prior knowledge outside the given document or context~\cite{talmor-etal-2019-commonsenseqa}. To extend the QA task Abductive Natural Language Inference ($\alpha$NLI), Abductive Natural Language Generation ($\alpha$NLG), a conditional generation task, is also proposed to explain given observations in natural language ~\cite{bhagavatula2020abductive}. CommonGen further explicitly tests models for the ability of generative commonsense reasoning due to its rigorous requirements on both relation reasoning and compositional generalization~\cite{lin-etal-2020-commongen}. Besides general commonsense evaluation tasks evaluating how well models understand daily scenarios, specific commonsense knowledge ones are further designed for different scenarios. SocialIQA, a large-scale benchmark for social commonsense reasoning, is challenging even for PTMs~\cite{sap-etal-2019-social}. Beside human interactions, physical interactions are also important in commonsense knowledge, hence the task of PIQA is introduced for physical commonsense reasoning~\cite{Bisk_Zellers_Lebras_Gao_Choi_2020}. Temporal commonsense is crucial for understanding the timing of events, for example duration, frequency, and order, leading to correct reasoning. McTaco defines five classes of temporal commonsense~\cite{zhou-etal-2019-going}, while TRACIE evaluates models' temporal understanding of implicit events~\cite{ZRNKSR21}.

\subsection{Characteristics of Different Pre-trained Big Models}
In the aforementioned paragraphs, we give a review to the main streams of multi-modal pre-trained models and highlight the features of each model in Table~\ref{PTMList1}, Table~\ref{PTMList2}, and Table~\ref{PTMList3}. In this subsection, we compare and analyze the characteristics of these models. 
Specifically, the early multi-modal pre-trained big models usually design an interactive learning module, for example, the ViLBERT~\cite{lu2019vilbert}, LXMERT~\cite{tan2019lxmert}. They integrate the co-attention or cross-attention mechanism into their framework to boost the feature representation between multiple inputs. Actually, these models obey the idea of interactive fusion of traditional small models. This allows for seamless integration with numerous downstream tasks and providing a high degree of flexibility. In contrast, many current big models directly process the inputs using projection layers and feed them into a unified network like the Transformers, including Unicoder-VL~\cite{li2020Unicodervl}, VideoBERT~\cite{sun2019videobert}, UniVL~\cite{luo2020univl}. More and more works demonstrate that the powerful Transformer network can achieve comparable or event better performance.

There are also some works make full use of existing big models and carry out secondary development to achieve a higher performance~\cite{zhong2021regionclip, guzhov2022audioclip}. To address the issues caused by shortage of paired multi-modal data, some researchers propose to training their model using unpaird data~\cite{li2021Uvisualbert}. These models show the great potential of processing massive multi-modal data. Unlike general big models, some models are specifically designed for a specific task or domain, like the e-commerce, or Indoor navigation. This provides conditions and convenience for fully mining more detailed domain knowledge assist the pre-training process.

\section{Downstream Tasks}  \label{downstreamTasks} 
\begin{figure}
\center
\includegraphics[width=3in]{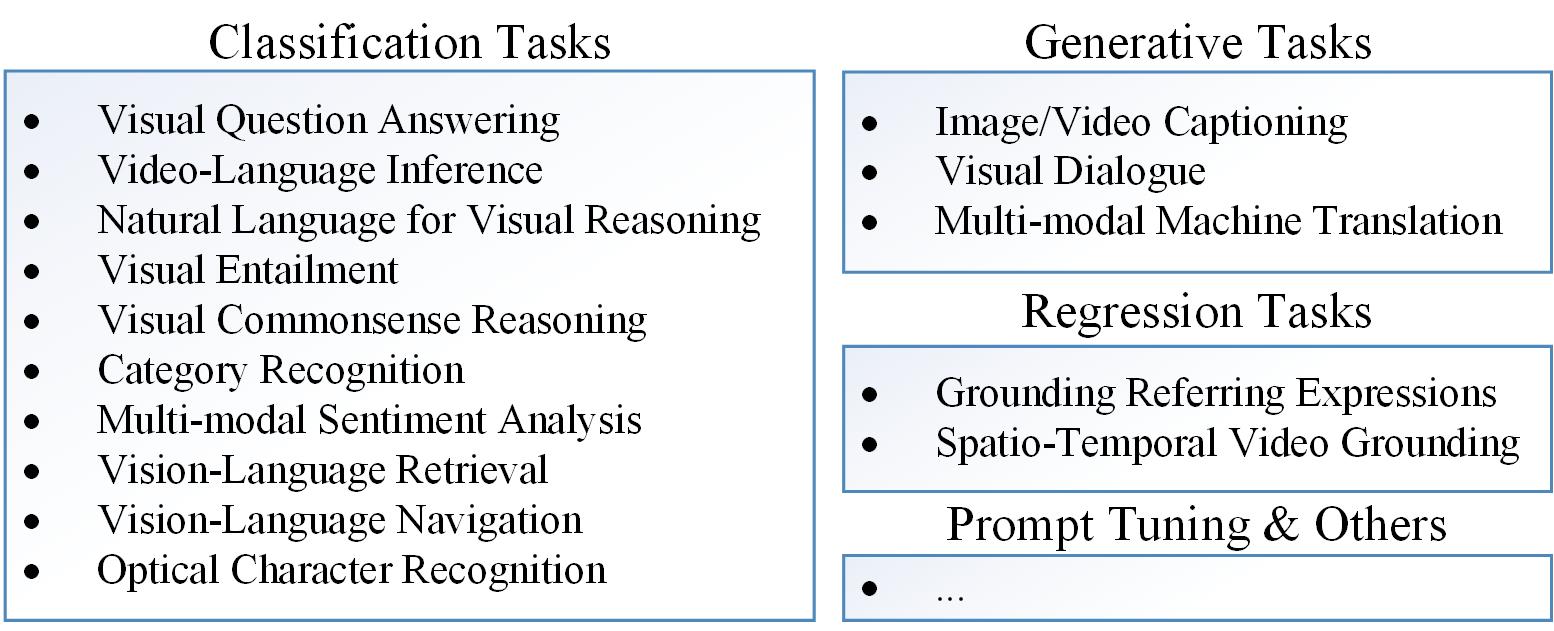}
\caption{An overview of downstream tasks reviewed in this paper.}  
\label{downstreamTASKs}
\end{figure}

After the pre-training phase, the researchers usually test their model on many downstream tasks to validate the powerful ability. Specifically, the generative tasks, classification tasks, regression tasks are adopted for the validation which will be discussed below. 
As a new learning paradigm, the prompt learning which target at modifying the downstream tasks to fit the pre-trained big model draws more and more attention. In this part, several representative prompt learning algorithms are also reviewed.
An overview of these downstream tasks are visualized in Fig.~\ref{downstreamTASKs}.  

\subsection{Generative Tasks} 

\noindent \textbf{Image/Video Captioning} attempt to describe content of input image or video using a couple of sentences. Usually, a visual encoder is used to encode the input image/video, then, a language decoder is adopted for sentence prediction in a word by word manner. NoCaps~\cite{agrawal2019nocaps} is proposed by Agrawal et al. in 2019. It is also an image captioning task but focus on developing generalized captioning models.

\noindent \textbf{Visual Dialogue (VD)} attempt to let the AI agent to talk with humans by holding a meaningful dialog about the visual content~\cite{das2017visualDialog}.

\noindent \textbf{Multi-modal Machine Translation (MMT)} is a task that targets translating the source sentence into a different language based on the paired image~\cite{yang2020visualMMMT}.

\subsection{Classification Tasks} 

\noindent \textbf{Visual Question Answering (VQA)} model is provided with an image and a question, and asked to produce an answer~\cite{antol2015vqa}. The relations between GQA~\cite{hudson2019gqa} and VQA is similar to the NoCaps and the standard captioning task. It is introduced to address key drawbacks of previous VQA datasets, and generate novel and diverse questions from a robust question engine, which sufficiently considers the content and structure.

\noindent \textbf{Video-Language Inference (VLI)} is proposed by Liu et al.~\cite{liu2020violin} in year 2020, which aims at understanding the video and text multimodal data.

\noindent \textbf{Natural Language for Visual Reasoning (NLVR)} can be seen as a binary classification problem. As noted in ~\cite{suhr2017corpus}, the model needs to judge the authenticity of a statement for the image.

\noindent \textbf{Visual Entailment (VE)}~\cite{xie2019visualentailment} is a triplet-label classification problem derived from Text Entailment (TE) task~\cite{dagan2005textentailment}. The VE model needs to predict whether the given image semantically entails the text. The three labels are \emph{entailment}, \emph{neutral} or \emph{contradiction}.

\noindent \textbf{Visual Commonsense Reasoning (VCR)}~\cite{zellers2019VCR} is a variation of VQA, which require a machine to provide a rationale justification and answer correctly for the given challenging problem.

\noindent \textbf{Category Recognition (CR)} is a classification problem which attempt to predict the category of given image. Many computer vision tasks are belong to this downstream task, such as pedestrian attribute recognition~\cite{wang2022pedestriansurvey}, action recognition~\cite{wang2021actionclip}.

\noindent \textbf{Multi-modal Sentiment Analysis (MSA)} is a multi-modal fusion task proposed for sentiment analysis~\cite{ghosal2018contextual}, which attempt to aggregate various homogeneous and/or heterogeneous modalities for more accurate reason. The modalities can be text, visual and acoustic, etc.

\noindent \textbf{Vision-Language Retrieval (VLR)} can be used in many applications, such as text-based person search~\cite{li2017personsearch}, or general object retrieval based on language~\cite{chen2021deepimgretievalsurvey}.

\noindent \textbf{Vision-Language Navigation (VLN)}~\cite{gu2022VLNavigationSurvey, park2022visualnavigation} is task that the agents learn to navigate in 3D indoor environments following the given natural language instruction. A benchmark for the popular VLN can be found at the following \href{leaderboard}{https://eval.ai/web/challenges/challenge-page/97/leaderboard/270}.

\noindent \textbf{Optical Character Recognition (OCR)} target at convert the images of Diverse text information into machine-encoded text. 
Usually, the OCR system contains both text detection and text recognition modules.

\subsection{Regression Tasks} 
\noindent \textbf{Grounding Referring Expressions (GRE)} takes the visual image and language description as input, and output the location of target object described by the language~\cite{zhang2018grounding, yang2019GRE, ding2022exploringgrounding}. Similar tasks defined on videos are termed \textbf{Spatio-Temporal Video Grounding (STVG)}~\cite{tang2021HCSTVG} or \textbf{Tracking by Natural Language}~\cite{wang2021TNL2K, wang2018describe, feng2021SNLT}.

\subsection{Prompt Learning}
To make full use of pre-trained big models, the prompt learning (also called prompt tuning) is proposed to re-formulate the downstream tasks to fit the objectives of pre-trained models, including CPT~\cite{yao2021cpt}, CPL~\cite{he2022cpl}. Also, some prompt tuning schemes are designed to fix the parameters of the large model and adjust the parameters as little as possible to achieve good results, such as the VPT~\cite{jia2022VPT}, CoOp~\cite{zhou2021learningprompt}, CoCoOp~\cite{zhou2022conditionalPrompt}. To be specific, the VPT~\cite{jia2022VPT} fixes the parameters of ViT models and integrates the prompt vectors as additional input. It achieves good performance even only tune the parameters of classification head and prompts. CoOp~\cite{zhou2021learningprompt} achieves huge improvements by tuning the context words into a set of learnable prompt vectors. Conditional Context Optimization (CoCoOp)~\cite{zhou2022conditionalPrompt} is developed based on CoOp which learns an external network to generate input-conditional tokens for each image. It addresses the issue of class shift significantly using such dynamic prompts.

\section{Experimental Analysis}  \label{experimentalAnalysis}
Considering the complexity and numbers of MM-PTMs, it is almost impossible to reproduce pre-training tasks in a short amount of time. Therefore, the experiments and related analyses of the pre-training are ignored in this paper. However, we still want to summarize a more complete review paper for the readers, thus, we extract the experimental results of the corresponding downstream tasks from their paper and compare them to the shared benchmark datasets. More detailed results can be found in Table~\ref{PTMList1} and Table~\ref{PTMList2}.

\begin{figure*}[!htp]
\center
\includegraphics[width=6in]{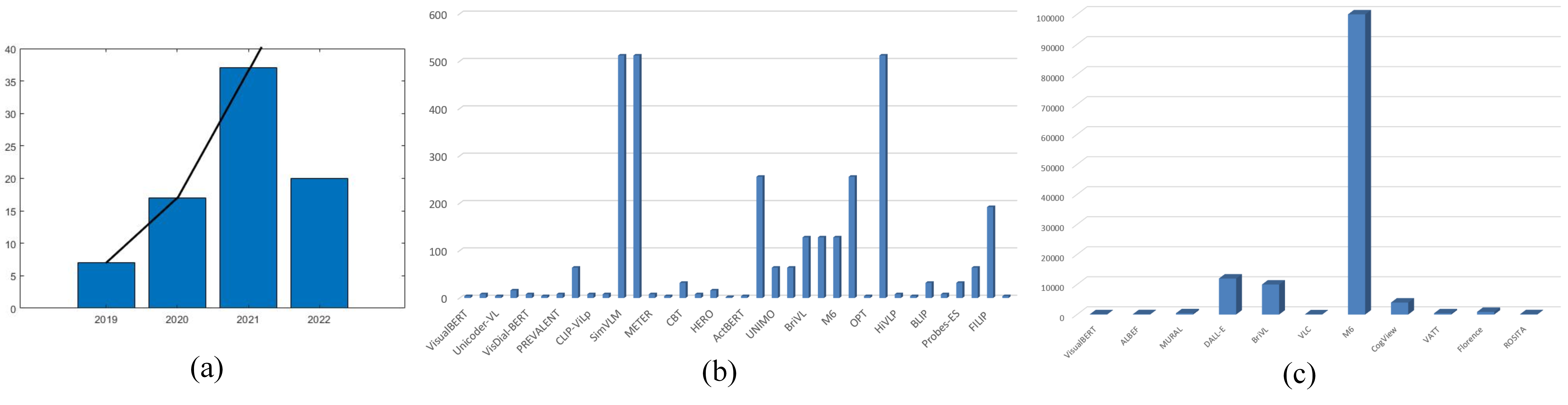}
\caption{(a). Number of MM-PTMs papers published from year 2019 to 2022; (b). Number of GPUs used for pre-training of selected models; (c). Parameters of selected MM-PTMs.}  
\label{modelsGPUsParmas}
\end{figure*}

\subsection{Model Parameters and Training Information}
As shown in Fig.~\ref{modelsGPUsParmas} (a), the large-scale MM-PTMs are emerging in the year 2019 and the number of papers shows an increasing trend year by year \footnote{Note that only half a year's results (the year 2022, from January to June) have been counted.}. 
From the Fig.~\ref{modelsGPUsParmas} (b), it is easy to find that current large-scale PTMs are optimized on servers with more than 8 GPUs. Also, many of them are trained using more than 100 GPUs, such as BriVL (128)~\cite{huo2021wenlan}, VLC (128)~\cite{Gui2022VLC}, M6 (128)~\cite{lin2021m6}, SimVLM (512)~\cite{wang2021simvlm}, MURAL (512)~\cite{jain2021mural}, CLIP (256)~\cite{li2021DeCLIP}, VATT (256)~\cite{akbari2021vatt}, Florence (512)~\cite{yuan2021florence}, FILIP (192)~\cite{yao2021filip}. Some MM-PTMs are trained on TPUs with massive chips, for example, the largest model of Flamingo~\cite{alayrac2022flamingo} is trained for 15 days on 1536 chips. From all these cases, we can see the huge demand of computing power for pre-trained big MM-PTMs.

Based on Fig.~\ref{modelsGPUsParmas} (c), it is also easy to find that many large-scale MM-PTMs are still with limited parameters, but some of them  indeed reached new heights. For example, the DALLE-E~\cite{ramesh2021DALLE} (12000 MB), BriVL~\cite{huo2021wenlan} (10000 MB), M6~\cite{lin2021m6} (100000 MB), and CogView~\cite{ding2021cogview} (4000 MB). The reasons for this phenomenon may be as follows: 
1). Many MM-PTMs are trained on several public datasets. The scale of parameters is greatly improved compared to traditional models, but not by a shocking amount. 
2). The development of big models is also limited by the need for large-scale computing power, and only a few giant companies or research institutes have such computing power platforms.

\begin{figure*}[!htp]
\center
\includegraphics[width=6in]{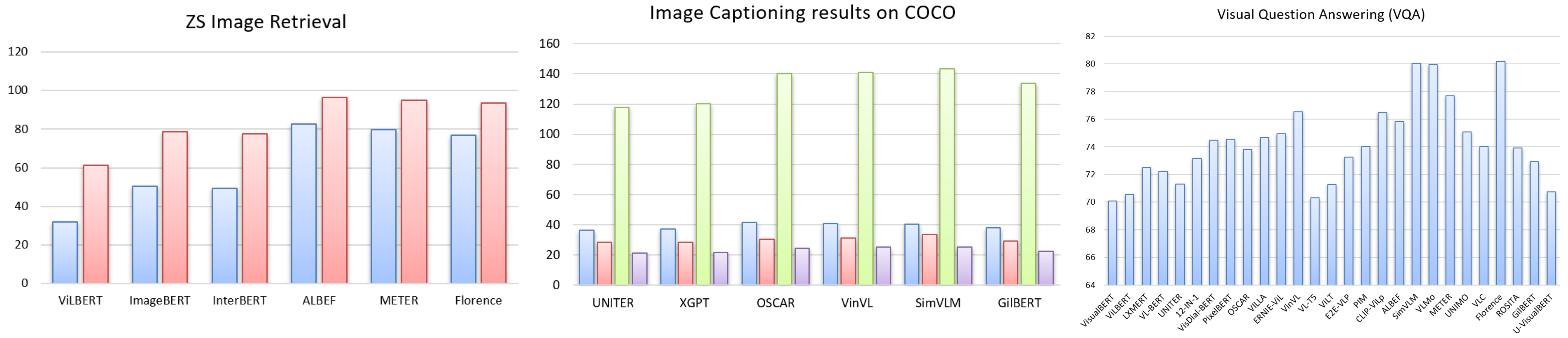}
\caption{Experimental results of selected MM-PTMs on zero-shot image retrieval (Rank-1, Rank-5), image captioning (BLEU, METEOR, CIDEr, SPICE), and visual question answering (Test-std).}  
\label{experimentResults}
\end{figure*}

\subsection{Performance on Representative Downstream Tasks} 
Here, we report the experimental results of zero-shot image retrieval, image captioning, and visual question answering. From Fig.~\ref{experimentResults} (a), we can find that the performance of different MM-PTMs have a big difference on the zero-shot image retrieval task. The blue and red vertical bar denotes the results of Rank-1 and Rank-5, respectively. Some models achieve high performance on this task which demonstrates the effectiveness of large-scale pre-training. For example, the ALBEF~\cite{li2021ALBEF} and METER~\cite{dou2021METER} achieves $82.80, 96.30$ and $79.60, 94.96$ on both evaluation metric.

For the image captioning task, we can find that the compared models achieved close performance on the COCO dataset according to Fig.~\ref{experimentResults} (b). Specifically, OSCAR~\cite{li2020oscar} obtains 41.7, 30.6, 140, 24.5; VinVL attains~\cite{zhang2021vinvl} 41, 31.1, 140.9, 25.2; SimVLM achieves~\cite{wang2021simvlm} 40.6, 33.7, 143.3, 25.4, respectively. These results are significantly better than traditional image captioning models pre-trained in a supervised manner through ImageNet~\cite{deng2009imagenet} classification task. Similar results can also be concluded from Fig.~\ref{experimentResults} (c).

\section{Research Directions} \label{researchDict}

Although the multi-modal pre-trained big models have obtained huge development, however, it is still a young research direction. Many problems and opportunities are still waiting for researchers to solve. In this section, we summarize several research points which are worthy to be tried.

\begin{figure}
\center
\includegraphics[width=3in]{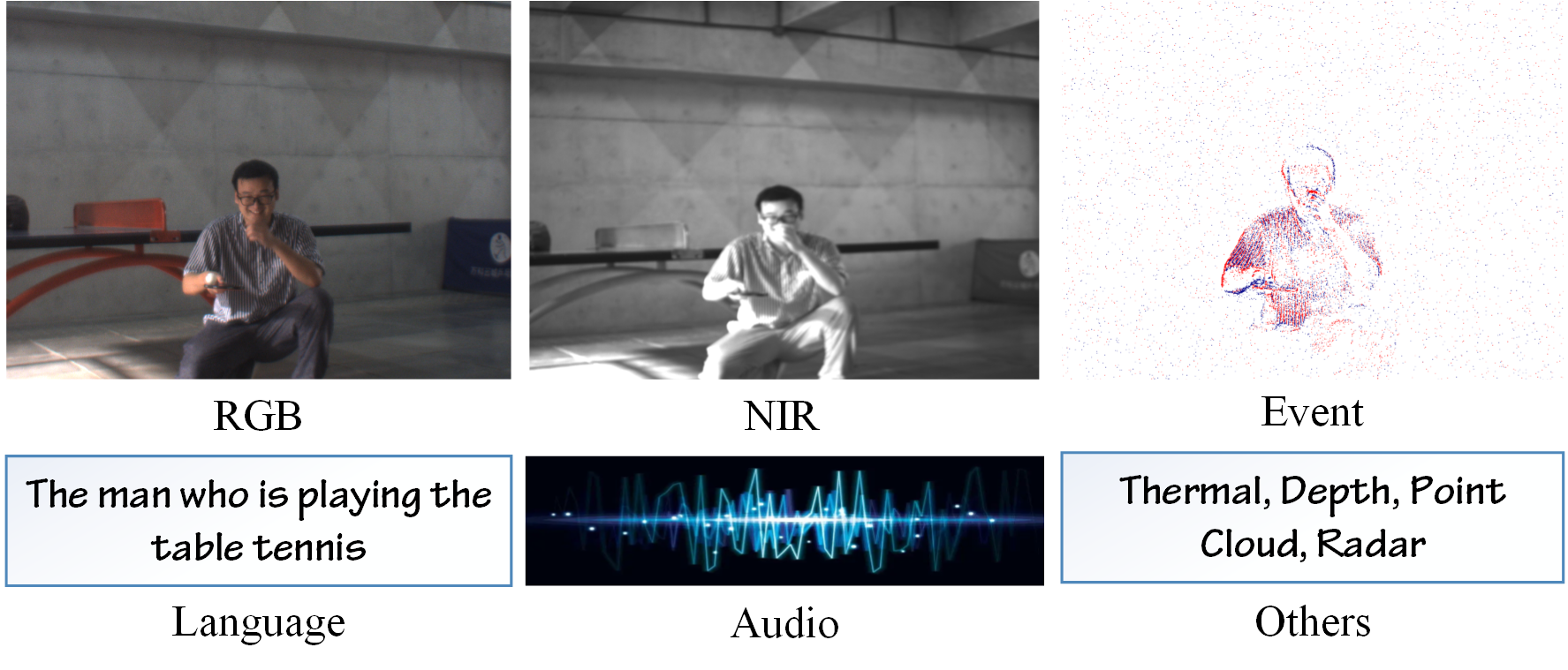}
\caption{Representative samples of mainstream modalities frequently used.}    
\label{more_modality}
\end{figure}

$\bullet$ \textbf{Pre-training on More Modalities}: 
Existing large-scale PTMs are usually pre-trained on two modalities, e.g., the vision and language. The missing of large amount aligned  multi-modal data may be a key reason. As an old saying goes, ``Sharpening your axe will not delay your job of chopping wood". The acquirement of real multi-modal data is the most important thing for large-scale pre-training, as shown in Fig.~\ref{more_modality}, such as visual image, text, audio, radar, event streams, depth image, thermal image, etc. To the best of our knowledge, no imaging device can capture so many modalities at the same time. Therefore, the manufacture of multi-modal imaging equipment can be a very significant thing. The pre-trained big model based on these data may have a wider potential for applications.

$\bullet$ \textbf{Incremental Learning based Pre-training}: 
Currently, existing pre-trained big methods are used for downstream tasks through feature finetuning or prompt learning~\cite{liu2021promptlearning}. This standard deep learning procedure works well in a short time, but pre-training is an expensive process. Specifically, the collection and cleaning of data, the electric charge used for pre-training, and the hardware device all cost a huge amount of human and material resources. When we gathered another group of data, the pre-training on the mixed data are expensive, redundant, and not environmentally friendly. However, seldom of them consider incremental learning for big models, and it is still unclear if the incremental learning algorithms developed for traditional deep learning work well for big models.  

In addition to the aforementioned data incremental learning, there are still many aspects that can be exploited for multi-modal pre-trained big modals. For example, the class (or category) incremental learning is a classical machine learning problem. Another interesting problem is modality-incremental learning, in another word, how to introduce and absorb the new modality into the already pre-trained multi-modal model. Because the new sensors (modalities) will appear at some indefinite time in the future, the designed multi-modal big models should be flexible enough to handle this situation.

$\bullet$ \textbf{Knowledge Enhanced Multi-Modal Pre-training}:   
Based on aforementioned reviews on MM-PTMs, we can find that the study of knowledge-assisted pre-training is still in the starting stage. Current works simply adopt external knowledge-graph or knowledge base in the pre-training phase, but they are usually single-modal, independent of multi-modal data, and limited to improving the understanding of data for models. Although commonsense knowledge is more ubiquitous, it is also abstract and introduces ambiguities, leading to challenges when applying to specific data. Therefore, we believe that further explorations on knowledge enhanced multi-modal pre-training are worth investigating. First, specified knowledge for multi-modal data is demanded to collect or extract through self-supervised learning. Second, more general knowledge fusion methods designed for multi-modal data are needed, beyond the limitations of vision and language modalities. Third, knowledge evaluation tasks specific for pre-training are required to inspect the enhancement of knowledge at this early stage, because pre-training is the first phase of the entire training procedure while downstream tasks are to be determined.

$\bullet$ \textbf{Fine-grained Multi-Modal Pre-training}:   
Most existing MM-PTMs are pre-trained from a global-view, for example, the researchers adopt the matching between the whole image and language as a supervised signal for the pre-training. The representative works are CLIP~\cite{radford2021CLIP}, ALIGN~\cite{jia2021ALIGN}, etc. Note that, the fine-grained local information mining or instance-level pre-training may further improve the overall performance of multi-modal pre-training. Some researchers have exploited the possibilities of fine-grained pre-training strategies~\cite{zhan2021product1m}. We hope more researchers can focus on this direction to further boost the final results.

$\bullet$ \textbf{Multi-Modal Pre-trained Model based Prompt Learning}:  
Current pre-trained big models are usually used in a ``pretrain-finetuning" way, specifically, the users need to initialize their model using pre-trained weights, then, finetune on downstream tasks. Although it works well in many tasks, however, the finetune maybe not be the most direct way. Because current multi-modal big models are pre-trained via modality matching, masked token prediction, and the downstream tasks are usually classification and regression tasks. Therefore, it exists a gap between multi-modal pre-training and finetuning. Recently, a new framework (termed prompt learning) is developed for big model based downstream tasks, which slickly transforms the setting of downstream tasks to make them consistent with pre-training~\cite{liu2021promptlearning}. Many works have demonstrated its effectiveness~\cite{yao2021cpt, zhou2022conditionalPrompt, zhou2021learningprompt, zhu2022promptUIC, li2022clipevent} in CV and NLP tasks. The research in this direction is also interesting and has great potential.

$\bullet$ \textbf{Migration of techniques developed for small-scale models}: The small-scale multi-modal models have been exploited for many years, and many representative models are proposed for deep multi-modal tasks~\cite{wang2015robust, wang2022MFGNet, lee2018stacked}. Among these works, diffusion, cross-attention, and dynamic neural networks are useful for specific multi-modal tasks. Part of these techniques is exploited in VL-PTMs, such as the cross-attention based ViLBERT~\cite{lu2019vilbert}. There are still many algorithms or tricks that have not yet been explored on large model tasks. We believe the transfer from small-scale to large-scale PTMs is worthy to be studied.

\textcolor{black}{$\bullet$ \textbf{Coupling and decoupling problems in cross-modal pre-training models}: The coupling involves establishing the correlation between different modalities and the ``cross" can be only realized through such correlation. The decoupling can further expand the modality dynamically. It is worth studying how to give feasible solutions to the two problems from the aspect of framework design.  }

\section{Conclusion}  
We give a comprehensive review of large-scale Multi-Modal Pre-Trained Models (MM-PTMs) in this paper. Firstly, we introduce the background of MM-PTMs, with a focus on conventional deep learning, and pre-training in NLP, CV, and speech. Then, the task definition, key challenges, and benefits of MM-PTMs are discussed. After that, we dive into the reviews of MM-PTMs and discuss the pre-training data, objectives, networks, knowledge enhanced pre-training, etc. We review the downstream tasks including generative, classification, and regression tasks, and also give an overview of model parameters of MM-PTMs and hardware for the pre-training. Experimental results of several representative tasks are also discussed and visualized. Finally, we point out some research directions that are worth to be focused on. We summarize this paper and hope our survey can provide some useful insights for the MM-PTMs.

\section*{Acknowledgement} 
This work is supported by Key-Area Research and Development Program of Guangdong Province (No. 2021B0101400002), Peng Cheng Laboratory Key Research Project (No. PCL2021A07), Multi-source Cross-platform Video Analysis and Understanding for Intelligent Perception in Smart City (No. U20B2052), National Natural Science Foundation of China (No. 61872256, 62102205).

\section*{Publish Information} 
\textbf{Name of the Journal}: Machine Intelligence Research \\ 
\textbf{Links}: \url{https://link.springer.com/article/10.1007/s11633-022-1410-8} \\ 
\textbf{DOI}: \url{10.1007/s11633-022-1410-8}


{\scriptsize
\bibliographystyle{unsrt}
\bibliography{reference}

\begin{thebibliography}{100}

\bibitem{krizhevsky2012AlexNet}
Alex Krizhevsky, Ilya Sutskever, and Geoffrey~E Hinton.
\newblock Imagenet classification with deep convolutional neural networks.
\newblock {\em Advances in neural information processing systems}, 25, 2012.

\bibitem{deng2009imagenet}
Jia Deng, Wei Dong, Richard Socher, Li-Jia Li, Kai Li, and Li~Fei-Fei.
\newblock Imagenet: A large-scale hierarchical image database.
\newblock In {\em 2009 IEEE conference on computer vision and pattern
  recognition}, pages 248--255. Ieee, 2009.

\bibitem{simonyan2014VGG}
Karen Simonyan and Andrew Zisserman.
\newblock Very deep convolutional networks for large-scale image recognition.
\newblock {\em arXiv preprint arXiv:1409.1556}, 2014.

\bibitem{he2016resnet}
Kaiming He, Xiangyu Zhang, Shaoqing Ren, and Jian Sun.
\newblock Deep residual learning for image recognition.
\newblock In {\em Proceedings of the IEEE conference on computer vision and
  pattern recognition}, pages 770--778, 2016.

\bibitem{szegedy2017inception}
Christian Szegedy, Sergey Ioffe, Vincent Vanhoucke, and Alexander~A Alemi.
\newblock Inception-v4, inception-resnet and the impact of residual connections
  on learning.
\newblock In {\em Thirty-first AAAI conference on artificial intelligence},
  2017.

\bibitem{hochreiter1997long}
Sepp Hochreiter and J{\"u}rgen Schmidhuber.
\newblock Long short-term memory.
\newblock {\em Neural computation}, 9(8):1735--1780, 1997.

\bibitem{pennington2014glove}
Jeffrey Pennington, Richard Socher, and Christopher~D Manning.
\newblock Glove: Global vectors for word representation.
\newblock In {\em Proceedings of the 2014 conference on empirical methods in
  natural language processing (EMNLP)}, pages 1532--1543, 2014.

\bibitem{kiros2015skipthoughtvectors}
Ryan Kiros, Yukun Zhu, Russ~R Salakhutdinov, Richard Zemel, Raquel Urtasun,
  Antonio Torralba, and Sanja Fidler.
\newblock Skip-thought vectors.
\newblock {\em Advances in neural information processing systems}, 28, 2015.

\bibitem{vaswani2017attention}
Ashish Vaswani, Noam Shazeer, Niki Parmar, Jakob Uszkoreit, Llion Jones,
  Aidan~N Gomez, {\L}ukasz Kaiser, and Illia Polosukhin.
\newblock Attention is all you need.
\newblock {\em Advances in neural information processing systems}, 30, 2017.

\bibitem{kenton2019bert}
Jacob Devlin Ming-Wei~Chang Kenton and Lee~Kristina Toutanova.
\newblock Bert: Pre-training of deep bidirectional transformers for language
  understanding.
\newblock In {\em Proceedings of NAACL-HLT}, pages 4171--4186, 2019.

\bibitem{xia2021xgpt}
Qiaolin Xia, Haoyang Huang, Nan Duan, Dongdong Zhang, Lei Ji, Zhifang Sui,
  Edward Cui, Taroon Bharti, and Ming Zhou.
\newblock Xgpt: Cross-modal generative pre-training for image captioning.
\newblock In {\em CCF International Conference on Natural Language Processing
  and Chinese Computing}, pages 786--797. Springer, 2021.

\bibitem{brown2020language}
Tom Brown, Benjamin Mann, Nick Ryder, Melanie Subbiah, Jared~D Kaplan, Prafulla
  Dhariwal, Arvind Neelakantan, Pranav Shyam, Girish Sastry, Amanda Askell,
  et~al.
\newblock Language models are few-shot learners.
\newblock {\em Advances in neural information processing systems},
  33:1877--1901, 2020.

\bibitem{raffel2020T5}
Colin Raffel, Noam Shazeer, Adam Roberts, Katherine Lee, Sharan Narang, Michael
  Matena, Yanqi Zhou, Wei Li, and Peter~J Liu.
\newblock Exploring the limits of transfer learning with a unified text-to-text
  transformer.
\newblock {\em Journal of Machine Learning Research}, 21:1--67, 2020.

\bibitem{yang2019xlnet}
Zhilin Yang, Zihang Dai, Yiming Yang, Jaime Carbonell, Russ~R Salakhutdinov,
  and Quoc~V Le.
\newblock Xlnet: Generalized autoregressive pretraining for language
  understanding.
\newblock {\em Advances in neural information processing systems}, 32, 2019.

\bibitem{dosovitskiy2020ViT}
Alexey Dosovitskiy, Lucas Beyer, Alexander Kolesnikov, Dirk Weissenborn,
  Xiaohua Zhai, Thomas Unterthiner, Mostafa Dehghani, Matthias Minderer, Georg
  Heigold, Sylvain Gelly, et~al.
\newblock An image is worth 16x16 words: Transformers for image recognition at
  scale.
\newblock In {\em International Conference on Learning Representations}, 2020.

\bibitem{liu2021swinTransformer}
Ze~Liu, Yutong Lin, Yue Cao, Han Hu, Yixuan Wei, Zheng Zhang, Stephen Lin, and
  Baining Guo.
\newblock Swin transformer: Hierarchical vision transformer using shifted
  windows.
\newblock In {\em Proceedings of the IEEE/CVF International Conference on
  Computer Vision}, pages 10012--10022, 2021.

\bibitem{li2020oscar}
Xiujun Li, Xi~Yin, Chunyuan Li, Pengchuan Zhang, Xiaowei Hu, Lei Zhang, Lijuan
  Wang, Houdong Hu, Li~Dong, Furu Wei, et~al.
\newblock Oscar: Object-semantics aligned pre-training for vision-language
  tasks.
\newblock In {\em European Conference on Computer Vision}, pages 121--137.
  Springer, 2020.

\bibitem{chen2020uniter}
Yen-Chun Chen, Linjie Li, Licheng Yu, Ahmed El~Kholy, Faisal Ahmed, Zhe Gan,
  Yu~Cheng, and Jingjing Liu.
\newblock Uniter: Universal image-text representation learning.
\newblock In {\em European conference on computer vision}, pages 104--120.
  Springer, 2020.

\bibitem{li2021DeCLIP}
Yangguang Li, Feng Liang, Lichen Zhao, Yufeng Cui, Wanli Ouyang, Jing Shao,
  Fengwei Yu, and Junjie Yan.
\newblock Supervision exists everywhere: A data efficient contrastive
  language-image pre-training paradigm.
\newblock {\em arXiv preprint arXiv:2110.05208}, 2021.

\bibitem{huang2020pixelBERT}
Zhicheng Huang, Zhaoyang Zeng, Bei Liu, Dongmei Fu, and Jianlong Fu.
\newblock Pixel-bert: Aligning image pixels with text by deep multi-modal
  transformers.
\newblock {\em arXiv preprint arXiv:2004.00849}, 2020.

\bibitem{jia2021ALIGN}
Chao Jia, Yinfei Yang, Ye~Xia, Yi-Ting Chen, Zarana Parekh, Hieu Pham, Quoc Le,
  Yun-Hsuan Sung, Zhen Li, and Tom Duerig.
\newblock Scaling up visual and vision-language representation learning with
  noisy text supervision.
\newblock In {\em International Conference on Machine Learning}, pages
  4904--4916. PMLR, 2021.

\bibitem{liu2021opt}
Jing Liu, Xinxin Zhu, Fei Liu, Longteng Guo, Zijia Zhao, Mingzhen Sun, Weining
  Wang, Hanqing Lu, Shiyu Zhou, Jiajun Zhang, et~al.
\newblock Opt: Omni-perception pre-trainer for cross-modal understanding and
  generation.
\newblock {\em arXiv preprint arXiv:2107.00249}, 2021.

\bibitem{cheng2022hybridDistillation}
De~Cheng, Jingyu Zhou, Nannan Wang, and Xinbo Gao.
\newblock Hybrid dynamic contrast and probability distillation for unsupervised
  person re-id.
\newblock {\em IEEE Transactions on Image Processing}, 31:3334--3346, 2022.

\bibitem{chen2022vlp}
Feilong Chen, Duzhen Zhang, Minglun Han, Xiuyi Chen, Jing Shi, Shuang Xu, and
  Bo~Xu.
\newblock Vlp: A survey on vision-language pre-training.
\newblock {\em arXiv preprint arXiv:2202.09061}, 2022.

\bibitem{du2022VLPsurvey}
Yifan Du, Zikang Liu, Junyi Li, and Wayne~Xin Zhao.
\newblock A survey of vision-language pre-trained models.
\newblock {\em arXiv preprint arXiv:2202.10936}, 2022.

\bibitem{zaib2020shortsurvey}
Munazza Zaib, Quan~Z Sheng, and Wei Emma~Zhang.
\newblock A short survey of pre-trained language models for conversational ai-a
  new age in nlp.
\newblock In {\em Proceedings of the Australasian Computer Science Week
  Multiconference}, pages 1--4, 2020.

\bibitem{zhangHanqing2022survey}
Hanqing Zhang, Haolin Song, Shaoyu Li, Ming Zhou, and Dawei Song.
\newblock A survey of controllable text generation using transformer-based
  pre-trained language models.
\newblock {\em arXiv preprint arXiv:2201.05337}, 2022.

\bibitem{yang2021survey}
Jian Yang, Gang Xiao, Yulong Shen, Wei Jiang, Xinyu Hu, Ying Zhang, and Jinghui
  Peng.
\newblock A survey of knowledge enhanced pre-trained models.
\newblock {\em arXiv preprint arXiv:2110.00269}, 2021.

\bibitem{yin2022survey}
Da~Yin, Li~Dong, Hao Cheng, Xiaodong Liu, Kai-Wei Chang, Furu Wei, and Jianfeng
  Gao.
\newblock A survey of knowledge-intensive nlp with pre-trained language models.
\newblock {\em arXiv preprint arXiv:2202.08772}, 2022.

\bibitem{bhargava2022CKRSurvey}
Prajjwal Bhargava and Vincent Ng.
\newblock Commonsense knowledge reasoning and generation with pre-trained
  language models: A survey.
\newblock {\em arXiv preprint arXiv:2201.12438}, 2022.

\bibitem{liu2020contextualembedsurvey}
Qi~Liu, Matt~J Kusner, and Phil Blunsom.
\newblock A survey on contextual embeddings.
\newblock {\em arXiv preprint arXiv:2003.07278}, 2020.

\bibitem{liu2021prepromptpredsurvey}
Pengfei Liu, Weizhe Yuan, Jinlan Fu, Zhengbao Jiang, Hiroaki Hayashi, and
  Graham Neubig.
\newblock Pre-train, prompt, and predict: A systematic survey of prompting
  methods in natural language processing.
\newblock {\em arXiv preprint arXiv:2107.13586}, 2021.

\bibitem{wang2021pretrainBiomedicalSurvey}
Benyou Wang, Qianqian Xie, Jiahuan Pei, Prayag Tiwari, Zhao Li, et~al.
\newblock Pre-trained language models in biomedical domain: A systematic
  survey.
\newblock {\em arXiv preprint arXiv:2110.05006}, 2021.

\bibitem{qiu2020pretrainNLPsurvey}
Xipeng Qiu, Tianxiang Sun, Yige Xu, Yunfan Shao, Ning Dai, and Xuanjing Huang.
\newblock Pre-trained models for natural language processing: A survey.
\newblock {\em Science China Technological Sciences}, 63(10):1872--1897, 2020.

\bibitem{han2021pretrainSurveys}
Xu~Han, Zhengyan Zhang, Ning Ding, Yuxian Gu, Xiao Liu, Yuqi Huo, Jiezhong Qiu,
  Yuan Yao, Ao~Zhang, Liang Zhang, et~al.
\newblock Pre-trained models: Past, present and future.
\newblock {\em AI Open}, 2:225--250, 2021.

\bibitem{du2022surveyVLP}
Yifan Du, Zikang Liu, Junyi Li, and Wayne~Xin Zhao.
\newblock A survey of vision-language pre-trained models.
\newblock {\em arXiv preprint arXiv:2202.10936}, 2022.

\bibitem{ruan2022TransformerVideosurvey}
Ludan Ruan and Qin Jin.
\newblock Survey: Transformer based video-language pre-training.
\newblock {\em AI Open}, 2022.

\bibitem{li2022VLISurvey}
Feng Li, Hao Zhang, Yi-Fan Zhang, Shilong Liu, Jian Guo, Lionel~M Ni, PengChuan
  Zhang, and Lei Zhang.
\newblock Vision-language intelligence: Tasks, representation learning, and
  large models.
\newblock {\em arXiv preprint arXiv:2203.01922}, 2022.

\bibitem{han2022transformersurvey}
Kai Han, Yunhe Wang, Hanting Chen, Xinghao Chen, Jianyuan Guo, Zhenhua Liu,
  Yehui Tang, An~Xiao, Chunjing Xu, Yixing Xu, et~al.
\newblock A survey on vision transformer.
\newblock {\em IEEE Transactions on Pattern Analysis and Machine Intelligence},
  2022.

\bibitem{khan2021transformerssurvey}
Salman Khan, Muzammal Naseer, Munawar Hayat, Syed~Waqas Zamir, Fahad~Shahbaz
  Khan, and Mubarak Shah.
\newblock Transformers in vision: A survey.
\newblock {\em ACM Computing Surveys (CSUR)}, 2021.

\bibitem{liu2021surveyTransformers}
Yang Liu, Yao Zhang, Yixin Wang, Feng Hou, Jin Yuan, Jiang Tian, Yang Zhang,
  Zhongchao Shi, Jianping Fan, and Zhiqiang He.
\newblock A survey of visual transformers.
\newblock {\em arXiv preprint arXiv:2111.06091}, 2021.

\bibitem{selva2022videoSurvey}
Javier Selva, Anders~S Johansen, Sergio Escalera, Kamal Nasrollahi, Thomas~B
  Moeslund, and Albert Clap{\'e}s.
\newblock Video transformers: A survey.
\newblock {\em arXiv preprint arXiv:2201.05991}, 2022.

\bibitem{guo2022threats}
Shangwei Guo, Chunlong Xie, Jiwei Li, Lingjuan Lyu, and Tianwei Zhang.
\newblock Threats to pre-trained language models: Survey and taxonomy.
\newblock {\em arXiv preprint arXiv:2202.06862}, 2022.

\bibitem{garrido2021survey}
Ismael Garrido-Mu{\~n}oz, Arturo Montejo-R{\'a}ez, Fernando
  Mart{\'\i}nez-Santiago, and L~Alfonso Ure{\~n}a-L{\'o}pez.
\newblock A survey on bias in deep nlp.
\newblock {\em Applied Sciences}, 11(7):3184, 2021.

\bibitem{meade2021empiricalSurvey}
Nicholas Meade, Elinor Poole-Dayan, and Siva Reddy.
\newblock An empirical survey of the effectiveness of debiasing techniques for
  pre-trained language models.
\newblock {\em arXiv preprint arXiv:2110.08527}, 2021.

\bibitem{kaliyar2020BERTsurvey}
Rohit~Kumar Kaliyar.
\newblock A multi-layer bidirectional transformer encoder for pre-trained word
  embedding: A survey of bert.
\newblock In {\em 2020 10th International Conference on Cloud Computing, Data
  Science \& Engineering (Confluence)}, pages 336--340. IEEE, 2020.

\bibitem{pengJiajia2021survey}
Jiajia Peng and Kaixu Han.
\newblock Survey of pre-trained models for natural language processing.
\newblock In {\em 2021 International Conference on Electronic Communications,
  Internet of Things and Big Data (ICEIB)}, pages 277--280. IEEE, 2021.

\bibitem{yuan2022roadmapBIGModels}
Sha Yuan, Hanyu Zhao, Shuai Zhao, Jiahong Leng, Yangxiao Liang, Xiaozhi Wang,
  Jifan Yu, Xin Lv, Zhou Shao, Jiaao He, et~al.
\newblock A roadmap for big model.
\newblock {\em arXiv preprint arXiv:2203.14101}, 2022.

\bibitem{Long2022VLPTsurvey}
Soyeon Caren~Han Siqu~Long, Feiqi~Cao and Haiqing Yang.
\newblock Vision-and-language pretrained models: A survey.
\newblock In {\em IJCAI}, 2022.

\bibitem{Xu2021MMTransSurvey}
Xu~Peng, Zhu Xiatian, and A.~Clifton David.
\newblock Multimodal learning with transformers: A survey.
\newblock {\em arXiv preprint arXiv:2206.06488}, 2022.

\bibitem{lecun1998LeNet}
Y.~Lecun, L.~Bottou, Y.~Bengio, and P.~Haffner.
\newblock Gradient-based learning applied to document recognition.
\newblock {\em Proceedings of the IEEE}, 86(11):2278--2324, 1998.

\bibitem{Huang_2017CVPR_densenet}
Gao Huang, Zhuang Liu, Laurens van~der Maaten, and Kilian~Q. Weinberger.
\newblock Densely connected convolutional networks.
\newblock In {\em Proceedings of the IEEE Conference on Computer Vision and
  Pattern Recognition (CVPR)}, July 2017.

\bibitem{qiu2020pre}
Xipeng Qiu, Tianxiang Sun, Yige Xu, Yunfan Shao, Ning Dai, and Xuanjing Huang.
\newblock Pre-trained models for natural language processing: A survey.
\newblock {\em Science China Technological Sciences}, 63(10):1872--1897, 2020.

\bibitem{zaib2020short}
Munazza Zaib, Quan~Z Sheng, and Wei Emma~Zhang.
\newblock A short survey of pre-trained language models for conversational ai-a
  new age in nlp.
\newblock In {\em Proceedings of the Australasian Computer Science Week
  Multiconference}, pages 1--4, 2020.

\bibitem{min2021recent}
Bonan Min, Hayley Ross, Elior Sulem, Amir Pouran~Ben Veyseh, Thien~Huu Nguyen,
  Oscar Sainz, Eneko Agirre, Ilana Heinz, and Dan Roth.
\newblock Recent advances in natural language processing via large pre-trained
  language models: A survey.
\newblock {\em arXiv preprint arXiv:2111.01243}, 2021.

\bibitem{liu2021pre}
Pengfei Liu, Weizhe Yuan, Jinlan Fu, Zhengbao Jiang, Hiroaki Hayashi, and
  Graham Neubig.
\newblock Pre-train, prompt, and predict: A systematic survey of prompting
  methods in natural language processing.
\newblock {\em arXiv preprint arXiv:2107.13586}, 2021.

\bibitem{wang2019glue}
Alex Wang, Amanpreet Singh, Julian Michael, Felix Hill, Omer Levy, and Samuel~R
  Bowman.
\newblock Glue: A multi-task benchmark and analysis platform for natural
  language understanding.
\newblock In {\em 7th International Conference on Learning Representations,
  ICLR 2019}, 2019.

\bibitem{radford2018improving}
Alec Radford, Karthik Narasimhan, Tim Salimans, and Ilya Sutskever.
\newblock Improving language understanding by generative pre-training.
\newblock 2018.

\bibitem{radford2019language}
Alec Radford, Jeffrey Wu, Rewon Child, David Luan, Dario Amodei, Ilya
  Sutskever, et~al.
\newblock Language models are unsupervised multitask learners.
\newblock {\em OpenAI blog}, 1(8):9, 2019.

\bibitem{rosset2020turing}
Corby Rosset.
\newblock Turing-nlg: A 17-billion-parameter language model by microsoft.
\newblock {\em Microsoft Blog}, 1(2), 2020.

\bibitem{zeng2021pangualpha}
Wei Zeng, Xiaozhe Ren, Teng Su, Hui Wang, Yi~Liao, Zhiwei Wang, Xin Jiang,
  ZhenZhang Yang, Kaisheng Wang, Xiaoda Zhang, et~al.
\newblock Pangu-$alpha$: Large-scale autoregressive pretrained chinese language
  models with auto-parallel computation.
\newblock {\em arXiv preprint arXiv:2104.12369}, 2021.

\bibitem{wei2019nezha}
Junqiu Wei, Xiaozhe Ren, Xiaoguang Li, Wenyong Huang, Yi~Liao, Yasheng Wang,
  Jiashu Lin, Xin Jiang, Xiao Chen, and Qun Liu.
\newblock Nezha: Neural contextualized representation for chinese language
  understanding.
\newblock {\em arXiv preprint arXiv:1909.00204}, 2019.

\bibitem{chen2020generative}
Mark Chen, Alec Radford, Rewon Child, Jeffrey Wu, Heewoo Jun, David Luan, and
  Ilya Sutskever.
\newblock Generative pretraining from pixels.
\newblock In {\em International Conference on Machine Learning}, pages
  1691--1703. PMLR, 2020.

\bibitem{dosovitskiy2020image}
Alexey Dosovitskiy, Lucas Beyer, Alexander Kolesnikov, Dirk Weissenborn,
  Xiaohua Zhai, Thomas Unterthiner, Mostafa Dehghani, Matthias Minderer, Georg
  Heigold, Sylvain Gelly, et~al.
\newblock An image is worth 16x16 words: Transformers for image recognition at
  scale.
\newblock In {\em International Conference on Learning Representations}, 2020.

\bibitem{carion2020end}
Nicolas Carion, Francisco Massa, Gabriel Synnaeve, Nicolas Usunier, Alexander
  Kirillov, and Sergey Zagoruyko.
\newblock End-to-end object detection with transformers.
\newblock In {\em European conference on computer vision}, pages 213--229.
  Springer, 2020.

\bibitem{zheng2021rethinking}
Sixiao Zheng, Jiachen Lu, Hengshuang Zhao, Xiatian Zhu, Zekun Luo, Yabiao Wang,
  Yanwei Fu, Jianfeng Feng, Tao Xiang, Philip~HS Torr, et~al.
\newblock Rethinking semantic segmentation from a sequence-to-sequence
  perspective with transformers.
\newblock In {\em Proceedings of the IEEE/CVF conference on computer vision and
  pattern recognition}, pages 6881--6890, 2021.

\bibitem{chen2021pre}
Hanting Chen, Yunhe Wang, Tianyu Guo, Chang Xu, Yiping Deng, Zhenhua Liu, Siwei
  Ma, Chunjing Xu, Chao Xu, and Wen Gao.
\newblock Pre-trained image processing transformer.
\newblock In {\em Proceedings of the IEEE/CVF Conference on Computer Vision and
  Pattern Recognition}, pages 12299--12310, 2021.

\bibitem{he2021masked}
Kaiming He, Xinlei Chen, Saining Xie, Yanghao Li, Piotr Doll{\'a}r, and Ross
  Girshick.
\newblock Masked autoencoders are scalable vision learners.
\newblock {\em arXiv preprint arXiv:2111.06377}, 2021.

\bibitem{bao2021beit}
Hangbo Bao, Li~Dong, and Furu Wei.
\newblock Beit: Bert pre-training of image transformers.
\newblock {\em arXiv preprint arXiv:2106.08254}, 2021.

\bibitem{dong2021peco}
Xiaoyi Dong, Jianmin Bao, Ting Zhang, Dongdong Chen, Weiming Zhang, Lu~Yuan,
  Dong Chen, Fang Wen, and Nenghai Yu.
\newblock Peco: Perceptual codebook for bert pre-training of vision
  transformers.
\newblock {\em arXiv preprint arXiv:2111.12710}, 2021.

\bibitem{schneider2019wav2vec}
Steffen Schneider, Alexei Baevski, Ronan Collobert, and Michael Auli.
\newblock wav2vec: Unsupervised pre-training for speech recognition.
\newblock {\em arXiv preprint arXiv:1904.05862}, 2019.

\bibitem{baevski2019effectiveness}
Alexei Baevski, Michael Auli, and Abdelrahman Mohamed.
\newblock Effectiveness of self-supervised pre-training for speech recognition.
\newblock {\em arXiv preprint arXiv:1911.03912}, 2019.

\bibitem{hsu2021hubert}
Wei-Ning Hsu, Benjamin Bolte, Yao-Hung~Hubert Tsai, Kushal Lakhotia, Ruslan
  Salakhutdinov, and Abdelrahman Mohamed.
\newblock Hubert: Self-supervised speech representation learning by masked
  prediction of hidden units.
\newblock {\em IEEE/ACM Transactions on Audio, Speech, and Language
  Processing}, 29:3451--3460, 2021.

\bibitem{baevski2020wav2vec}
Alexei Baevski, Yuhao Zhou, Abdelrahman Mohamed, and Michael Auli.
\newblock wav2vec 2.0: A framework for self-supervised learning of speech
  representations.
\newblock {\em Advances in Neural Information Processing Systems},
  33:12449--12460, 2020.

\bibitem{chung2021w2v}
Yu-An Chung, Yu~Zhang, Wei Han, Chung-Cheng Chiu, James Qin, Ruoming Pang, and
  Yonghui Wu.
\newblock W2v-bert: Combining contrastive learning and masked language modeling
  for self-supervised speech pre-training.
\newblock {\em arXiv preprint arXiv:2108.06209}, 2021.

\bibitem{zhu2022promptUIC}
Peipei Zhu, Xiao Wang, Lin Zhu, Zhenglong Sun, Weishi Zheng, Yaowei Wang, and
  Changwen Chen.
\newblock Prompt-based learning for unpaired image captioning.
\newblock {\em arXiv preprint arXiv:2205.13125}, 2022.

\bibitem{radford2021CLIP}
Alec Radford, Jong~Wook Kim, Chris Hallacy, Aditya Ramesh, Gabriel Goh,
  Sandhini Agarwal, Girish Sastry, Amanda Askell, Pamela Mishkin, Jack Clark,
  et~al.
\newblock Learning transferable visual models from natural language
  supervision.
\newblock In {\em International Conference on Machine Learning}, pages
  8748--8763. PMLR, 2021.

\bibitem{xing2022classprompt}
Yinghui Xing, Qirui Wu, De~Cheng, Shizhou Zhang, Guoqiang Liang, and Yanning
  Zhang.
\newblock Class-aware visual prompt tuning for vision-language pre-trained
  model.
\newblock {\em arXiv preprint arXiv:2208.08340}, 2022.

\bibitem{ordonez2011im2text}
Vicente Ordonez, Girish Kulkarni, and Tamara Berg.
\newblock Im2text: Describing images using 1 million captioned photographs.
\newblock {\em Advances in neural information processing systems}, 24, 2011.

\bibitem{young2014image}
Peter Young, Alice Lai, Micah Hodosh, and Julia Hockenmaier.
\newblock From image descriptions to visual denotations: New similarity metrics
  for semantic inference over event descriptions.
\newblock {\em Transactions of the Association for Computational Linguistics},
  2:67--78, 2014.

\bibitem{lin2014coco}
Tsung-Yi Lin, Michael Maire, Serge Belongie, James Hays, Pietro Perona, Deva
  Ramanan, Piotr Doll{\'a}r, and C~Lawrence Zitnick.
\newblock Microsoft coco: Common objects in context.
\newblock In {\em European conference on computer vision}, pages 740--755.
  Springer, 2014.

\bibitem{krishna2017visualgenome}
Ranjay Krishna, Yuke Zhu, Oliver Groth, Justin Johnson, Kenji Hata, Joshua
  Kravitz, Stephanie Chen, Yannis Kalantidis, Li-Jia Li, David~A Shamma, et~al.
\newblock Visual genome: Connecting language and vision using crowdsourced
  dense image annotations.
\newblock {\em International journal of computer vision}, 123(1):32--73, 2017.

\bibitem{goyal2017VQA}
Yash Goyal, Tejas Khot, Douglas Summers-Stay, Dhruv Batra, and Devi Parikh.
\newblock Making the v in vqa matter: Elevating the role of image understanding
  in visual question answering.
\newblock In {\em Proceedings of the IEEE conference on computer vision and
  pattern recognition}, pages 6904--6913, 2017.

\bibitem{rostamzadeh2018fashion}
Negar Rostamzadeh, Seyedarian Hosseini, Thomas Boquet, Wojciech Stokowiec, Ying
  Zhang, Christian Jauvin, and Chris Pal.
\newblock Fashion-gen: The generative fashion dataset and challenge.
\newblock {\em arXiv preprint arXiv:1806.08317}, 2018.

\bibitem{sharma2018conceptual}
Piyush Sharma, Nan Ding, Sebastian Goodman, and Radu Soricut.
\newblock Conceptual captions: A cleaned, hypernymed, image alt-text dataset
  for automatic image captioning.
\newblock In {\em Proceedings of the 56th Annual Meeting of the Association for
  Computational Linguistics (Volume 1: Long Papers)}, pages 2556--2565, 2018.

\bibitem{hudson2019gqa}
Drew~A Hudson and Christopher~D Manning.
\newblock Gqa: A new dataset for real-world visual reasoning and compositional
  question answering.
\newblock In {\em Proceedings of the IEEE/CVF conference on computer vision and
  pattern recognition}, pages 6700--6709, 2019.

\bibitem{qi2020imagebert}
Di~Qi, Lin Su, Jia Song, Edward Cui, Taroon Bharti, and Arun Sacheti.
\newblock Imagebert: Cross-modal pre-training with large-scale weak-supervised
  image-text data.
\newblock {\em arXiv preprint arXiv:2001.07966}, 2020.

\bibitem{changpinyo2021conceptual}
Soravit Changpinyo, Piyush Sharma, Nan Ding, and Radu Soricut.
\newblock Conceptual 12m: Pushing web-scale image-text pre-training to
  recognize long-tail visual concepts.
\newblock In {\em Proceedings of the IEEE/CVF Conference on Computer Vision and
  Pattern Recognition}, pages 3558--3568, 2021.

\bibitem{jia2021scaling}
Chao Jia, Yinfei Yang, Ye~Xia, Yi-Ting Chen, Zarana Parekh, Hieu Pham, Quoc Le,
  Yun-Hsuan Sung, Zhen Li, and Tom Duerig.
\newblock Scaling up visual and vision-language representation learning with
  noisy text supervision.
\newblock In {\em International Conference on Machine Learning}, pages
  4904--4916. PMLR, 2021.

\bibitem{lei2018tvqa}
Jie Lei, Licheng Yu, Mohit Bansal, and Tamara Berg.
\newblock Tvqa: Localized, compositional video question answering.
\newblock In {\em Proceedings of the 2018 Conference on Empirical Methods in
  Natural Language Processing}, pages 1369--1379, 2018.

\bibitem{miech2019howto100m}
Antoine Miech, Dimitri Zhukov, Jean-Baptiste Alayrac, Makarand Tapaswi, Ivan
  Laptev, and Josef Sivic.
\newblock Howto100m: Learning a text-video embedding by watching hundred
  million narrated video clips.
\newblock In {\em Proceedings of the IEEE/CVF International Conference on
  Computer Vision}, pages 2630--2640, 2019.

\bibitem{bain2021frozen}
Max Bain, Arsha Nagrani, G{\"u}l Varol, and Andrew Zisserman.
\newblock Frozen in time: A joint video and image encoder for end-to-end
  retrieval.
\newblock In {\em Proceedings of the IEEE/CVF International Conference on
  Computer Vision}, pages 1728--1738, 2021.

\bibitem{thomee2016yfcc100m}
Bart Thomee, David~A Shamma, Gerald Friedland, Benjamin Elizalde, Karl Ni,
  Douglas Poland, Damian Borth, and Li-Jia Li.
\newblock Yfcc100m: The new data in multimedia research.
\newblock {\em Communications of the ACM}, 59(2):64--73, 2016.

\bibitem{schuhmann2021laion}
Christoph Schuhmann, Robert Kaczmarczyk, Aran Komatsuzaki, Aarush Katta,
  Richard Vencu, Romain Beaumont, Jenia Jitsev, Theo Coombes, and Clayton
  Mullis.
\newblock Laion-400m: Open dataset of clip-filtered 400 million image-text
  pairs.
\newblock In {\em NeurIPS Workshop Datacentric AI}, number FZJ-2022-00923.
  J{\"u}lich Supercomputing Center, 2021.

\bibitem{desai2021redcaps}
Karan Desai, Gaurav Kaul, Zubin Aysola, and Justin Johnson.
\newblock {RedCaps: Web-curated image-text data created by the people, for the
  people}.
\newblock In {\em NeurIPS Datasets and Benchmarks}, 2021.

\bibitem{gu2022wukong}
Jiaxi Gu, Xiaojun Meng, Guansong Lu, Lu~Hou, Minzhe Niu, Hang Xu, Xiaodan
  Liang, Wei Zhang, Xin Jiang, and Chunjing Xu.
\newblock Wukong: 100 million large-scale chinese cross-modal pre-training
  dataset and a foundation framework, 2022.

\bibitem{parekh2021crisscrossed}
Zarana Parekh, Jason Baldridge, Daniel Cer, Austin Waters, and Yinfei Yang.
\newblock Crisscrossed captions: Extended intramodal and intermodal semantic
  similarity judgments for ms-coco.
\newblock In {\em Proceedings of the 16th Conference of the European Chapter of
  the Association for Computational Linguistics: Main Volume}, pages
  2855--2870, 2021.

\bibitem{zhan2021product1m}
Xunlin Zhan, Yangxin Wu, Xiao Dong, Yunchao Wei, Minlong Lu, Yichi Zhang, Hang
  Xu, and Xiaodan Liang.
\newblock Product1m: Towards weakly supervised instance-level product retrieval
  via cross-modal pretraining.
\newblock In {\em Proceedings of the IEEE/CVF International Conference on
  Computer Vision}, pages 11782--11791, 2021.

\bibitem{srinivasan2021wit}
Krishna Srinivasan, Karthik Raman, Jiecao Chen, Michael Bendersky, and Marc
  Najork.
\newblock Wit: Wikipedia-based image text dataset for multimodal multilingual
  machine learning.
\newblock In {\em Proceedings of the 44th International ACM SIGIR Conference on
  Research and Development in Information Retrieval}, pages 2443--2449, 2021.

\bibitem{sun2017JFT300M}
Chen Sun, Abhinav Shrivastava, Saurabh Singh, and Abhinav Gupta.
\newblock Revisiting unreasonable effectiveness of data in deep learning era.
\newblock In {\em Proceedings of the IEEE international conference on computer
  vision}, pages 843--852, 2017.

\bibitem{yang2021JFT3B}
Jianwei Yang, Chunyuan Li, Pengchuan Zhang, Xiyang Dai, Bin Xiao, Lu~Yuan, and
  Jianfeng Gao.
\newblock Focal self-attention for local-global interactions in vision
  transformers.
\newblock {\em arXiv preprint arXiv:2107.00641}, 2021.

\bibitem{mahajan2018exploring}
Dhruv Mahajan, Ross Girshick, Vignesh Ramanathan, Kaiming He, Manohar Paluri,
  Yixuan Li, Ashwin Bharambe, and Laurens Van Der~Maaten.
\newblock Exploring the limits of weakly supervised pretraining.
\newblock In {\em Proceedings of the European conference on computer vision
  (ECCV)}, pages 181--196, 2018.

\bibitem{lin2021m6}
Junyang Lin, Rui Men, An~Yang, Chang Zhou, Ming Ding, Yichang Zhang, Peng Wang,
  Ang Wang, Le~Jiang, Xianyan Jia, et~al.
\newblock M6: A chinese multimodal pretrainer.
\newblock {\em arXiv preprint arXiv:2103.00823}, 2021.

\bibitem{dong2021m5product}
Xiao Dong, Xunlin Zhan, Yangxin Wu, Yunchao Wei, Xiaoyong Wei, Minlong Lu, and
  Xiaodan Liang.
\newblock M5product: A multi-modal pretraining benchmark for e-commercial
  product downstream tasks.
\newblock {\em arXiv preprint arXiv:2109.04275}, 2021.

\bibitem{pont2020connecting}
Jordi Pont-Tuset, Jasper Uijlings, Soravit Changpinyo, Radu Soricut, and
  Vittorio Ferrari.
\newblock Connecting vision and language with localized narratives.
\newblock In {\em European Conference on Computer Vision}, pages 647--664.
  Springer, 2020.

\bibitem{huo2021wenlan}
Yuqi Huo, Manli Zhang, Guangzhen Liu, Haoyu Lu, Yizhao Gao, Guoxing Yang,
  Jingyuan Wen, Heng Zhang, Baogui Xu, Weihao Zheng, et~al.
\newblock Wenlan: Bridging vision and language by large-scale multi-modal
  pre-training.
\newblock {\em arXiv preprint arXiv:2103.06561}, 2021.

\bibitem{Sha2022WuDaoMM}
Leng Jiahong Xue Zhao Zhao~Hanyu Sha~Yuan, Zhao~Shuai and Tang Jie.
\newblock Wudaomm: A large-scale multi-modal dataset for pre-training models.
\newblock {\em arXiv preprint arXiv:2203.11480}, 2022.

\bibitem{chenmep}
Delong Chen, Fan Liu, Xiaoyu Du, Ruizhuo Gao, and Feng Xu.
\newblock Mep-3m: A large-scale multi-modal e-commerce products dataset.

\bibitem{fei2021wenlanV2}
Nanyi Fei, Zhiwu Lu, Yizhao Gao, Guoxing Yang, Yuqi Huo, Jingyuan Wen, Haoyu
  Lu, Ruihua Song, Xin Gao, Tao Xiang, et~al.
\newblock Wenlan 2.0: Make ai imagine via a multimodal foundation model.
\newblock {\em arXiv preprint arXiv:2110.14378}, 2021.

\bibitem{hodosh2013framing}
Micah Hodosh, Peter Young, and Julia Hockenmaier.
\newblock Framing image description as a ranking task: Data, models and
  evaluation metrics.
\newblock {\em Journal of Artificial Intelligence Research}, 47:853--899, 2013.

\bibitem{chen2015cococaptions}
Xinlei Chen, Hao Fang, Tsung-Yi Lin, Ramakrishna Vedantam, Saurabh Gupta, Piotr
  Doll{\'a}r, and C~Lawrence Zitnick.
\newblock Microsoft coco captions: Data collection and evaluation server.
\newblock {\em arXiv preprint arXiv:1504.00325}, 2015.

\bibitem{zhou2017ade20k}
Bolei Zhou, Hang Zhao, Xavier Puig, Sanja Fidler, Adela Barriuso, and Antonio
  Torralba.
\newblock Scene parsing through ade20k dataset.
\newblock In {\em Proceedings of the IEEE conference on computer vision and
  pattern recognition}, pages 633--641, 2017.

\bibitem{zhang2021vinvl}
Pengchuan Zhang, Xiujun Li, Xiaowei Hu, Jianwei Yang, Lei Zhang, Lijuan Wang,
  Yejin Choi, and Jianfeng Gao.
\newblock Vinvl: Revisiting visual representations in vision-language models.
\newblock In {\em Proceedings of the IEEE/CVF Conference on Computer Vision and
  Pattern Recognition}, pages 5579--5588, 2021.

\bibitem{li2020Unicodervl}
Gen Li, Nan Duan, Yuejian Fang, Ming Gong, and Daxin Jiang.
\newblock Unicoder-vl: A universal encoder for vision and language by
  cross-modal pre-training.
\newblock In {\em Proceedings of the AAAI Conference on Artificial
  Intelligence}, volume~34, pages 11336--11344, 2020.

\bibitem{lin2020interbert}
Junyang Lin, An~Yang, Yichang Zhang, Jie Liu, Jingren Zhou, and Hongxia Yang.
\newblock Interbert: Vision-and-language interaction for multi-modal
  pretraining.
\newblock {\em arXiv preprint arXiv:2003.13198}, 2020.

\bibitem{wang2021simvlm}
Zirui Wang, Jiahui Yu, Adams~Wei Yu, Zihang Dai, Yulia Tsvetkov, and Yuan Cao.
\newblock Simvlm: Simple visual language model pretraining with weak
  supervision.
\newblock {\em arXiv preprint arXiv:2108.10904}, 2021.

\bibitem{tan2019lxmert}
Hao Tan and Mohit Bansal.
\newblock Lxmert: Learning cross-modality encoder representations from
  transformers.
\newblock In {\em Proceedings of the 2019 Conference on Empirical Methods in
  Natural Language Processing and the 9th International Joint Conference on
  Natural Language Processing (EMNLP-IJCNLP)}, pages 5100--5111, 2019.

\bibitem{ELMo2018}
Matthew~E. Peters, Mark Neumann, Mohit Iyyer, Matt Gardner, Christopher Clark,
  Kenton Lee, and Luke Zettlemoyer.
\newblock Deep contextualized word representations.
\newblock In {\em Proceedings of the 2018 Conference of the North {A}merican
  Chapter of the Association for Computational Linguistics: Human Language
  Technologies, Volume 1 (Long Papers)}, pages 2227--2237, New Orleans,
  Louisiana, June 2018. Association for Computational Linguistics.

\bibitem{dong2019unilm}
Li~Dong, Nan Yang, Wenhui Wang, Furu Wei, Xiaodong Liu, Yu~Wang, Jianfeng Gao,
  Ming Zhou, and Hsiao-Wuen Hon.
\newblock Unified language model pre-training for natural language
  understanding and generation.
\newblock {\em Advances in Neural Information Processing Systems}, 32, 2019.

\bibitem{peyre2019computationalOT}
Gabriel Peyr{\'e}, Marco Cuturi, et~al.
\newblock Computational optimal transport: With applications to data science.
\newblock {\em Foundations and Trends{\textregistered} in Machine Learning},
  11(5-6):355--607, 2019.

\bibitem{xie2020fast}
Yujia Xie, Xiangfeng Wang, Ruijia Wang, and Hongyuan Zha.
\newblock A fast proximal point method for computing exact wasserstein
  distance.
\newblock In {\em Uncertainty in artificial intelligence}, pages 433--453.
  PMLR, 2020.

\bibitem{hao2020PREVALENT}
Weituo Hao, Chunyuan Li, Xiujun Li, Lawrence Carin, and Jianfeng Gao.
\newblock Towards learning a generic agent for vision-and-language navigation
  via pre-training.
\newblock In {\em Proceedings of the IEEE/CVF Conference on Computer Vision and
  Pattern Recognition}, pages 13137--13146, 2020.

\bibitem{yu2021ernieViL}
Fei Yu, Jiji Tang, Weichong Yin, Yu~Sun, Hao Tian, Hua Wu, and Haifeng Wang.
\newblock Ernie-vil: Knowledge enhanced vision-language representations through
  scene graphs.
\newblock In {\em Proceedings of the AAAI Conference on Artificial
  Intelligence}, volume~35, pages 3208--3216, 2021.

\bibitem{zhuge2021kaleidoBERT}
Mingchen Zhuge, Dehong Gao, Deng-Ping Fan, Linbo Jin, Ben Chen, Haoming Zhou,
  Minghui Qiu, and Ling Shao.
\newblock Kaleido-bert: Vision-language pre-training on fashion domain.
\newblock In {\em Proceedings of the IEEE/CVF Conference on Computer Vision and
  Pattern Recognition}, pages 12647--12657, 2021.

\bibitem{xu2021E2EVLP}
Haiyang Xu, Ming Yan, Chenliang Li, Bin Bi, Songfang Huang, Wenming Xiao, and
  Fei Huang.
\newblock E2e-vlp: End-to-end vision-language pre-training enhanced by visual
  learning.
\newblock In {\em Proceedings of the 59th Annual Meeting of the Association for
  Computational Linguistics and the 11th International Joint Conference on
  Natural Language Processing (Volume 1: Long Papers)}, pages 503--513, 2021.

\bibitem{li2020hero}
Linjie Li, Yen-Chun Chen, Yu~Cheng, Zhe Gan, Licheng Yu, and Jingjing Liu.
\newblock Hero: Hierarchical encoder for video+ language omni-representation
  pre-training.
\newblock In {\em Proceedings of the 2020 Conference on Empirical Methods in
  Natural Language Processing (EMNLP)}, pages 2046--2065, 2020.

\bibitem{ling2022vision}
Yan Ling, Rui Xia, et~al.
\newblock Vision-language pre-training for multimodal aspect-based sentiment
  analysis.
\newblock {\em arXiv preprint arXiv:2204.07955}, 2022.

\bibitem{cui2021rosita}
Yuhao Cui, Zhou Yu, Chunqi Wang, Zhongzhou Zhao, Ji~Zhang, Meng Wang, and Jun
  Yu.
\newblock Rosita: Enhancing vision-and-language semantic alignments via
  cross-and intra-modal knowledge integration.
\newblock In {\em Proceedings of the 29th ACM International Conference on
  Multimedia}, pages 797--806, 2021.

\bibitem{guo2022attentionSurvey}
Meng-Hao Guo, Tian-Xing Xu, Jiang-Jiang Liu, Zheng-Ning Liu, Peng-Tao Jiang,
  Tai-Jiang Mu, Song-Hai Zhang, Ralph~R Martin, Ming-Ming Cheng, and Shi-Min
  Hu.
\newblock Attention mechanisms in computer vision: A survey.
\newblock {\em Computational Visual Media}, pages 1--38, 2022.

\bibitem{li2021ALBEF}
Junnan Li, Ramprasaath Selvaraju, Akhilesh Gotmare, Shafiq Joty, Caiming Xiong,
  and Steven Chu~Hong Hoi.
\newblock Align before fuse: Vision and language representation learning with
  momentum distillation.
\newblock {\em Advances in Neural Information Processing Systems}, 34, 2021.

\bibitem{yang2022Icode}
Ziyi Yang, Yuwei Fang, Chenguang Zhu, Reid Pryzant, Dongdong Chen, Yu~Shi,
  Yichong Xu, Yao Qian, Mei Gao, Yi-Ling Chen, et~al.
\newblock i-code: An integrative and composable multimodal learning framework.
\newblock {\em arXiv preprint arXiv:2205.01818}, 2022.

\bibitem{SuoS0021ijcai}
Wei Suo, Mengyang Sun, Peng Wang, and Qi~Wu.
\newblock Proposal-free one-stage referring expression via grid-word
  cross-attention.
\newblock In Zhi{-}Hua Zhou, editor, {\em Proceedings of the Thirtieth
  International Joint Conference on Artificial Intelligence, {IJCAI} 2021,
  Virtual Event / Montreal, Canada, 19-27 August 2021}, pages 1032--1038.
  ijcai.org, 2021.

\bibitem{zhu2020actbert}
Linchao Zhu and Yi~Yang.
\newblock Actbert: Learning global-local video-text representations.
\newblock In {\em Proceedings of the IEEE/CVF conference on computer vision and
  pattern recognition}, pages 8746--8755, 2020.

\bibitem{wang2021actionclip}
Mengmeng Wang, Jiazheng Xing, and Yong Liu.
\newblock Actionclip: A new paradigm for video action recognition.
\newblock {\em arXiv preprint arXiv:2109.08472}, 2021.

\bibitem{li2022clipevent}
Manling Li, Ruochen Xu, Shuohang Wang, Luowei Zhou, Xudong Lin, Chenguang Zhu,
  Michael Zeng, Heng Ji, and Shih-Fu Chang.
\newblock Clip-event: Connecting text and images with event structures.
\newblock {\em arXiv preprint arXiv:2201.05078}, 2022.

\bibitem{cui2022CLIPbenchmark}
Yufeng Cui, Lichen Zhao, Feng Liang, Yangguang Li, and Jing Shao.
\newblock Democratizing contrastive language-image pre-training: A clip
  benchmark of data, model, and supervision.
\newblock {\em arXiv preprint arXiv:2203.05796}, 2022.

\bibitem{shen2021CLIPViL}
Sheng Shen, Liunian~Harold Li, Hao Tan, Mohit Bansal, Anna Rohrbach, Kai-Wei
  Chang, Zhewei Yao, and Kurt Keutzer.
\newblock How much can clip benefit vision-and-language tasks?
\newblock {\em arXiv preprint arXiv:2107.06383}, 2021.

\bibitem{Chen2022ProtoCLIP}
Chen Delong, Wu~Zhao, Liu Fan, Yang Zaiquan, Huang Yixiang, Bao Yiping, and
  Zhou Erjin.
\newblock Prototypical contrastive language image pretraining.
\newblock {\em arXiv preprint arXiv:2206.10996}, 2022.

\bibitem{li2019visualbert}
Liunian~Harold Li, Mark Yatskar, Da~Yin, Cho-Jui Hsieh, and Kai-Wei Chang.
\newblock Visualbert: A simple and performant baseline for vision and language.
\newblock {\em arXiv preprint arXiv:1908.03557}, 2019.

\bibitem{lu2019vilbert}
Jiasen Lu, Dhruv Batra, Devi Parikh, and Stefan Lee.
\newblock Vilbert: Pretraining task-agnostic visiolinguistic representations
  for vision-and-language tasks.
\newblock {\em Advances in neural information processing systems}, 32, 2019.

\bibitem{alberti2019B2T2}
Chris Alberti, Jeffrey Ling, Michael Collins, and David Reitter.
\newblock Fusion of detected objects in text for visual question answering.
\newblock In {\em Proceedings of the 2019 Conference on Empirical Methods in
  Natural Language Processing and the 9th International Joint Conference on
  Natural Language Processing (EMNLP-IJCNLP)}, pages 2131--2140, 2019.

\bibitem{su2019vlbert}
Weijie Su, Xizhou Zhu, Yue Cao, Bin Li, Lewei Lu, Furu Wei, and Jifeng Dai.
\newblock Vl-bert: Pre-training of generic visual-linguistic representations.
\newblock In {\em International Conference on Learning Representations}, 2019.

\bibitem{zhouLuowei2020VLP}
Luowei Zhou, Hamid Palangi, Lei Zhang, Houdong Hu, Jason Corso, and Jianfeng
  Gao.
\newblock Unified vision-language pre-training for image captioning and vqa.
\newblock In {\em Proceedings of the AAAI Conference on Artificial
  Intelligence}, volume~34, pages 13041--13049, 2020.

\bibitem{lu20201212in1}
Jiasen Lu, Vedanuj Goswami, Marcus Rohrbach, Devi Parikh, and Stefan Lee.
\newblock 12-in-1: Multi-task vision and language representation learning.
\newblock In {\em Proceedings of the IEEE/CVF Conference on Computer Vision and
  Pattern Recognition}, pages 10437--10446, 2020.

\bibitem{murahari2020vdialogbert}
Vishvak Murahari, Dhruv Batra, Devi Parikh, and Abhishek Das.
\newblock Large-scale pretraining for visual dialog: A simple state-of-the-art
  baseline.
\newblock In {\em European Conference on Computer Vision}, pages 336--352.
  Springer, 2020.

\bibitem{gaoPyramidCLIP2022}
Gao Yuting, Liu Jinfeng, Xu~Zihan, Zhang Jun, Li~Ke, and Shen Chunhua.
\newblock Pyramidclip: Hierarchical feature alignment for vision-language model
  pretraining.
\newblock In {\em arXiv:2204.14095}, 2022.

\bibitem{gao2020fashionbert}
Dehong Gao, Linbo Jin, Ben Chen, Minghui Qiu, Peng Li, Yi~Wei, Yi~Hu, and Hao
  Wang.
\newblock Fashionbert: Text and image matching with adaptive loss for
  cross-modal retrieval.
\newblock In {\em Proceedings of the 43rd International ACM SIGIR Conference on
  Research and Development in Information Retrieval}, pages 2251--2260, 2020.

\bibitem{gan2020VILLA}
Zhe Gan, Yen-Chun Chen, Linjie Li, Chen Zhu, Yu~Cheng, and Jingjing Liu.
\newblock Large-scale adversarial training for vision-and-language
  representation learning.
\newblock {\em Advances in Neural Information Processing Systems},
  33:6616--6628, 2020.

\bibitem{song2021kvlbert}
Dandan Song, Siyi Ma, Zhanchen Sun, Sicheng Yang, and Lejian Liao.
\newblock Kvl-bert: Knowledge enhanced visual-and-linguistic bert for visual
  commonsense reasoning.
\newblock {\em Knowledge-Based Systems}, 230:107408, 2021.

\bibitem{cho2021VLT5}
Jaemin Cho, Jie Lei, Hao Tan, and Mohit Bansal.
\newblock Unifying vision-and-language tasks via text generation.
\newblock In {\em International Conference on Machine Learning}, pages
  1931--1942. PMLR, 2021.

\bibitem{kim2021vilt}
Wonjae Kim, Bokyung Son, and Ildoo Kim.
\newblock Vilt: Vision-and-language transformer without convolution or region
  supervision.
\newblock In {\em International Conference on Machine Learning}, pages
  5583--5594. PMLR, 2021.

\bibitem{kamath2021mdetr}
Aishwarya Kamath, Mannat Singh, Yann LeCun, Gabriel Synnaeve, Ishan Misra, and
  Nicolas Carion.
\newblock Mdetr-modulated detection for end-to-end multi-modal understanding.
\newblock In {\em Proceedings of the IEEE/CVF International Conference on
  Computer Vision}, pages 1780--1790, 2021.

\bibitem{huang2021SOHO}
Zhicheng Huang, Zhaoyang Zeng, Yupan Huang, Bei Liu, Dongmei Fu, and Jianlong
  Fu.
\newblock Seeing out of the box: End-to-end pre-training for vision-language
  representation learning.
\newblock In {\em Proceedings of the IEEE/CVF Conference on Computer Vision and
  Pattern Recognition}, pages 12976--12985, 2021.

\bibitem{xue2021VisualParsing}
Hongwei Xue, Yupan Huang, Bei Liu, Houwen Peng, Jianlong Fu, Houqiang Li, and
  Jiebo Luo.
\newblock Probing inter-modality: Visual parsing with self-attention for
  vision-and-language pre-training.
\newblock {\em Advances in Neural Information Processing Systems}, 34, 2021.

\bibitem{jain2021mural}
Aashi Jain, Mandy Guo, Krishna Srinivasan, Ting Chen, Sneha Kudugunta, Chao
  Jia, Yinfei Yang, and Jason Baldridge.
\newblock Mural: multimodal, multitask retrieval across languages.
\newblock {\em arXiv preprint arXiv:2109.05125}, 2021.

\bibitem{wang2021vlmo}
Wenhui Wang, Hangbo Bao, Li~Dong, and Furu Wei.
\newblock Vlmo: Unified vision-language pre-training with
  mixture-of-modality-experts.
\newblock {\em arXiv preprint arXiv:2111.02358}, 2021.

\bibitem{dou2021METER}
Zi-Yi Dou, Yichong Xu, Zhe Gan, Jianfeng Wang, Shuohang Wang, Lijuan Wang,
  Chenguang Zhu, Zicheng Liu, Michael Zeng, et~al.
\newblock An empirical study of training end-to-end vision-and-language
  transformers.
\newblock {\em arXiv preprint arXiv:2111.02387}, 2021.

\bibitem{sun2019videobert}
Chen Sun, Austin Myers, Carl Vondrick, Kevin Murphy, and Cordelia Schmid.
\newblock Videobert: A joint model for video and language representation
  learning.
\newblock In {\em Proceedings of the IEEE/CVF International Conference on
  Computer Vision}, pages 7464--7473, 2019.

\bibitem{sun2019CBT}
Chen Sun, Fabien Baradel, Kevin Murphy, and Cordelia Schmid.
\newblock Learning video representations using contrastive bidirectional
  transformer.
\newblock {\em arXiv preprint arXiv:1906.05743}, 2019.

\bibitem{luo2020univl}
Huaishao Luo, Lei Ji, Botian Shi, Haoyang Huang, Nan Duan, Tianrui Li, Jason
  Li, Taroon Bharti, and Ming Zhou.
\newblock Univl: A unified video and language pre-training model for multimodal
  understanding and generation.
\newblock {\em arXiv preprint arXiv:2002.06353}, 2020.

\bibitem{urooj2020mmftBERT}
Aisha Urooj, Amir Mazaheri, Mubarak Shah, et~al.
\newblock Mmft-bert: Multimodal fusion transformer with bert encodings for
  visual question answering.
\newblock In {\em Findings of the Association for Computational Linguistics:
  EMNLP 2020}, pages 4648--4660, 2020.

\bibitem{yan2021RegionLearner}
Rui Yan, Mike~Zheng Shou, Yixiao Ge, Alex~Jinpeng Wang, Xudong Lin, Guanyu Cai,
  and Jinhui Tang.
\newblock Video-text pre-training with learned regions.
\newblock {\em arXiv preprint arXiv:2112.01194}, 2021.

\bibitem{li2020unimo}
Wei Li, Can Gao, Guocheng Niu, Xinyan Xiao, Hao Liu, Jiachen Liu, Hua Wu, and
  Haifeng Wang.
\newblock Unimo: Towards unified-modal understanding and generation via
  cross-modal contrastive learning.
\newblock {\em arXiv preprint arXiv:2012.15409}, 2020.

\bibitem{ramesh2021DALLE}
Aditya Ramesh, Mikhail Pavlov, Gabriel Goh, Scott Gray, Chelsea Voss, Alec
  Radford, Mark Chen, and Ilya Sutskever.
\newblock Zero-shot text-to-image generation.
\newblock In {\em International Conference on Machine Learning}, pages
  8821--8831. PMLR, 2021.

\bibitem{Gui2022VLC}
Alex Hauptmann Yonatan Bisk Jianfeng~Gao Liangke~Gui, Qiuyuan~Huang.
\newblock Training vision-language transformers from captions alone.
\newblock {\em arXiv preprint arXiv:2205.09256}, 2022.

\bibitem{ding2021cogview}
Ming Ding, Zhuoyi Yang, Wenyi Hong, Wendi Zheng, Chang Zhou, Da~Yin, Junyang
  Lin, Xu~Zou, Zhou Shao, Hongxia Yang, et~al.
\newblock Cogview: Mastering text-to-image generation via transformers.
\newblock {\em Advances in Neural Information Processing Systems}, 34, 2021.

\bibitem{akbari2021vatt}
Hassan Akbari, Liangzhe Yuan, Rui Qian, Wei-Hong Chuang, Shih-Fu Chang, Yin
  Cui, and Boqing Gong.
\newblock Vatt: Transformers for multimodal self-supervised learning from raw
  video, audio and text.
\newblock {\em Advances in Neural Information Processing Systems}, 34, 2021.

\bibitem{yuan2021florence}
Lu~Yuan, Dongdong Chen, Yi-Ling Chen, Noel Codella, Xiyang Dai, Jianfeng Gao,
  Houdong Hu, Xuedong Huang, Boxin Li, Chunyuan Li, et~al.
\newblock Florence: A new foundation model for computer vision.
\newblock {\em arXiv preprint arXiv:2111.11432}, 2021.

\bibitem{Bakkali2022VLCDoC}
Mickael Coustaty Marçal Rusiñol Oriol Ramos~Terrades Souhail~Bakkali,
  Zuheng~Ming.
\newblock Hivlp: Hierarchical vision-language pre-training for fast image-text
  retrieval.
\newblock {\em arXiv preprint arXiv:2205.12029}, 2022.

\bibitem{wei2022mvp}
Longhui Wei, Lingxi Xie, Wengang Zhou, Houqiang Li, and Qi~Tian.
\newblock Mvp: Multimodality-guided visual pre-training.
\newblock {\em arXiv preprint arXiv:2203.05175}, 2022.

\bibitem{hong2021gilbert}
Weixiang Hong, Kaixiang Ji, Jiajia Liu, Jian Wang, Jingdong Chen, and Wei Chu.
\newblock Gilbert: Generative vision-language pre-training for image-text
  retrieval.
\newblock In {\em Proceedings of the 44th International ACM SIGIR Conference on
  Research and Development in Information Retrieval}, pages 1379--1388, 2021.

\bibitem{Lu2022COTS}
Yuqi Huo Yizhao Gao Zhiwu Lu Ji-Rong~Wen Haoyu~Lu, Nanyi~Fei.
\newblock Cots: Collaborative two-stream vision-language pre-training model for
  cross-modal retrieval.
\newblock In {\em arXiv:2204.07441}, 2022.

\bibitem{li2021Uvisualbert}
Liunian~Harold Li, Haoxuan You, Zhecan Wang, Alireza Zareian, Shih-Fu Chang,
  and Kai-Wei Chang.
\newblock Unsupervised vision-and-language pre-training without parallel images
  and captions.
\newblock In {\em Proceedings of the 2021 Conference of the North American
  Chapter of the Association for Computational Linguistics: Human Language
  Technologies}, pages 5339--5350, 2021.

\bibitem{alayrac2022flamingo}
Jean-Baptiste Alayrac, Jeff Donahue, Pauline Luc, Antoine Miech, Iain Barr,
  Yana Hasson, Karel Lenc, Arthur Mensch, Katie Millican, Malcolm Reynolds,
  et~al.
\newblock Flamingo: a visual language model for few-shot learning.
\newblock {\em arXiv preprint arXiv:2204.14198}, 2022.

\bibitem{ni2021m3p}
Minheng Ni, Haoyang Huang, Lin Su, Edward Cui, Taroon Bharti, Lijuan Wang,
  Dongdong Zhang, and Nan Duan.
\newblock M3p: Learning universal representations via multitask multilingual
  multimodal pre-training.
\newblock In {\em Proceedings of the IEEE/CVF Conference on Computer Vision and
  Pattern Recognition}, pages 3977--3986, 2021.

\bibitem{li2022BLIP}
Junnan Li, Dongxu Li, Caiming Xiong, and Steven Hoi.
\newblock Blip: Bootstrapping language-image pre-training for unified
  vision-language understanding and generation, 2022.

\bibitem{wu2021nvwa}
Chenfei Wu, Jian Liang, Lei Ji, Fan Yang, Yuejian Fang, Daxin Jiang, and Nan
  Duan.
\newblock N$\backslash$" uwa: Visual synthesis pre-training for neural visual
  world creation.
\newblock {\em arXiv preprint arXiv:2111.12417}, 2021.

\bibitem{yang2022vision}
Jinyu Yang, Jiali Duan, Son Tran, Yi~Xu, Sampath Chanda, Liqun Chen, Belinda
  Zeng, Trishul Chilimbi, and Junzhou Huang.
\newblock Vision-language pre-training with triple contrastive learning.
\newblock {\em arXiv preprint arXiv:2202.10401}, 2022.

\bibitem{weim5product}
Minlong~Lu Wei, Yaowei Wang, and Xiaodan Liang.
\newblock M5product: Self-harmonized contrastive learning for e-commercial
  multi-modal pretraining.

\bibitem{yan2022clinical}
Bin Yan and Mingtao Pei.
\newblock Clinical-bert: Vision-language pre-training for radiograph diagnosis
  and reports generation.
\newblock 2022.

\bibitem{zhong2021regionclip}
Yiwu Zhong, Jianwei Yang, Pengchuan Zhang, Chunyuan Li, Noel Codella,
  Liunian~Harold Li, Luowei Zhou, Xiyang Dai, Lu~Yuan, Yin Li, et~al.
\newblock Regionclip: Region-based language-image pretraining.
\newblock {\em arXiv preprint arXiv:2112.09106}, 2021.

\bibitem{liang2022visual}
Xiwen Liang, Fengda Zhu, Lingling Li, Hang Xu, and Xiaodan Liang.
\newblock Visual-language navigation pretraining via prompt-based environmental
  self-exploration.
\newblock {\em arXiv preprint arXiv:2203.04006}, 2022.

\bibitem{li2021GLIP}
Liunian~Harold Li*, Pengchuan Zhang*, Haotian Zhang*, Jianwei Yang, Chunyuan
  Li, Yiwu Zhong, Lijuan Wang, Lu~Yuan, Lei Zhang, Jenq-Neng Hwang, Kai-Wei
  Chang, and Jianfeng Gao.
\newblock Grounded language-image pre-training.
\newblock In {\em CVPR}, 2022.

\bibitem{Xie2022zeroR2D2}
Xie Chunyu, Cai Heng, Song Jianfei, Li~Jincheng, Kong Fanjing, Wu~Xiaoyu,
  Morimitsu Henrique, Yao Lin, Wang Dexin, Leng Dawei, Ji~Xiangyang, and Deng
  Yafeng.
\newblock Zero and r2d2: A large-scale chinese cross-modal benchmark and a
  vision-language framework.
\newblock {\em arXiv preprint arXiv:2205.03860}, 2022.

\bibitem{mu2021slip}
Norman Mu, Alexander Kirillov, David Wagner, and Saining Xie.
\newblock Slip: Self-supervision meets language-image pre-training.
\newblock {\em arXiv preprint arXiv:2112.12750}, 2021.

\bibitem{yao2021filip}
Lewei Yao, Runhui Huang, Lu~Hou, Guansong Lu, Minzhe Niu, Hang Xu, Xiaodan
  Liang, Zhenguo Li, Xin Jiang, and Chunjing Xu.
\newblock Filip: Fine-grained interactive language-image pre-training.
\newblock {\em arXiv preprint arXiv:2111.07783}, 2021.

\bibitem{li2021semvlp}
Chenliang Li, Ming Yan, Haiyang Xu, Fuli Luo, Wei Wang, Bin Bi, and Songfang
  Huang.
\newblock Semvlp: Vision-language pre-training by aligning semantics at
  multiple levels.
\newblock {\em arXiv preprint arXiv:2103.07829}, 2021.

\bibitem{yu2022coca}
Jiahui Yu, Zirui Wang, Vijay Vasudevan, Legg Yeung, Mojtaba Seyedhosseini, and
  Yonghui Wu.
\newblock Coca: Contrastive captioners are image-text foundation models.
\newblock {\em arXiv preprint arXiv:2205.01917}, 2022.

\bibitem{Chen2022HiVLP}
Jiaxin Shi Duzhen Zhang Jianlong~Chang Feilong~Chen, Xiuyi~Chen and Qi~Tian.
\newblock Hivlp: Hierarchical vision-language pre-training for fast image-text
  retrieval.
\newblock {\em arXiv preprint arXiv:2205.12105}, 2022.

\bibitem{guzhov2022audioclip}
Andrey Guzhov, Federico Raue, J{\"o}rn Hees, and Andreas Dengel.
\newblock Audioclip: Extending clip to image, text and audio.
\newblock In {\em ICASSP 2022-2022 IEEE International Conference on Acoustics,
  Speech and Signal Processing (ICASSP)}, pages 976--980. IEEE, 2022.

\bibitem{bao2022vlbeit}
Hangbo Bao, Wenhui Wang, Li~Dong, and Furu Wei.
\newblock Vl-beit: Generative vision-language pretraining.
\newblock {\em arXiv preprint arXiv:2206.01127}, 2022.

\bibitem{seo2022MVGPT}
Paul~Hongsuck Seo, Arsha Nagrani, Anurag Arnab, and Cordelia Schmid.
\newblock End-to-end generative pretraining for multimodal video captioning.
\newblock {\em arXiv preprint arXiv:2201.08264}, 2022.

\bibitem{Fan2022MMKD}
Fan Zhihao, Wei Zhongyu, Chen Jingjing, Wang Siyuan, Li~Zejun, Xu~Jiarong, and
  Huang Xuanjing.
\newblock A unified continuous learning framework for multi-modal knowledge
  discovery and pre-training.
\newblock {\em arXiv preprint arXiv:2206.05555}, 2022.

\bibitem{Zhang2022GLIPv2}
Zhang Haotian, Zhang Pengchuan, Hu~Xiaowei, Chen Yen-Chun, Harold~Li Liunian,
  Dai Xiyang, Wang Lijuan, Yuan Lu, Hwang Jenq-Neng, and Gao Jianfeng.
\newblock Glipv2: Unifying localization and vision-language understanding.
\newblock {\em arXiv preprint arXiv:2206.05836}, 2022.

\bibitem{Basil2022LIMoE}
Mustafa Basil, Riquelme Carlos, Puigcerver Joan, Jenatton Rodolphe, and Houlsby
  Neil.
\newblock Multimodal contrastive learning with limoe: the language-image
  mixture of experts.
\newblock {\em arXiv preprint arXiv:2206.02770}, 2022.

\bibitem{wang2022vlmixer}
Wang Teng, Jiang Wenhao, Lu~Zhichao, Zheng Feng, Cheng Ran, Yin Chengguo, and
  Ping Luo.
\newblock Vlmixer: Unpaired vision-language pre-training via cross-modal
  cutmix.
\newblock {\em arXiv preprint arXiv:2206.08919}, 2022.

\bibitem{NIPS2013_1cecc7a7}
Antoine Bordes, Nicolas Usunier, Alberto Garcia-Duran, Jason Weston, and Oksana
  Yakhnenko.
\newblock Translating embeddings for modeling multi-relational data.
\newblock In C.J. Burges, L.~Bottou, M.~Welling, Z.~Ghahramani, and K.Q.
  Weinberger, editors, {\em Advances in Neural Information Processing Systems},
  volume~26. Curran Associates, Inc., 2013.

\bibitem{Wang_Zhang_Feng_Chen_2014}
Zhen Wang, Jianwen Zhang, Jianlin Feng, and Zheng Chen.
\newblock Knowledge graph embedding by translating on hyperplanes.
\newblock {\em Proceedings of the AAAI Conference on Artificial Intelligence},
  28(1), Jun. 2014.

\bibitem{ji-etal-2015-knowledge}
Guoliang Ji, Shizhu He, Liheng Xu, Kang Liu, and Jun Zhao.
\newblock Knowledge graph embedding via dynamic mapping matrix.
\newblock In {\em Proceedings of the 53rd Annual Meeting of the Association for
  Computational Linguistics and the 7th International Joint Conference on
  Natural Language Processing (Volume 1: Long Papers)}, pages 687--696,
  Beijing, China, July 2015. Association for Computational Linguistics.

\bibitem{10.5555/2886521.2886624}
Yankai Lin, Zhiyuan Liu, Maosong Sun, Yang Liu, and Xuan Zhu.
\newblock Learning entity and relation embeddings for knowledge graph
  completion.
\newblock In {\em Proceedings of the Twenty-Ninth AAAI Conference on Artificial
  Intelligence}, AAAI'15, page 2181–2187. AAAI Press, 2015.

\bibitem{Ji_Liu_He_Zhao_2016}
Guoliang Ji, Kang Liu, Shizhu He, and Jun Zhao.
\newblock Knowledge graph completion with adaptive sparse transfer matrix.
\newblock {\em Proceedings of the AAAI Conference on Artificial Intelligence},
  30(1), Feb. 2016.

\bibitem{10.5555/3104482.3104584}
Maximilian Nickel, Volker Tresp, and Hans-Peter Kriegel.
\newblock A three-way model for collective learning on multi-relational data.
\newblock In {\em Proceedings of the 28th International Conference on
  International Conference on Machine Learning}, ICML'11, page 809–816,
  Madison, WI, USA, 2011. Omnipress.

\bibitem{NIPS2013_b337e84d}
Richard Socher, Danqi Chen, Christopher~D Manning, and Andrew Ng.
\newblock Reasoning with neural tensor networks for knowledge base completion.
\newblock In C.J. Burges, L.~Bottou, M.~Welling, Z.~Ghahramani, and K.Q.
  Weinberger, editors, {\em Advances in Neural Information Processing Systems},
  volume~26. Curran Associates, Inc., 2013.

\bibitem{https://doi.org/10.48550/arxiv.1412.6575}
Bishan Yang, Wen-tau Yih, Xiaodong He, Jianfeng Gao, and Li~Deng.
\newblock Embedding entities and relations for learning and inference in
  knowledge bases, 2014.

\bibitem{Bordes2014}
Antoine Bordes, Xavier Glorot, Jason Weston, and Yoshua Bengio.
\newblock A semantic matching energy function for learning with
  multi-relational data.
\newblock {\em Machine Learning}, 94(2):233--259, Feb 2014.

\bibitem{Nickel_Rosasco_Poggio_2016}
Maximilian Nickel, Lorenzo Rosasco, and Tomaso Poggio.
\newblock Holographic embeddings of knowledge graphs.
\newblock {\em Proceedings of the AAAI Conference on Artificial Intelligence},
  30(1), Mar. 2016.

\bibitem{ae482107de73461787258f805cf8f4ed}
Joan Bruna, Wojciech Zaremba, Arthur Szlam, and Yann Lecun.
\newblock Spectral networks and locally connected networks on graphs.
\newblock In {\em International Conference on Learning Representations
  (ICLR2014), CBLS, April 2014}, 2014.

\bibitem{kipf2017semi}
Thomas~N. Kipf and Max Welling.
\newblock Semi-supervised classification with graph convolutional networks.
\newblock In {\em International Conference on Learning Representations (ICLR)},
  2017.

\bibitem{https://doi.org/10.48550/arxiv.1611.07308}
Thomas~N. Kipf and Max Welling.
\newblock Variational graph auto-encoders, 2016.

\bibitem{NIPS2017_5dd9db5e}
Will Hamilton, Zhitao Ying, and Jure Leskovec.
\newblock Inductive representation learning on large graphs.
\newblock In I.~Guyon, U.~Von Luxburg, S.~Bengio, H.~Wallach, R.~Fergus,
  S.~Vishwanathan, and R.~Garnett, editors, {\em Advances in Neural Information
  Processing Systems}, volume~30. Curran Associates, Inc., 2017.

\bibitem{https://doi.org/10.48550/arxiv.1710.10903}
Petar Veličković, Guillem Cucurull, Arantxa Casanova, Adriana Romero, Pietro
  Liò, and Yoshua Bengio.
\newblock Graph attention networks, 2017.

\bibitem{10.1007/978-3-319-93417-4_38}
Michael Schlichtkrull, Thomas~N. Kipf, Peter Bloem, Rianne van den Berg, Ivan
  Titov, and Max Welling.
\newblock Modeling relational data with graph convolutional networks.
\newblock In Aldo Gangemi, Roberto Navigli, Maria-Esther Vidal, Pascal Hitzler,
  Rapha{\"e}l Troncy, Laura Hollink, Anna Tordai, and Mehwish Alam, editors,
  {\em The Semantic Web}, pages 593--607, Cham, 2018. Springer International
  Publishing.

\bibitem{Shang_Tang_Huang_Bi_He_Zhou_2019}
Chao Shang, Yun Tang, Jing Huang, Jinbo Bi, Xiaodong He, and Bowen Zhou.
\newblock End-to-end structure-aware convolutional networks for knowledge base
  completion.
\newblock {\em Proceedings of the AAAI Conference on Artificial Intelligence},
  33(01):3060--3067, Jul. 2019.

\bibitem{Dettmers_Minervini_Stenetorp_Riedel_2018}
Tim Dettmers, Pasquale Minervini, Pontus Stenetorp, and Sebastian Riedel.
\newblock Convolutional 2d knowledge graph embeddings.
\newblock {\em Proceedings of the AAAI Conference on Artificial Intelligence},
  32(1), Apr. 2018.

\bibitem{KBGAT2019}
Deepak Nathani, Jatin Chauhan, Charu Sharma, and Manohar Kaul.
\newblock Learning attention-based embeddings for relation prediction in
  knowledge graphs.
\newblock In {\em Proceedings of the 57th Annual Meeting of the Association for
  Computational Linguistics}. Association for Computational Linguistics, 2019.

\bibitem{vashishth2020compositionbased}
Shikhar Vashishth, Soumya Sanyal, Vikram Nitin, and Partha Talukdar.
\newblock Composition-based multi-relational graph convolutional networks.
\newblock In {\em International Conference on Learning Representations}, 2020.

\bibitem{li-etal-2020-enhancing}
Yanzeng Li, Bowen Yu, Xue Mengge, and Tingwen Liu.
\newblock Enhancing pre-trained {C}hinese character representation with
  word-aligned attention.
\newblock In {\em Proceedings of the 58th Annual Meeting of the Association for
  Computational Linguistics}, pages 3442--3448, Online, July 2020. Association
  for Computational Linguistics.

\bibitem{ke-etal-2020-sentilare}
Pei Ke, Haozhe Ji, Siyang Liu, Xiaoyan Zhu, and Minlie Huang.
\newblock {S}enti{LARE}: Sentiment-aware language representation learning with
  linguistic knowledge.
\newblock In {\em Proceedings of the 2020 Conference on Empirical Methods in
  Natural Language Processing (EMNLP)}, pages 6975--6988, Online, November
  2020. Association for Computational Linguistics.

\bibitem{roberts-etal-2020-much}
Adam Roberts, Colin Raffel, and Noam Shazeer.
\newblock How much knowledge can you pack into the parameters of a language
  model?
\newblock In {\em Proceedings of the 2020 Conference on Empirical Methods in
  Natural Language Processing (EMNLP)}, pages 5418--5426, Online, November
  2020. Association for Computational Linguistics.

\bibitem{sachan-etal-2021-syntax}
Devendra Sachan, Yuhao Zhang, Peng Qi, and William~L. Hamilton.
\newblock Do syntax trees help pre-trained transformers extract information?
\newblock In {\em Proceedings of the 16th Conference of the European Chapter of
  the Association for Computational Linguistics: Main Volume}, Online, April
  2021. Association for Computational Linguistics.

\bibitem{zhou-etal-2020-limit}
Junru Zhou, Zhuosheng Zhang, Hai Zhao, and Shuailiang Zhang.
\newblock {LIMIT}-{BERT} : Linguistics informed multi-task {BERT}.
\newblock In {\em Findings of the Association for Computational Linguistics:
  EMNLP 2020}, pages 4450--4461, Online, November 2020. Association for
  Computational Linguistics.

\bibitem{zhang2019ernie}
Zhengyan Zhang, Xu~Han, Zhiyuan Liu, Xin Jiang, Maosong Sun, and Qun Liu.
\newblock {ERNIE}: Enhanced language representation with informative entities.
\newblock In {\em Proceedings of ACL 2019}, 2019.

\bibitem{peters-etal-2019-knowledge}
Matthew~E. Peters, Mark Neumann, Robert Logan, Roy Schwartz, Vidur Joshi,
  Sameer Singh, and Noah~A. Smith.
\newblock Knowledge enhanced contextual word representations.
\newblock In {\em Proceedings of the 2019 Conference on Empirical Methods in
  Natural Language Processing and the 9th International Joint Conference on
  Natural Language Processing (EMNLP-IJCNLP)}, pages 43--54, Hong Kong, China,
  November 2019. Association for Computational Linguistics.

\bibitem{ijcai2017-179}
Peng Wang, Qi~Wu, Chunhua Shen, Anthony Dick, and Anton van~den Hengel.
\newblock Explicit knowledge-based reasoning for visual question answering.
\newblock In {\em Proceedings of the Twenty-Sixth International Joint
  Conference on Artificial Intelligence, {IJCAI-17}}, pages 1290--1296, 2017.

\bibitem{8046084}
Peng Wang, Qi~Wu, Chunhua Shen, Anthony Dick, and Anton van~den Hengel.
\newblock Fvqa: Fact-based visual question answering.
\newblock {\em IEEE Transactions on Pattern Analysis and Machine Intelligence},
  40(10):2413--2427, 2018.

\bibitem{10.1007/978-3-319-10590-1_4}
Jia Deng, Nan Ding, Yangqing Jia, Andrea Frome, Kevin Murphy, Samy Bengio, Yuan
  Li, Hartmut Neven, and Hartwig Adam.
\newblock Large-scale object classification using label relation graphs.
\newblock In David Fleet, Tomas Pajdla, Bernt Schiele, and Tinne Tuytelaars,
  editors, {\em Computer Vision -- ECCV 2014}, pages 48--64, Cham, 2014.
  Springer International Publishing.

\bibitem{kwiatkowski-etal-2019-natural}
Tom Kwiatkowski, Jennimaria Palomaki, Olivia Redfield, Michael Collins, Ankur
  Parikh, Chris Alberti, Danielle Epstein, Illia Polosukhin, Jacob Devlin,
  Kenton Lee, Kristina Toutanova, Llion Jones, Matthew Kelcey, Ming-Wei Chang,
  Andrew~M. Dai, Jakob Uszkoreit, Quoc Le, and Slav Petrov.
\newblock Natural questions: A benchmark for question answering research.
\newblock {\em Transactions of the Association for Computational Linguistics},
  7:452--466, 2019.

\bibitem{yang-etal-2018-hotpotqa}
Zhilin Yang, Peng Qi, Saizheng Zhang, Yoshua Bengio, William Cohen, Ruslan
  Salakhutdinov, and Christopher~D. Manning.
\newblock {H}otpot{QA}: A dataset for diverse, explainable multi-hop question
  answering.
\newblock In {\em Proceedings of the 2018 Conference on Empirical Methods in
  Natural Language Processing}, pages 2369--2380, Brussels, Belgium,
  October-November 2018. Association for Computational Linguistics.

\bibitem{clark-etal-2019-boolq}
Christopher Clark, Kenton Lee, Ming-Wei Chang, Tom Kwiatkowski, Michael
  Collins, and Kristina Toutanova.
\newblock {B}ool{Q}: Exploring the surprising difficulty of natural yes/no
  questions.
\newblock In {\em Proceedings of the 2019 Conference of the North {A}merican
  Chapter of the Association for Computational Linguistics: Human Language
  Technologies, Volume 1 (Long and Short Papers)}, pages 2924--2936,
  Minneapolis, Minnesota, June 2019. Association for Computational Linguistics.

\bibitem{thorne-etal-2018-fever}
James Thorne, Andreas Vlachos, Christos Christodoulopoulos, and Arpit Mittal.
\newblock {FEVER}: a large-scale dataset for fact extraction and
  {VER}ification.
\newblock In {\em Proceedings of the 2018 Conference of the North {A}merican
  Chapter of the Association for Computational Linguistics: Human Language
  Technologies, Volume 1 (Long Papers)}, pages 809--819, New Orleans,
  Louisiana, June 2018. Association for Computational Linguistics.

\bibitem{10.1145/2661829.2661887}
Zhaochen Guo and Denilson Barbosa.
\newblock Robust entity linking via random walks.
\newblock In {\em Proceedings of the 23rd ACM International Conference on
  Conference on Information and Knowledge Management}, CIKM '14, page
  499–508, New York, NY, USA, 2014. Association for Computing Machinery.

\bibitem{talmor-etal-2019-commonsenseqa}
Alon Talmor, Jonathan Herzig, Nicholas Lourie, and Jonathan Berant.
\newblock {C}ommonsense{QA}: A question answering challenge targeting
  commonsense knowledge.
\newblock In {\em Proceedings of the 2019 Conference of the North {A}merican
  Chapter of the Association for Computational Linguistics: Human Language
  Technologies, Volume 1 (Long and Short Papers)}, pages 4149--4158,
  Minneapolis, Minnesota, June 2019. Association for Computational Linguistics.

\bibitem{bhagavatula2020abductive}
Chandra Bhagavatula, Ronan~Le Bras, Chaitanya Malaviya, Keisuke Sakaguchi, Ari
  Holtzman, Hannah Rashkin, Doug Downey, Wen tau Yih, and Yejin Choi.
\newblock Abductive commonsense reasoning.
\newblock In {\em International Conference on Learning Representations}, 2020.

\bibitem{lin-etal-2020-commongen}
Bill~Yuchen Lin, Wangchunshu Zhou, Ming Shen, Pei Zhou, Chandra Bhagavatula,
  Yejin Choi, and Xiang Ren.
\newblock {C}ommon{G}en: A constrained text generation challenge for generative
  commonsense reasoning.
\newblock In {\em Findings of the Association for Computational Linguistics:
  EMNLP 2020}, pages 1823--1840, Online, November 2020. Association for
  Computational Linguistics.

\bibitem{sap-etal-2019-social}
Maarten Sap, Hannah Rashkin, Derek Chen, Ronan Le~Bras, and Yejin Choi.
\newblock Social {IQ}a: Commonsense reasoning about social interactions.
\newblock In {\em Proceedings of the 2019 Conference on Empirical Methods in
  Natural Language Processing and the 9th International Joint Conference on
  Natural Language Processing (EMNLP-IJCNLP)}, pages 4463--4473, Hong Kong,
  China, November 2019. Association for Computational Linguistics.

\bibitem{Bisk_Zellers_Lebras_Gao_Choi_2020}
Yonatan Bisk, Rowan Zellers, Ronan Le~bras, Jianfeng Gao, and Yejin Choi.
\newblock Piqa: Reasoning about physical commonsense in natural language.
\newblock {\em Proceedings of the AAAI Conference on Artificial Intelligence},
  34(05):7432--7439, Apr. 2020.

\bibitem{zhou-etal-2019-going}
Ben Zhou, Daniel Khashabi, Qiang Ning, and Dan Roth.
\newblock {``}going on a vacation{''} takes longer than {``}going for a
  walk{''}: A study of temporal commonsense understanding.
\newblock In {\em Proceedings of the 2019 Conference on Empirical Methods in
  Natural Language Processing and the 9th International Joint Conference on
  Natural Language Processing (EMNLP-IJCNLP)}, pages 3363--3369, Hong Kong,
  China, November 2019. Association for Computational Linguistics.

\bibitem{ZRNKSR21}
Ben Zhou, Kyle Richardson, Qiang Ning, Tushar Khot, Ashish Sabharwal, and Dan
  Roth.
\newblock Temporal reasoning on implicit events from distant supervision.
\newblock In {\em NAACL}, 2021.

\bibitem{agrawal2019nocaps}
Harsh Agrawal, Karan Desai, Yufei Wang, Xinlei Chen, Rishabh Jain, Mark
  Johnson, Dhruv Batra, Devi Parikh, Stefan Lee, and Peter Anderson.
\newblock Nocaps: Novel object captioning at scale.
\newblock In {\em Proceedings of the IEEE/CVF International Conference on
  Computer Vision}, pages 8948--8957, 2019.

\bibitem{das2017visualDialog}
Abhishek Das, Satwik Kottur, Khushi Gupta, Avi Singh, Deshraj Yadav,
  Jos{\'e}~MF Moura, Devi Parikh, and Dhruv Batra.
\newblock Visual dialog.
\newblock In {\em Proceedings of the IEEE conference on computer vision and
  pattern recognition}, pages 326--335, 2017.

\bibitem{yang2020visualMMMT}
Pengcheng Yang, Boxing Chen, Pei Zhang, and Xu~Sun.
\newblock Visual agreement regularized training for multi-modal machine
  translation.
\newblock In {\em Proceedings of the AAAI Conference on Artificial
  Intelligence}, volume~34, pages 9418--9425, 2020.

\bibitem{antol2015vqa}
Stanislaw Antol, Aishwarya Agrawal, Jiasen Lu, Margaret Mitchell, Dhruv Batra,
  C~Lawrence Zitnick, and Devi Parikh.
\newblock Vqa: Visual question answering.
\newblock In {\em Proceedings of the IEEE international conference on computer
  vision}, pages 2425--2433, 2015.

\bibitem{liu2020violin}
Jingzhou Liu, Wenhu Chen, Yu~Cheng, Zhe Gan, Licheng Yu, Yiming Yang, and
  Jingjing Liu.
\newblock Violin: A large-scale dataset for video-and-language inference.
\newblock In {\em Proceedings of the IEEE/CVF Conference on Computer Vision and
  Pattern Recognition}, pages 10900--10910, 2020.

\bibitem{suhr2017corpus}
Alane Suhr, Mike Lewis, James Yeh, and Yoav Artzi.
\newblock A corpus of natural language for visual reasoning.
\newblock In {\em Proceedings of the 55th Annual Meeting of the Association for
  Computational Linguistics (Volume 2: Short Papers)}, pages 217--223, 2017.

\bibitem{xie2019visualentailment}
Ning Xie, Farley Lai, Derek Doran, and Asim Kadav.
\newblock Visual entailment: A novel task for fine-grained image understanding.
\newblock {\em arXiv preprint arXiv:1901.06706}, 2019.

\bibitem{dagan2005textentailment}
Ido Dagan, Oren Glickman, and Bernardo Magnini.
\newblock The pascal recognising textual entailment challenge.
\newblock In {\em Machine Learning Challenges Workshop}, pages 177--190.
  Springer, 2005.

\bibitem{zellers2019VCR}
Rowan Zellers, Yonatan Bisk, Ali Farhadi, and Yejin Choi.
\newblock From recognition to cognition: Visual commonsense reasoning.
\newblock In {\em Proceedings of the IEEE/CVF conference on computer vision and
  pattern recognition}, pages 6720--6731, 2019.

\bibitem{wang2022pedestriansurvey}
Xiao Wang, Shaofei Zheng, Rui Yang, Aihua Zheng, Zhe Chen, Jin Tang, and Bin
  Luo.
\newblock Pedestrian attribute recognition: A survey.
\newblock {\em Pattern Recognition}, 121:108220, 2022.

\bibitem{ghosal2018contextual}
Deepanway Ghosal, Md~Shad Akhtar, Dushyant Chauhan, Soujanya Poria, Asif Ekbal,
  and Pushpak Bhattacharyya.
\newblock Contextual inter-modal attention for multi-modal sentiment analysis.
\newblock In {\em proceedings of the 2018 conference on empirical methods in
  natural language processing}, pages 3454--3466, 2018.

\bibitem{li2017personsearch}
Shuang Li, Tong Xiao, Hongsheng Li, Bolei Zhou, Dayu Yue, and Xiaogang Wang.
\newblock Person search with natural language description.
\newblock In {\em Proceedings of the IEEE Conference on Computer Vision and
  Pattern Recognition}, pages 1970--1979, 2017.

\bibitem{chen2021deepimgretievalsurvey}
Wei Chen, Yang Liu, Weiping Wang, Erwin~M Bakker, TK~Georgiou, Paul Fieguth,
  Li~Liu, and MSK Lew.
\newblock Deep image retrieval: A survey.
\newblock {\em ArXiv}, 2021.

\bibitem{gu2022VLNavigationSurvey}
Jing Gu, Eliana Stefani, Qi~Wu, Jesse Thomason, and Xin~Eric Wang.
\newblock Vision-and-language navigation: A survey of tasks, methods, and
  future directions.
\newblock {\em arXiv preprint arXiv:2203.12667}, 2022.

\bibitem{park2022visualnavigation}
Sang-Min Park and Young-Gab Kim.
\newblock Visual language navigation: a survey and open challenges.
\newblock {\em Artificial Intelligence Review}, pages 1--63, 2022.

\bibitem{zhang2018grounding}
Hanwang Zhang, Yulei Niu, and Shih-Fu Chang.
\newblock Grounding referring expressions in images by variational context.
\newblock In {\em Proceedings of the IEEE Conference on Computer Vision and
  Pattern Recognition}, pages 4158--4166, 2018.

\bibitem{yang2019GRE}
Sibei Yang, Guanbin Li, and Yizhou Yu.
\newblock Cross-modal relationship inference for grounding referring
  expressions.
\newblock In {\em Proceedings of the IEEE/CVF Conference on Computer Vision and
  Pattern Recognition}, pages 4145--4154, 2019.

\bibitem{ding2022exploringgrounding}
Xinpeng Ding, Nannan Wang, Shiwei Zhang, Ziyuan Huang, Xiaomeng Li, Mingqian
  Tang, Tongliang Liu, and Xinbo Gao.
\newblock Exploring language hierarchy for video grounding.
\newblock {\em IEEE Transactions on Image Processing}, 31:4693--4706, 2022.

\bibitem{tang2021HCSTVG}
Zongheng Tang, Yue Liao, Si~Liu, Guanbin Li, Xiaojie Jin, Hongxu Jiang, Qian
  Yu, and Dong Xu.
\newblock Human-centric spatio-temporal video grounding with visual
  transformers.
\newblock {\em IEEE Transactions on Circuits and Systems for Video Technology},
  2021.

\bibitem{wang2021TNL2K}
Xiao Wang, Xiujun Shu, Zhipeng Zhang, Bo~Jiang, Yaowei Wang, Yonghong Tian, and
  Feng Wu.
\newblock Towards more flexible and accurate object tracking with natural
  language: Algorithms and benchmark.
\newblock In {\em Proceedings of the IEEE/CVF Conference on Computer Vision and
  Pattern Recognition}, pages 13763--13773, 2021.

\bibitem{wang2018describe}
Xiao Wang, Chenglong Li, Rui Yang, Tianzhu Zhang, Jin Tang, and Bin Luo.
\newblock Describe and attend to track: Learning natural language guided
  structural representation and visual attention for object tracking.
\newblock {\em arXiv preprint arXiv:1811.10014}, 2018.

\bibitem{feng2021SNLT}
Qi~Feng, Vitaly Ablavsky, Qinxun Bai, and Stan Sclaroff.
\newblock Siamese natural language tracker: Tracking by natural language
  descriptions with siamese trackers.
\newblock In {\em Proceedings of the IEEE/CVF Conference on Computer Vision and
  Pattern Recognition}, pages 5851--5860, 2021.

\bibitem{yao2021cpt}
Yuan Yao, Ao~Zhang, Zhengyan Zhang, Zhiyuan Liu, Tat-Seng Chua, and Maosong
  Sun.
\newblock Cpt: Colorful prompt tuning for pre-trained vision-language models.
\newblock {\em arXiv preprint arXiv:2109.11797}, 2021.

\bibitem{he2022cpl}
Xuehai He, Diji Yang, Weixi Feng, Tsu-Jui Fu, Arjun Akula, Varun Jampani,
  Pradyumna Narayana, Sugato Basu, William~Yang Wang, and Xin~Eric Wang.
\newblock Cpl: Counterfactual prompt learning for vision and language models.
\newblock {\em arXiv preprint arXiv:2210.10362}, 2022.

\bibitem{jia2022VPT}
Menglin Jia, Luming Tang, Bor-Chun Chen, Claire Cardie, Serge Belongie, Bharath
  Hariharan, and Ser-Nam Lim.
\newblock Visual prompt tuning.
\newblock {\em arXiv preprint arXiv:2203.12119}, 2022.

\bibitem{zhou2021learningprompt}
Kaiyang Zhou, Jingkang Yang, Chen~Change Loy, and Ziwei Liu.
\newblock Learning to prompt for vision-language models.
\newblock {\em International Journal of Computer Vision}, 130(9):2337--2348,
  2022.

\bibitem{zhou2022conditionalPrompt}
Kaiyang Zhou, Jingkang Yang, Chen~Change Loy, and Ziwei Liu.
\newblock Conditional prompt learning for vision-language models.
\newblock In {\em Proceedings of the IEEE/CVF Conference on Computer Vision and
  Pattern Recognition}, pages 16816--16825, 2022.

\bibitem{liu2021promptlearning}
Pengfei Liu, Weizhe Yuan, Jinlan Fu, Zhengbao Jiang, Hiroaki Hayashi, and
  Graham Neubig.
\newblock Pre-train, prompt, and predict: A systematic survey of prompting
  methods in natural language processing.
\newblock {\em arXiv preprint arXiv:2107.13586}, 2021.

\bibitem{wang2015robust}
Qingzheng Wang, Shuai Li, Hong Qin, and Aimin Hao.
\newblock Robust multi-modal medical image fusion via anisotropic heat
  diffusion guided low-rank structural analysis.
\newblock {\em Information fusion}, 26:103--121, 2015.

\bibitem{wang2022MFGNet}
Xiao Wang, Xiujun Shu, Shilliang Zhang, Bo~Jiang, Yaowei Wang, Yonghong Tian,
  and Feng Wu.
\newblock Mfgnet: Dynamic modality-aware filter generation for rgb-t tracking.
\newblock {\em IEEE Transactions on Multimedia}, 2022.

\bibitem{lee2018stacked}
Kuang-Huei Lee, Xi~Chen, Gang Hua, Houdong Hu, and Xiaodong He.
\newblock Stacked cross attention for image-text matching.
\newblock In {\em Proceedings of the European Conference on Computer Vision
  (ECCV)}, pages 201--216, 2018.

\end{thebibliography}
}

\end{document}